\documentclass{article}

\usepackage{longcat_style}

\settitlelogo{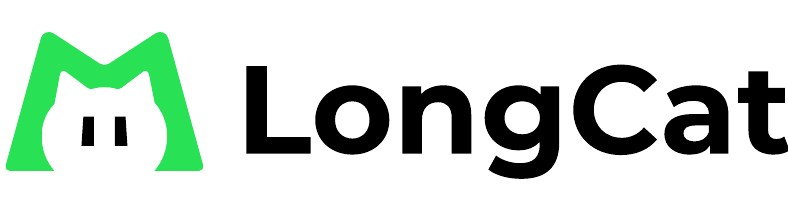}

\renewcommand{\headeright}{Preprint}
\renewcommand{\shorttitle}{\benchmark: A Benchmark for Interactive Video World Model Evaluation}

\usepackage[utf8]{inputenc}
\usepackage[T1]{fontenc}
\usepackage{hyperref}
\usepackage{xcolor}
\hypersetup{
    colorlinks=true,
    linkcolor=blue,
    urlcolor=blue,
    citecolor=blue,
}

\usepackage{float}
\usepackage[section]{placeins}
\usepackage{url}
\usepackage{booktabs}
\usepackage{amsmath}
\usepackage{amsfonts}
\usepackage{nicefrac}
\usepackage{microtype}
\usepackage{graphicx}
\usepackage[numbers,compress]{natbib}
\usepackage{doi}
\usepackage{xspace}
\usepackage{enumitem}
\usepackage{array}
\usepackage{adjustbox}
\usepackage{colortbl}
\usepackage{multirow}
\usepackage{makecell}
\usepackage{pifont}
\usepackage{subcaption}
\usepackage{fontawesome5}
\usepackage[most]{tcolorbox}
\tcbuselibrary{skins,breakable}
\usepackage{soul}
\setuldepth{a}
\usepackage{titletoc}
\usepackage[capitalize]{cleveref}

\setlist[itemize]{leftmargin=*}
\setlist[enumerate]{leftmargin=*}

\newcommand{\pub}[1]{\color{gray}{\scriptsize{[{#1}]}}}

\definecolor{badgegray}{RGB}{246,247,249}
\definecolor{badgeborder}{RGB}{215,218,224}
\definecolor{badgetext}{RGB}{45,50,60}
\definecolor{githubblack}{RGB}{36,41,47}
\definecolor{hfamber}{RGB}{255,188,66}
\definecolor{homepageblue}{RGB}{66,133,244}

\newtcbox{\linkbadgebox}{
  on line,
  arc=8pt,
  boxrule=0.4pt,
  colback=badgegray,
  colframe=badgeborder,
  boxsep=1pt,
  left=5pt,
  right=5pt,
  top=2pt,
  bottom=2pt,
  tcbox raise base,
  nobeforeafter
}

\newcommand{\linkbadge}[3]{%
  \href{#1}{%
    \linkbadgebox{%
      \textcolor{#2}{#3}%
    }%
  }%
}

\newcommand{\homepage}[1]{%
  \linkbadge{#1}{badgetext}{\textcolor{homepageblue}{\faGlobe}~\textbf{Homepage}}%
}

\newcommand{\github}[1]{%
  \linkbadge{#1}{badgetext}{\textcolor{githubblack}{\faGithub}~\textbf{GitHub}}%
}

\newcommand{\huggingface}[1]{%
  \linkbadge{#1}{badgetext}{%
    \raisebox{-0.12em}{\includegraphics[height=1em]{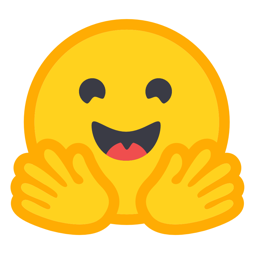}}~\textbf{HuggingFace}%
  }%
}

\emergencystretch=\maxdimen
\providecommand{\thinslash}{\hspace{1pt}/\hspace{1pt}}
\providecommand{\tagbox}[2]{\smash{\setlength{\fboxsep}{2.3pt}\raisebox{0.5pt}{\colorbox{#1}{#2}}}\vphantom{#2}}

\providecolor{dimVQ}{RGB}{191,219,254}
\providecolor{dimSA}{RGB}{187,247,208}
\providecolor{dimIA}{RGB}{253,213,159}
\providecolor{dimCO}{RGB}{216,180,254}
\providecolor{dimPH}{RGB}{253,182,182}

\newcommand{\cmark}{\ding{51}}
\newcommand{\xmark}{\textcolor{gray!25}{\ding{55}}}

\makeatletter
\newcommand{\thickhline}{%
    \noalign{\ifnum0=`}\fi \hrule height 1pt
    \futurelet\reserved@a\@xhline
}

\newcommand{\mytoprule}{%
    \noalign{\ifnum0=`}\fi
    \vskip\abovetopsep
    \hrule height\heavyrulewidth
    \vskip\belowrulesep
    \futurelet\reserved@a\@xhline
}
\newcommand{\mymidrule}{%
    \noalign{\ifnum0=`}\fi
    \vskip\aboverulesep
    \hrule height\lightrulewidth
    \vskip\belowrulesep
    \futurelet\reserved@a\@xhline
}
\newcommand{\mybottomrule}{%
    \noalign{\ifnum0=`}\fi
    \vskip\aboverulesep
    \hrule height\heavyrulewidth
    \vskip\belowbottomsep
    \futurelet\reserved@a\@xhline
}
\makeatother

\newcommand{\vs}{\textit{vs.}\xspace}
\newcommand{\myparagraph}[1]{\vspace{-0.1em}\noindent\textbf{#1}\hspace{0.3em}}

\newcommand{\icon}[1]{\raisebox{-0.15em}{\makebox[1.2em][c]{\includegraphics[height=1em,width=1em,keepaspectratio]{figures/icon/#1}}}\hspace{0.2em}}

\DeclareRobustCommand{\benchmark}{\textsc{WBench}\xspace}
\newcommand{\DoAppendixToC}{%
  \ifdefined\startcontents
    \startcontents
    \printcontents{}{1}{\noindent\textbf{\Large{Appendix Contents}}\par\vspace{0.5em}}%
    \vspace{0.5em}%
  \fi
}

\newcommand{\eg}{\textit{e.g.}\xspace}

\newcommand{\etc}{\textit{etc.}\xspace}

\newcommand{\rk}[1]{\rlap{\,{\fontsize{4}{4.5}\selectfont\ifnum#1>5\relax\textcolor{gray!40}{(#1)}\else(#1)\fi}}}

\newcommand{\nummodel}{{20}\xspace}
\newcommand{\numvideo}{289\xspace}
\newcommand{\numturn}{1,058\xspace}
\newcommand{\numnavi}{158\xspace}
\newcommand{\numsubmetric}{{22}\xspace}

\title{\benchmark: A Comprehensive Multi-turn Benchmark for Interactive Video World Model Evaluation}

\author{
  Kaining Ying$^{1}$\thanks{Equal contribution.} \quad
  Hengrui Hu$^{1}$\footnotemark[1] \quad
  Siyu Ren$^{2}$ \quad
  Jiamu Li$^{2}$ \quad
  Fengjiao Chen$^{2}$ \\
  \bf{Ziwen Wang$^{2}$ \quad Xuezhi Cao$^{2}$ \quad Xunliang Cai$^{2}$ \quad Henghui Ding$^{1}$\thanks{Correspondence: Henghui Ding \textless\texttt{hhding@fudan.edu.cn}\textgreater}} \\[4pt]
  $^{1}$Fudan University \quad $^{2}$Meituan Longcat Team \\[7pt]
  \homepage{https://meituan-longcat.github.io/WBench}
  \quad
  \github{https://github.com/meituan-longcat/WBench}
  \quad
  \huggingface{https://huggingface.co/datasets/meituan-longcat/wbench}
}

\begin{document}
\maketitle

\begin{figure}[H]
\centering
\includegraphics[width=1.0\linewidth]{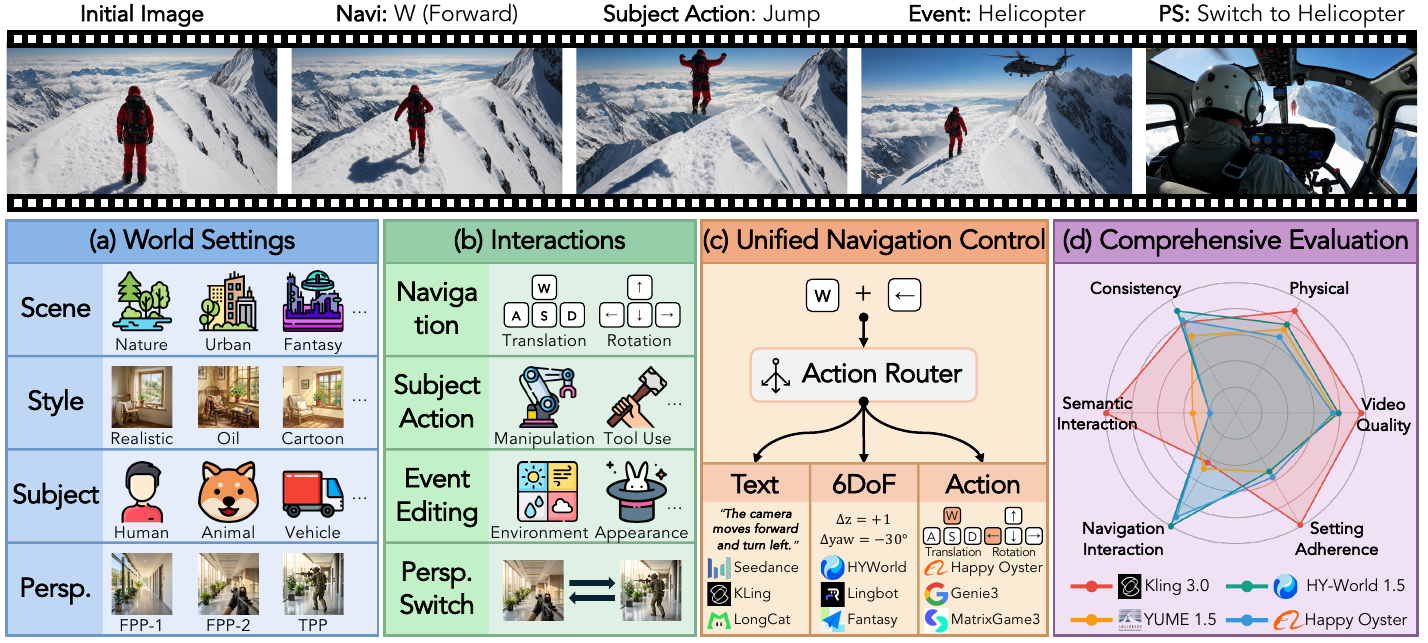}
\vspace{-5mm}
\caption[Overview of \benchmark]{Overview of \textbf{\benchmark}. \textbf{Top:} a multi-turn case with navigation, subject action, event editing, and perspective switching. \textbf{Bottom:} the benchmark design, including world settings, interaction taxonomy, unified navigation control, and evaluation over video quality, setting adherence, interaction adherence~(navigation and semantic interactions), consistency, and physics compliance.}
\label{fig:teaser}
\end{figure}

\begin{abstract}
Interactive world models are advancing rapidly, yet existing benchmarks cover only part of the required competencies, leaving no unified standard for systematic evaluation.
To fill this gap, we introduce \textbf{\benchmark}, a comprehensive multi-turn benchmark for interactive world model evaluation along five dimensions, namely video quality, setting adherence, interaction adherence, consistency, and physics compliance.
\benchmark contains \numvideo test cases and \numturn interaction turns, where each case specifies a world setting and a multi-turn interaction sequence, covering diverse scenes, styles, subjects, and both first- and third-person perspectives, together with four interaction types, including navigation, subject action, event editing, and perspective switching.
For navigation, \benchmark unifies text, 6-DoF pose, and discrete-action control, enabling evaluation of models with different native input interfaces.
Evaluation uses \numsubmetric automatic sub-metrics that combine specialist vision models with large multimodal models, and all metrics are validated against human judgments.
Across \nummodel state-of-the-art models, we find that no single model performs strongly across all dimensions. We provide detailed diagnostic insights into the characteristic strengths, weaknesses, and open challenges of each model.
Code and data are available at \href{https://github.com/meituan-longcat/WBench}{\texttt{\textcolor{blue}{https://github.com/meituan-longcat/WBench}}}.

\end{abstract}

\newpage
\tableofcontents
\newpage

\section{Introduction}
\label{sec:introduction}

Recent advances in video generation~\cite{ho2020denoising,kong2024hunyuanvideo,wan2025wan,polyak2024moviegen,brooks2024sora} have enabled interactive world models with controllable generation across games~\cite{bruce2024genie,deepmind2024genie2,ball2025genie3,he2025matrixgame2,he2026matrixgame3,li2025hunyuangamecraft,tang2025interbench}, autonomous driving~\cite{hu2023gaia1,gao2024vista}, embodied interaction~\cite{yang2023unisim,zhu2024irasim}, and open-domain scenarios~\cite{sun2025hyworldplay,chen2025yume,mao2025yume15,gao2026lingbot}.
However, evaluation remains fragmented, with many works relying on selected demos or task-specific protocols, making fair comparison and failure diagnosis difficult across visual quality, controllability, memory, and physics.

A capable interactive world model must fulfill five complementary roles, analogous to the subsystems of a game engine: a \emph{Renderer} for visually convincing video, a \emph{Director} for correct world initialization, a \emph{Controller} for faithful interaction execution, a \emph{Memory} for preserving world state across turns, and an \emph{Engine} for physically compliant world evolution. Existing benchmarks cover these roles only partially (\cref{tab:benchmark_comparison}). Video-generation benchmarks such as VBench~\cite{huang2024vbench,zheng2025vbench2} focus on perceptual quality without interactive control. World-model benchmarks evaluate more dimensions but remain limited in scope: WorldMark~\cite{xu2026worldmark} and MIND~\cite{ye2026mind} cover navigation and memory but lack semantic interactions, Omni-WorldBench~\cite{wu2026omniworldbench} adds causal interaction but supports only first-person view, and WorldLens~\cite{worldlens2025} evaluates multiple dimensions but is restricted to autonomous driving. None provides a unified protocol spanning open-domain scenes, both perspectives, and all four interaction types.

To address this gap, we introduce \benchmark, a comprehensive multi-turn benchmark for interactive world model evaluation. As shown in \cref{fig:teaser}, each test case is defined by a \emph{world setting} (scene, subject, style, and perspective) together with a multi-turn \emph{interaction} sequence. The top row illustrates a concrete case: a realistic snowy mountain scene with a human subject in third-person perspective, followed by forward navigation, a jump, the appearance of a helicopter, and a perspective switching to the cockpit. More broadly, the benchmark spans diverse open-domain scenes, rendering styles, subject categories, and both first- and third-person perspectives (\cref{fig:teaser}~\!(a)), with four interaction types shown in \cref{fig:teaser}~\!(b): navigation, subject action, event editing, and perspective switching. This design separates what the world \emph{is} from what the user \emph{requests}, making failure modes easier to locate: a model may render the initial scene well but ignore later actions, or follow a single instruction correctly but lose identity and spatial consistency over multiple turns.

\benchmark also supports fair comparison across different control paradigms. As shown in \cref{fig:teaser}~\!(c), navigation interactions are represented in three aligned forms, namely text, camera pose, and discrete action, so that models can be evaluated through their native interfaces. Accordingly, we adopt a dual-track evaluation protocol: all \nummodel models are compared on a shared navigation subset of \numnavi cases, while text-prompted I2V models are further evaluated on the full benchmark (\numvideo cases, \numturn turns). Evaluation uses \numsubmetric automatic sub-metrics combining specialist vision models and VLMs.

Experiments on \nummodel models reveal that:
1)~no model dominates all five dimensions,
2)~navigation is largely independent of other dimensions,
3)~camera control and perspective consistency are separate capabilities,
4)~physical correctness correlates with rendering quality rather than control,
5)~benchmark difficulty is structured by perspective, scene type, and subject category, and
6)~four interaction types degrade unevenly over turns, with navigation most fragile.

Our contributions are: 1)~a unified benchmark spanning five complementary evaluation dimensions with \numsubmetric fine-grained sub-metrics, 2)~a multi-turn dataset covering both perspectives, four interaction types, and a unified navigation interface enabling fair cross-paradigm comparison, and 3)~a fully automatic evaluation pipeline applied to \nummodel models, establishing diagnostic baselines and surfacing actionable insights for future model development.

\section{Related Work}
\label{sec:related_work}

\textbf{Video Generation Models.}
Video generation has evolved rapidly, from early U-Net-based diffusion models~\cite{ho2020denoising,ho2022video,blattmann2023align} to scalable Diffusion Transformers~\cite{yang2024cogvideox,polyak2024moviegen,kong2024hunyuanvideo} trained with flow-matching objectives on large-scale data, yielding longer, higher-resolution, and temporally coherent outputs.
Building on this foundation, the current frontier like Sora~2~\cite{openai2025sora2}, Kling~3.0~\cite{kuaishou2025kling3}, Veo~3~\cite{deepmind2025veo3}, Wan~2.7~\cite{wan2025wan27}, and others~\cite{gao2026seedance2,shengshu2025viduq3,kong2024hunyuanvideo,lightricks2025ltx,nvidia2025cosmos,cai2025longcat}, collectively advance cinematic quality, prompt adherence, efficient inference, physical grounding, and long-horizon continuation.
Despite these advances, evaluation still centers on distributional metrics (FID~\cite{heusel2017fid}, FVD~\cite{unterthiner2019fvd}), text-alignment scores, or multi-dimensional quality suites~\cite{huang2024vbench}, none of which probe interactive controllability or world-modeling competence.

\textbf{Interactive Video World Models.}
World models~\cite{ha2018world,lecun2022path} predict environment evolution in response to actions. While traditionally realized as latent state-space models, recent video generators have enabled a new paradigm: \emph{interactive video world models} that directly synthesize next frames from the current observation and an action signal, enabling closed-loop simulation. Although early systems also appeared in robotic manipulation and autonomous driving, such as UniSim~\cite{yang2023unisim}, IRASim~\cite{zhu2024irasim}, GAIA-1~\cite{hu2023gaia1}, and Vista~\cite{gao2024vista}, we focus on the open-domain branch most relevant to \benchmark.
Among world models evaluated in this work, YUME~1.5~\cite{mao2025yume15} represents language-driven interaction, using natural-language actions for multi-turn world evolution, while HY-World~1.5~\cite{sun2025hyworldplay} and LingBot-World~\cite{gao2026lingbot} represent camera-controlled generation with an emphasis on navigation and geometric consistency.
Action-conditioned systems such as Hunyuan-GameCraft~\cite{li2025hunyuangamecraft,tang2025interbench}, Matrix-Game~2.0~\cite{he2025matrixgame2}, and Matrix-Game~3.0~\cite{he2026matrixgame3} push real-time keyboard-and-mouse control, with Matrix-Game~3.0 further improving long-horizon consistency through explicit memory. Closed-source systems such as Genie~3~\cite{ball2025genie3}, Happy~Oyster~\cite{alibaba2026happyoyster}, and Marble~\cite{worldlabs2025marble} further highlight the momentum of this area.

\textbf{World Model Evaluation.}
As shown in \cref{tab:benchmark_comparison}, existing benchmarks fall into two broad groups.
Non-interactive suites such as VBench~\cite{huang2024vbench,huang2024vbenchplus,zheng2025vbench2}, EvalCrafter~\cite{liu2024evalcrafter}, and VideoPhy~\cite{bansal2024videophy,meng2024phygenbench} assess video quality, text alignment, or physical commonsense, but do not take action inputs or evaluate multi-turn interaction.
Among world model benchmarks, WorldScore~\cite{duan2024worldscore} evaluates camera-trajectory-conditioned generation, WorldModelBench~\cite{wang2024worldmodelbench} studies decision-oriented world-model quality, WorldArena~\cite{shang2026worldarena} targets embodied agents in closed domains, MIND~\cite{ye2026mind} probes closed-loop memory consistency, Omni-WorldBench~\cite{wu2026omniworldbench} focuses on causal interaction, WorldLens~\cite{worldlens2025} targets autonomous driving, and WorldMark~\cite{xu2026worldmark} measures navigation consistency.
Additional efforts examine complementary aspects~\cite{han2025videobench,liu2024fetv,gu2025phyworldbench,cai2025mmgr,upadhyay2026worldbench,lu2025_4dworldbench,gao2026wrena,lian2024worldsimbench,yue2025ewmbench,zhang2025worldinworld,tang2025interbench,zhou2026drivinggen}.
Despite this rapid progress, no existing benchmark jointly covers (i) diverse open-domain scenes, (ii) both first- and third-person perspectives with perspective-dependent action semantics, (iii) a comprehensive interaction taxonomy spanning navigation, subject action, event editing, and perspective switching, and (iv) multi-turn closed-loop evaluation targeting long-horizon consistency and physics compliance.
\benchmark fills this gap with a unified framework across all four axes, instantiated through \numsubmetric fine-grained automatic sub-metrics.

\begin{table}[t]
\caption[Comparison with representative benchmarks]{Comparison with representative benchmarks. Input: {\small\textsf{\textbf{T}}}\,=\,text, {\small\faVideo}\,=\,camera, {\small\faGamepad}\,=\,action. FPP\hspace{1pt}/\hspace{1pt}TPP: first-\hspace{1pt}/\hspace{1pt}third-person perspective. Navi, SA, EE, PS: navigation, subject action, event editing, and perspective switching. Qual, Adh, Inter, Cons, Phys: video quality, setting adherence, interaction adherence, consistency, and physics compliance. Full comparison is provided in \cref{tab:benchmark_comparison_full}.}
\vspace{+1mm}
\label{tab:benchmark_comparison}
\centering
\renewcommand{\arraystretch}{1.05}
\footnotesize
\setlength{\tabcolsep}{4.4pt}
\newcommand{\iOn}[1]{#1}
\newcommand{\iOff}[1]{\textcolor{black!15}{#1}}
\begin{tabular}{@{}l c cc cccc ccccc rr@{}}
\toprule
\multirow{2.5}{*}{\textbf{Benchmark}} & \multirow{2.5}{*}{\textbf{Input}} & \multicolumn{2}{c}{\textbf{Persp.}} & \multicolumn{4}{c}{\textbf{Interactions}} & \multicolumn{5}{c}{\textbf{Dimensions}} & \multicolumn{2}{c}{\textbf{Scale}} \\
\cmidrule(lr){3-4} \cmidrule(lr){5-8} \cmidrule(lr){9-13} \cmidrule(l){14-15}
& & \textbf{FPP} & \textbf{TPP} & \textbf{Navi} & \textbf{SA} & \textbf{EE} & \textbf{PS} & \textbf{Qual} & \textbf{Adh} & \textbf{Inter} & \textbf{Cons} & \textbf{Phys} & \textbf{Cases} & \textbf{Turns} \\
\midrule
VBench~\cite{huang2024vbench}                     & \iOn{\textsf{\textbf{T}}}\,\iOff{\faVideo}\,\iOff{\faGamepad}     & -- & -- & \xmark & \xmark & \xmark & \xmark & \cmark & \xmark & \xmark & \xmark & \xmark & 946   & 946   \\
WorldScore~\cite{duan2024worldscore}              & \iOn{\textsf{\textbf{T}}}\,\iOn{\faVideo}\,\iOff{\faGamepad}     & \cmark & \xmark & \cmark & \xmark & \cmark & \xmark & \cmark & \cmark & \xmark & \xmark & \xmark & 3,000 & 3,000 \\
WorldModelBench~\cite{wang2024worldmodelbench}    & \iOn{\textsf{\textbf{T}}}\,\iOff{\faVideo}\,\iOff{\faGamepad}     & \cmark & -- & \cmark & \cmark & \xmark & \xmark & \xmark & \cmark & \cmark & \xmark & \cmark & 350   & 350   \\
MIND~\cite{ye2026mind}                            & \iOff{\textsf{\textbf{T}}}\,\iOn{\faVideo}\,\iOff{\faGamepad}     & \cmark & -- & \cmark & \xmark & \xmark & \xmark & \cmark & \xmark & \cmark & \cmark & \xmark & 250   & --    \\
WorldArena~\cite{shang2026worldarena}             & \iOn{\textsf{\textbf{T}}}\,\iOn{\faVideo}\,\iOff{\faGamepad}      & \cmark & \xmark & \cmark & \cmark & \xmark & \xmark & \cmark & \xmark & \cmark & \cmark & \cmark & 500   & 500   \\
Omni-WorldBench~\cite{wu2026omniworldbench}       & \iOn{\textsf{\textbf{T}}}\,\iOn{\faVideo}\,\iOff{\faGamepad}      & \cmark & -- & \cmark & \cmark & \cmark & \xmark & \cmark & \xmark & \cmark & \cmark & \cmark & 1,068 & 1,068 \\
WorldLens~\cite{worldlens2025}                    & \iOff{\textsf{\textbf{T}}}\,\iOn{\faVideo}\,\iOff{\faGamepad}     & \cmark & \xmark & \cmark & \xmark & \xmark & \xmark & \cmark & \cmark & \cmark & \cmark & \cmark & 26k   & --   \\
WorldMark~\cite{xu2026worldmark}                  & \iOn{\textsf{\textbf{T}}}\,\iOn{\faVideo}\,\iOn{\faGamepad}     & \cmark & \cmark & \cmark & \xmark & \xmark & \xmark & \cmark & \xmark & \cmark & \cmark & \xmark & 500   & --    \\
\midrule
\textbf{\benchmark (Ours)}                        & \iOn{\textsf{\textbf{T}}}\,\iOn{\faVideo}\,\iOn{\faGamepad}       & \cmark & \cmark & \cmark & \cmark & \cmark & \cmark & \cmark & \cmark & \cmark & \cmark & \cmark & \numvideo  & \numturn \\
\bottomrule
\end{tabular}
\vspace{-1.6mm}
\end{table}

\section{\benchmark Dataset}
\label{sec:dataset}

\begin{figure}[t]
\captionsetup{skip=1mm}
\centering
\includegraphics[width=0.95\linewidth]  {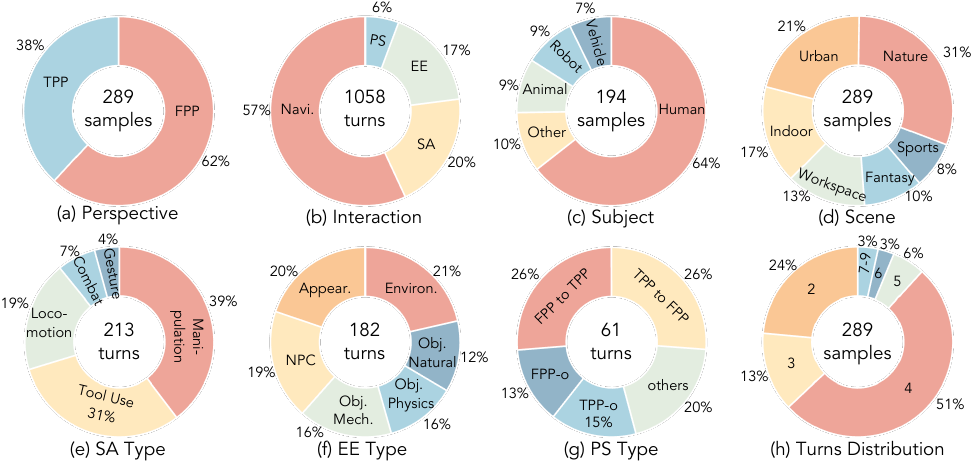}%
\caption[Dataset composition across eight axes]{Dataset composition of \benchmark across eight axes. We discuss these in \cref{sec:statistics}.}
\vspace{-2mm}
\label{fig:dataset_distribution}
\end{figure}

An interactive world model~\cite{ha2018world,lecun2022path} acts as a conditional generator that predicts the next observation $o_{t+1}$ given the historical observation $o_{\le t} = (o_0, \dots, o_t)$ and the action $a_{\le t} = (a_0, \dots, a_t)$:
\begin{equation}
    o_{t+1} \sim f_\theta(o_{t+1}|o_{\le t}, a_{\le t}).
\label{eq:world_model}
\end{equation}
To systematically evaluate this process, every case in \benchmark decomposes the inputs into two components: a \textbf{World Setting} $\mathcal{W}$ that defines the initial world state $o_0$, and an \textbf{Interaction sequence} $\mathcal{I} = (a_0, a_1, \dots, a_{T-1})$ that specifies the user control signals spanning $T$ consecutive turns.

\subsection{Dataset Construction}
\label{sec:ontology}

\myparagraph{World Settings.}
A world setting $\mathcal{W}$ is defined by four attributes:
\textbf{1)~Scene}, the environment type, spatial layout, and inherent dynamics, including both elements visible in the initial frame (\eg, terrain, buildings) and offscreen elements expected to appear during interaction (\eg, a river behind the camera);
\textbf{2)~Style}, the rendering appearance, such as realistic, cartoon, anime, cinematic, CG, or oil painting;
\textbf{3)~Perspective}, either first- or third-person;
and \textbf{4)~Subject}, the primary entity in the scene, such as a human, animal, vehicle, or robot. The subject attribute applies to all third-person cases and first-person cases where the viewer holds or controls a visible entity (\eg, a tool or an ego robot arm); environment-only first-person scenes have no associated subject.
These four attributes are composed into an \textbf{environment prompt} (Scene + Style) and a \textbf{subject prompt} (Perspective + Subject), which together with an initial frame form the input to each evaluated model.
Initial frames are generated by Nano Banana~2~\cite{google2025nanobanana2} and GPT-Image-1.5~\cite{openai2025gptimage}, supplemented by web-collected and manually captured images. All initial frames undergo manual verification for quality control.

\myparagraph{Interactions.}
Each case specifies a multi-turn interaction sequence drawn from four complementary types that can be freely composed within a single case, as shown in \cref{fig:teaser} (top).
\textbf{1)~Navigation} governs camera or ego-agent motion through four \emph{translational} controls \texttt{W}/\texttt{S}/\texttt{A}/\texttt{D} and four \emph{rotational} controls $\leftarrow$/$\rightarrow$/$\uparrow$/$\downarrow$, composable into compound actions such as \texttt{W}+$\leftarrow$. The same key drives the camera in first-person mode and the subject in third-person mode. Trajectories span six path topologies for motion diversity (\cref{app:nav_coverage}).
\textbf{2)~Subject Action} covers actions performed by the primary subject, including manipulation, locomotion, tool use, combat, and gestural interaction.
\textbf{3)~Event Editing} covers externally imposed changes to the environment, such as weather transitions, time-of-day shifts, object appearances, \etc
\textbf{4)~Perspective Switching} covers transitions between first- and third-person views, including same-subject switches, multi-subject switches, and scope mode transitions.

\myparagraph{Case Construction.}
Construction follows a setting-first principle: annotators design a world setting and then derive interaction sequences that are physically executable and semantically coherent within it (\eg, manipulation in a kitchen, weather transitions outdoors, and reasonable navigation trajectories). Multi-turn sequences respect causal ordering. We apply stratified sampling across scene, style, perspective, subject, and interaction type to ensure diverse coverage, with all selected cases undergoing manual review for prompt-frame consistency and inter-turn coherence.

\subsection{Dataset Statistics}
\label{sec:statistics}

\benchmark comprises \numvideo cases spanning \numturn interaction turns, with first-person cases at $62\%$ and third-person at $38\%$, see \cref{fig:dataset_distribution}~\!(a). Navigation is the most prevalent interaction ($57\%$), followed by subject action ($20\%$), event editing ($17\%$), and perspective switching ($6\%$), as shown in \cref{fig:dataset_distribution}~\!(b).

\myparagraph{Scene, Subject, and Style Diversity.} Scenes span six categories, led by nature ($31\%$) and urban environments ($21\%$), with indoor ($17\%$), works ($13\%$), fantasy ($10\%$), and sports ($8\%$) settings completing the spectrum, see \cref{fig:dataset_distribution}~\!(d). Across the $194$ cases with an explicit subject, humans dominate ($64\%$), followed by animals ($9\%$), robots ($9\%$), vehicles ($7\%$), and $10\%$ miscellaneous objects (\cref{fig:dataset_distribution}~~(c)). Photorealistic rendering covers $52\%$ of the cases, while the remaining $48\%$ span styles including anime, cartoon, CG, oil painting, ink wash, pencil sketch, and flat or abstract styles.

\myparagraph{Interaction Sub-type Taxonomy.}
As shown in \cref{fig:dataset_distribution}~~(e)(g), subject action is categorized into five sub-types, dominated by manipulation ($39\%$) and tool use ($31\%$), with locomotion, combat, and gestures comprising the remainder. Event editing covers six relatively balanced sub-types, including environment changes ($21\%$), appearance-state changes ($20\%$), NPC motion ($19\%$), and three types of object-state transitions involving mechanical, physical, and natural phenomena. Perspective switching consists of $61$ turns, including cross-perspective switches with each direction at $26\%$, intra-perspective switches(denoted by ``-o'' in the figure) accounting for $28\%$ in total, and other switches such as TPP-to-scope.

\myparagraph{Multi-turn Interaction Depth.}
Each case spans 2-9 interaction turns with an average of $3.7$, as shown in \cref{fig:dataset_distribution}~~(h). Four-turn cases are the most common ($51\%$) and mostly correspond to navigation trajectories, whereas the $12\%$ of longer 5--9-turn cases typically interleave subject action with event editing. Such multi-turn structure probes temporal consistency and long-horizon coherence, which single-turn benchmarks cannot assess. Further breakdowns of navigation coverage, evaluation activation, and lexical diversity are provided in \cref{app:dataset}.

\section{\benchmark Evaluation Suite}
\label{sec:evaluation_suite}

\benchmark decomposes evaluation into five complementary dimensions, each targeting a distinct aspect of world model fidelity.
In total, the evaluation suite comprises \numsubmetric fine-grained sub-metrics across these \textbf{five dimensions}. Detailed descriptions of each metric are provided in \cref{app:metric_details}.
All sub-metric scores are linearly rescaled to $[0, 100]$ for direct comparability across dimensions, with higher values indicating better performance.

\myparagraph{$\bullet$~\tagbox{dimVQ}{Video Quality.}}
\label{sec:d1_video_quality}
Video quality measures the perceptual quality of the generated video irrespective of the conditioning signal.
We adopt five sub-metrics from VBench~\cite{huang2024vbench}:
\tagbox{dimVQ}{V.1}\,Aesthetic Quality,
\tagbox{dimVQ}{V.2}\,Imaging Quality,
\tagbox{dimVQ}{V.3}\,Temporal Flickering,
\tagbox{dimVQ}{V.4}\,Dynamic Degree and
\tagbox{dimVQ}{V.5}\,Motion Smoothness,
plus \tagbox{dimVQ}{V.6}\,HPSv3-Norm~\cite{hpsv3}, a percentile-normalized human-preference reward score.

\myparagraph{$\bullet$~\tagbox{dimSA}{Setting Adherence.}}
Setting adherence measures whether the generated video faithfully reflects the specified world setting $\mathcal{W}$.
We evaluate two sub-metrics below:

\myparagraph{\tagbox{dimSA}{S.1}\,Scene Adherence.}
We decompose the environment prompt into an \emph{initially visible} part (\eg terrain, buildings in the initial frame) and an \emph{offscreen} part (\eg a river behind the camera) expected to appear later.
A VLM scores both components: whether initially visible elements remain consistent throughout, and whether described but offscreen elements eventually appear.

\myparagraph{\tagbox{dimSA}{S.2}\,Subject Adherence.}
We decompose the subject prompt into an \emph{appearance} part (\eg fur color, clothing) and a \emph{motion} part (\eg gait, agility).
A VLM\footnote{Unless otherwise noted, all VLM scoring in this paper uses \texttt{doubao-seed-2-0-lite-260215}.} scores whether the subject's visual attributes match the described appearance, and whether its movement style matches declared motion priors.

\myparagraph{$\bullet$~\tagbox{dimIA}{Interaction Adherence.}}
Interaction adherence evaluates whether the model correctly executes the requested interaction $\mathcal{I}$.
Navigation is assessed using geometric pose estimation, while the remaining three types are evaluated through structured VLM scoring with binary criteria per turn.

\myparagraph{\tagbox{dimIA}{I.1}\,Navigation Score.}
We estimate per-frame camera poses with MegaSaM~\cite{li2024megasam} and compare against a synthetic ground-truth trajectory built from the action sequence.
The GT encodes perspective-dependent semantics: first-person rotations produce heading changes, while third-person rotations produce orbital motion around the subject.
After alignment and arc-length resampling, we compute normalized Absolute Trajectory Error (nATE) as the accuracy term, and cross-turn trajectory consistency for repeated actions. The final score averages both.

\myparagraph{\tagbox{dimIA}{I.2}\,Event Editing and \tagbox{dimIA}{I.3}\,Subject Action Adherence.}
We use a unified turn-level VLM protocol for these two interaction types. For each turn, the VLM inspects the corresponding video segment with five binary checks derived from the action specification: change detection, event occurrence, completion, detail accuracy, and anomaly absence. Each satisfied check contributes one point, giving a $[0,5]$ grade that is averaged across turns per case then scaled to a 100-point score. The complete prompt templates and scoring details are provided in Appendix~\ref{app:event_edit_detail} and~\ref{app:subject_action_detail}.

\myparagraph{\tagbox{dimIA}{I.4}\,Perspective Switching Adherence.}
We score perspective switching with a stricter categorical protocol. The early and late frames of each relevant turn are jointly checked against three binary criteria: transition visibility, target-type consistency, and structural compliance of the new viewpoint. A turn is counted as successful only when all three hold, and the case score is the fraction percentage of successful turns. Details are provided in Appendix~\ref{app:persp_switch_detail}.

\myparagraph{$\bullet$~\tagbox{dimCO}{Consistency.}}
Consistency measures whether scene geometry, object appearance, and perspective anchoring remain stable as the camera moves and interactions accumulate.

\myparagraph{\tagbox{dimCO}{C.1}\,Spatial Consistency and \tagbox{dimCO}{C.2}\,Gated Spatial Consistency.}
For roundtrip trajectories~\cite{ye2026mind} (\eg $\text{left}\!\times\!2 \rightarrow \text{right}\!\times\!2$), we use MegaSaM-estimated camera poses to locate the return frame best matching the initial viewpoint, then compute DreamSim~\cite{fu2023dreamsim} perceptual similarity with the first frame.
The \emph{gated} variant additionally samples intermediate frames and computes their minimum similarity to the first frame, suppressing the score when the video barely moves.

\myparagraph{\tagbox{dimCO}{C.3}\,Segment Continuity.}
We use TransNetV2~\cite{soucek2020transnet} to detect unexpected hard cuts within each generated video.
The model-level score is the fraction of videos without any detected scene cuts.

\myparagraph{\tagbox{dimCO}{C.4}\,Perspective Consistency.}
We track the subject with SAM2~\cite{ravi2024sam2} and measure how stable its centroid remains across frames, weighted by the fraction of frames in which the subject is visible.

\myparagraph{\tagbox{dimCO}{C.5}\,Geometric Consistency and \tagbox{dimCO}{C.6}\,Photometric Consistency.}
We use Depth Anything 3~\cite{lin2025depthanything3} to estimate per-frame depth and camera poses, then reproject pixels across views.
Geometric consistency measures 3D structural coherence via reprojection displacement~\cite{an2026vggrpo}, while photometric consistency measures appearance stability via pixel-level PSNR between reprojected frame pairs~\cite{du2026videogpa}.

\myparagraph{\tagbox{dimCO}{C.7}\,Subject Consistency.}
We apply SAM2 masks to isolate the subject and retain only frames where it is visible, then average two complementary signals: DINOv2~\cite{oquab2024dinov2} adjacent-frame cosine similarity for local continuity, and CLIP first-frame anchored similarity for global drift detection.

\myparagraph{\tagbox{dimCO}{C.8}\,Background Consistency.}
Following VBench~\cite{huang2024vbench}, we measure the mean pairwise CLIP cosine similarity between consecutive frames, capturing temporal stability of the background appearance.

\vspace{1mm}
\myparagraph{$\bullet$~\tagbox{dimPH}{Physical.}}
Physical dimension assesses whether the generated world obeys declared physical rules, covering both high-level causal fidelity and low-level visual plausibility.

\myparagraph{\tagbox{dimPH}{P.1}\,Causal Fidelity.}
Causal fidelity is evaluated with a two-stage VLM protocol using three-point grading. Frames are uniformly sampled across all turns and fed to the VLM as a single sequence for holistic assessment.
\textbf{Stage~1} assesses \emph{global plausibility}, focusing on \emph{rendering-physics violations} such as motion continuity, object permanence, and character physics, as well as \emph{causal inconsistencies} where effects occur without causes, causes fail to produce effects, or unrelated objects unexpectedly appear. Instructed actions are excluded.
\textbf{Stage~2} assesses \emph{context-conditioned accuracy} over seven physics sub-dimensions: fluid and smoke, collision, surface tracks, deformation, wind, reflection, and human motion. For each case, a separate VLM assistant first identifies applicable sub-dimensions from scene metadata, action descriptions, and the initial frame under strict criteria. For example, \emph{fluid} is selected only when visible liquid or smoke is present. The selected sub-dimensions are manually verified and adjusted, kept fixed across all models, and used as the only sub-dimensions scored by the VLM evaluator.
The final per-case score averages the Stage~1 score and the mean Stage~2 score, then scales to 0--100. Further details on the dimension split and prompts are provided in \ref{app:causal_fidelity_detail}.

\myparagraph{\tagbox{dimPH}{P.2}\,Visual Plausibility.}
We fine-tune a pretrained Qwen3-VL-30B-A3B~\cite{Qwen3-VL} on in-house expert-annotated data to automatically score low-level physical artifacts such as geometric distortion, object penetration, and unnatural deformation, producing a continuous $[1, 5]$ score, normalized to $[0, 100]$.
This complements causal fidelity, which targets high-level rule compliance, by catching pervasive visual implausibilities that do not require case-specific questions. Details are in ~\cref{app:visual_plausibility_detail}.

\section{Experiments}
\label{sec:experiments}

\begin{table*}[t]
\caption[Main results on \benchmark]{Main results on \benchmark navigation split (\numnavi navigation cases). Columns from left to right: text-driven, camera-controlled, and action-conditioned models. Event editing, subject action, and perspective switching are from the non-navigation split (text-driven only). All scores $\in [0,100]$, higher is better. The full results are displayed in Appendix~
\ref{app:full_split_text}.}
\vspace{-2mm}
\label{tab:full_results_transposed}
\centering
\renewcommand{\arraystretch}{1.3}
\setlength{\tabcolsep}{2.33pt}
\scriptsize
\renewcommand{\icon}[1]{\raisebox{-0.15em}{\makebox[1.2em][c]{\includegraphics[height=1.1em,width=1.1em,keepaspectratio]{figures/icon/#1}}}\hspace{0.1em}}
\newcommand{\iconbg}[2]{\cellcolor{#1}\rule{0pt}{1.6em}\raisebox{-0.05em}{\makebox[1.2em][c]{\includegraphics[height=1.1em,width=1.1em,keepaspectratio]{figures/icon/#2}}}}
\newcommand{\g}[1]{\textcolor{gray}{#1}}
\begin{tabular}{@{}cl| *{9}{c} c |*{5}{c} c |*{6}{c} c @{}}
\thickhline
\textbf{} & \textbf{Metrics}
  & \rotatebox{90}{\scriptsize\scriptsize\textbf{Seedance 1.5}}
  & \rotatebox{90}{\scriptsize\textbf{Wan 2.7}}
  & \rotatebox{90}{\scriptsize\textbf{Kling 3.0}}
  & \rotatebox{90}{\scriptsize\textbf{YUME 1.5}}
  & \rotatebox{90}{\scriptsize\textbf{HY-Video 1.5}}
  & \rotatebox{90}{\scriptsize\textbf{LTX 2.3}}
  & \rotatebox{90}{\scriptsize\textbf{LongCat-Video}}
  & \rotatebox{90}{\scriptsize\scriptsize\textbf{Kairos 3.0}}
  & \rotatebox{90}{\scriptsize\textbf{Cosmos 2.5}}
  & \rotatebox{90}{\scriptsize\textit{Average}}
  & \rotatebox{90}{\scriptsize\textbf{LingBot-World}}
  & \rotatebox{90}{\scriptsize\scriptsize\textbf{HY-World 1.5}}
  & \rotatebox{90}{\scriptsize\textbf{Fantasy-World}}
  & \rotatebox{90}{\scriptsize\textbf{InSpatio-World}}
  & \rotatebox{90}{\scriptsize\textbf{Astra}}
  & \rotatebox{90}{\scriptsize\textit{Average}}
  & \rotatebox{90}{\scriptsize\textbf{Happy Oyster}}
  & \rotatebox{90}{\scriptsize\textbf{Matrix-Game 3.0 \ }}
  & \rotatebox{90}{\scriptsize\textbf{Genie 3}}
  & \rotatebox{90}{\scriptsize\textbf{Matrix-Game 2.0}}
  & \rotatebox{90}{\scriptsize\textbf{HY-GameCraft}}
  & \rotatebox{90}{\scriptsize\textbf{Infinite-World}}
  & \rotatebox{90}{\scriptsize\textit{Average}} \\
\noalign{\vskip -1pt}
\textbf{} & \textbf{}
  & \iconbg{blue!10}{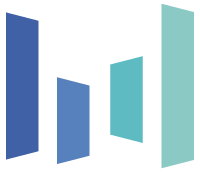}
  & \iconbg{blue!10}{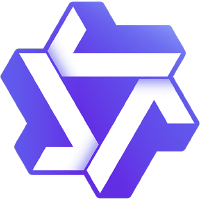}
  & \iconbg{blue!10}{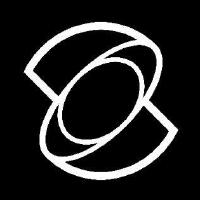}
  & \iconbg{blue!10}{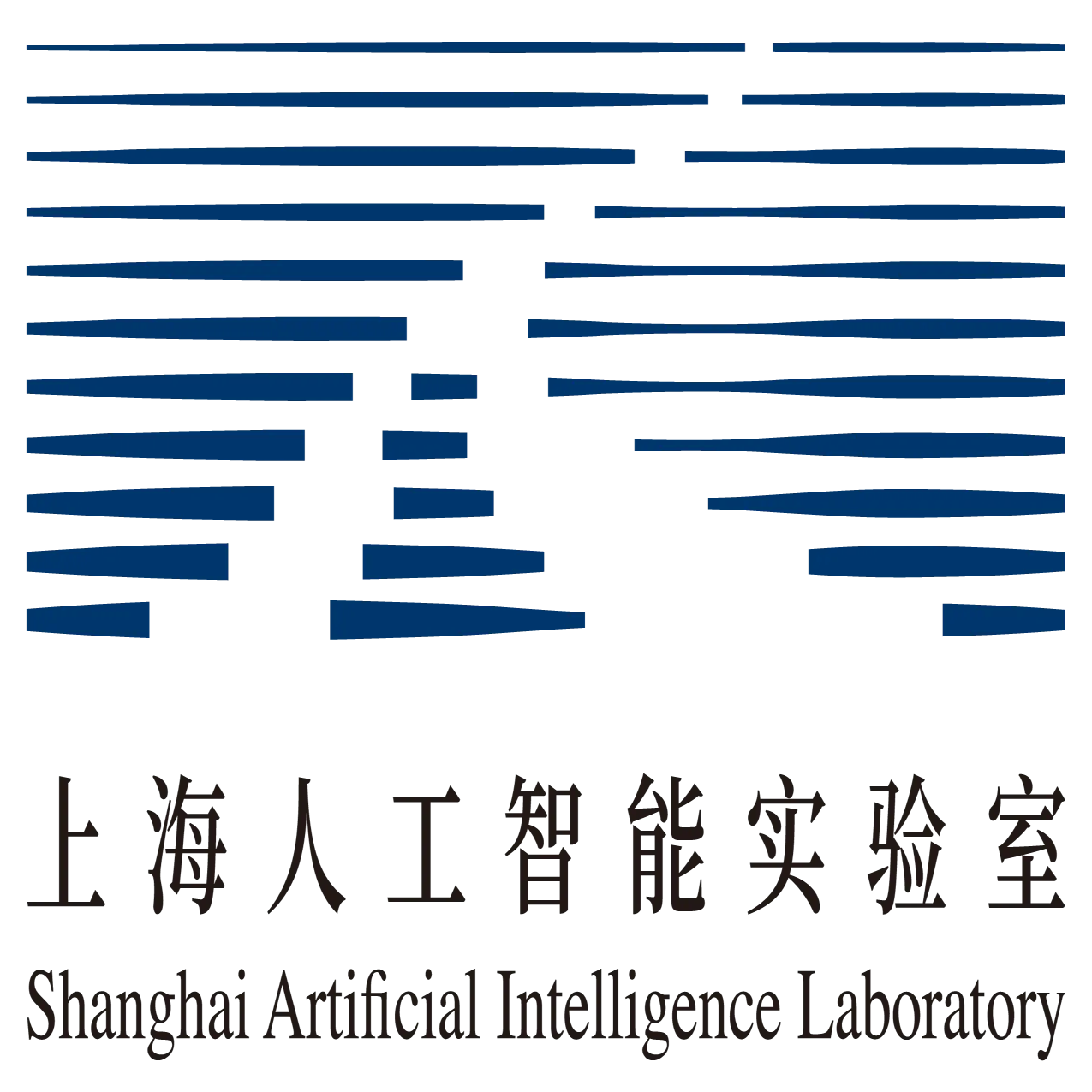}
  & \iconbg{blue!10}{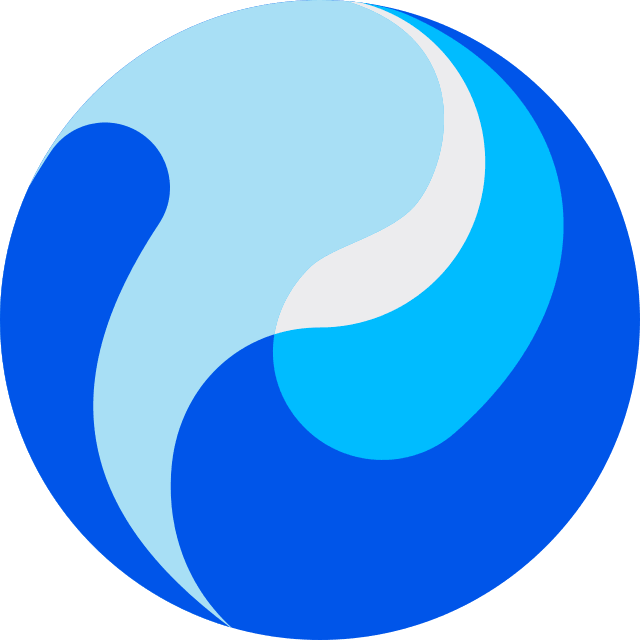}
  & \iconbg{blue!10}{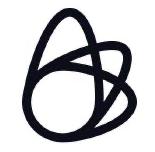}
  & \iconbg{blue!10}{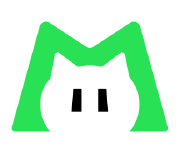}
  & \iconbg{blue!10}{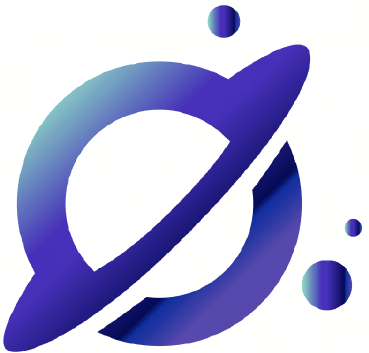}
  & \iconbg{blue!10}{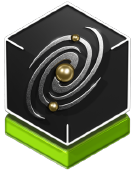}
  &
  & \iconbg{teal!10}{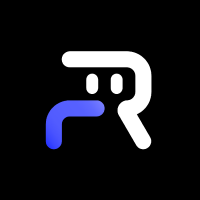}
  & \iconbg{teal!10}{hunyuan.png}
  & \iconbg{teal!10}{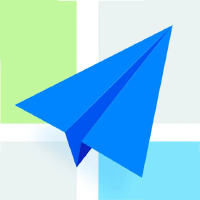}
  & \iconbg{teal!10}{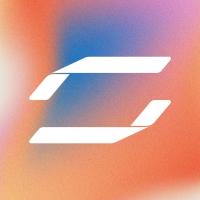}
  & \iconbg{teal!10}{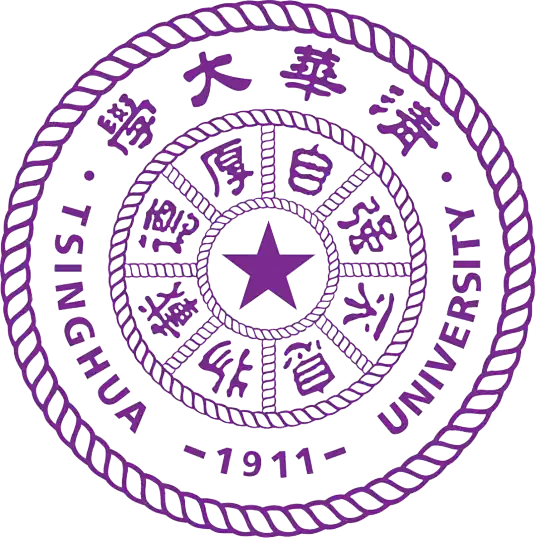}
  &
  & \iconbg{orange!10}{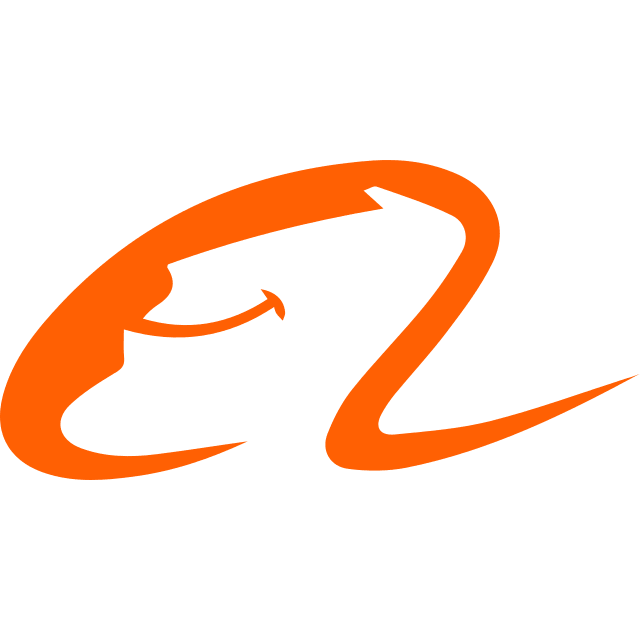}
  & \iconbg{orange!10}{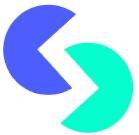}
  & \iconbg{orange!10}{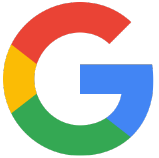}
  & \iconbg{orange!10}{skywork.jpeg}
  & \iconbg{orange!10}{hunyuan.png}
  & \iconbg{orange!10}{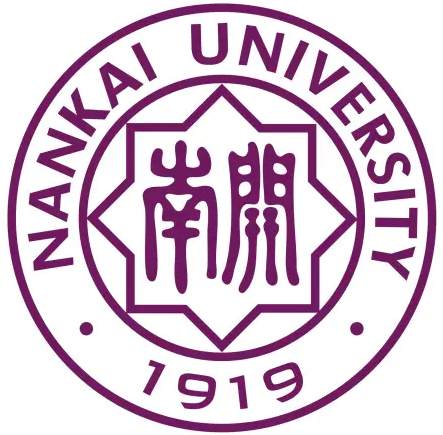}
  & \\
\hline
\multirow{7}{*}{\rotatebox{90}{\textbf{Video Quality}}}
  & Aesthetic   & 61.0 & 61.4 & 63.0 & 58.7 & \underline{63.4} & 57.9 & \textbf{66.5} & 59.9 & 61.8 & \g{61.5}& \textbf{66.9} & 60.1 & 63.0 & \underline{64.4} & 48.6 & \g{60.6}& \underline{56.6} & 46.4 & 51.6 & 54.0 & 52.6 & \textbf{58.7} & \g{53.3} \\
  & Imaging     & \underline{69.3} & 68.0 & 68.1 & 63.3 & 67.4 & 61.0 & \textbf{69.6} & 62.7 & 66.9 & \g{66.3}& \textbf{67.9} & 65.4 & 62.8 & \underline{67.6} & 52.5 & \g{63.2}& 63.9 & \textbf{70.0} & 59.3 & 60.3 & 58.7 & \underline{66.1} & \g{63.1} \\
  & Flickering  & 92.4 & 92.2 & 93.2 & 93.0 & 94.2 & 93.2 & \underline{94.8} & \textbf{95.4} & \underline{94.8} & \g{93.7}& 94.1 & 93.5 & \underline{95.8} & \textbf{96.0} & \textbf{96.0} & \g{95.1}& 94.0 & 86.3 & \textbf{95.0} & \underline{94.6} & 93.7 & 94.1 & \g{93.0} \\
  & Dynamic     & \underline{99.4} & \textbf{100.0} & 97.5 & 96.8 & 73.9 & 98.1 & 45.9 & 70.1 & 49.0 & \g{81.2}& 66.2 & \textbf{91.1} & 49.0 & 26.1 & \underline{79.6} & \g{62.4}& 94.2 & \textbf{97.5} & 92.4 & 94.9 & \underline{96.8} & 82.8 & \g{93.1} \\
  & Smoothness  & 97.5 & 96.3 & 97.6 & 97.0 & \textbf{98.7} & 96.4 & 97.9 & 97.5 & \underline{98.2} & \g{97.5}& 96.9 & \underline{98.1} & 97.9 & \textbf{98.8} & 97.7 & \g{97.9}& 97.0 & 95.4 & 97.8 & \textbf{98.2} & 97.6 & \underline{98.0} & \g{97.3} \\
  & HPSv3-Norm  & \underline{73.0} & 71.1 & 69.1 & 57.0 & 68.0 & 56.1 & \textbf{77.6} & 58.5 & 66.5 & \g{66.3}& \textbf{81.4} & 60.5 & 65.8 & \underline{76.1} & 28.0 & \g{62.4}& \underline{58.3} & 57.1 & 55.2 & 41.0 & 38.3 & \textbf{62.3} & \g{52.0} \\
  & \textbf{Average} & \textbf{82.1} & \underline{81.5} & 81.4 & 77.6 & 77.6 & 77.1 & 75.4 & 74.0 & 72.9 & \g{77.7}& \textbf{78.9} & \underline{78.1} & 72.4 & 71.5 & 67.1 & \g{73.6}& \textbf{77.3} & 75.5 & 75.2 & 73.8 & 73.0 & \underline{77.0} & \g{75.3} \\
\hline
\multirow{3}{*}{\rotatebox{90}{\textbf{Setting}}}
  & Scene       & 71.6 & \underline{88.3} & \textbf{89.0} & 53.1 & 77.5 & 81.3 & 53.1 & 52.2 & 72.4 & \g{70.9}& 51.6 & \textbf{53.5} & \underline{52.4} & 51.7 & 43.4 & \g{50.5}& \underline{57.4} & 48.9 & \textbf{61.1} & 49.4 & 50.6 & 54.0 & \g{53.6} \\
  & Subject     & \underline{94.2} & \textbf{94.6} & 92.9 & 91.7 & 93.6 & 89.2 & 91.5 & 88.5 & \underline{94.2} & \g{92.3}& \textbf{93.6} & 90.8 & 90.1 & \underline{91.1} & 75.9 & \g{88.3}& \textbf{91.1} & 78.4 & 83.8 & \underline{84.9} & 82.5 & 84.5 & \g{84.2} \\
  & \textbf{Average} & 82.9 & \textbf{91.4} & \underline{91.0} & 72.4 & 85.5 & 85.2 & 72.3 & 70.3 & 83.3 & \g{81.6}& \textbf{72.6} & \underline{72.2} & 71.2 & 71.4 & 59.7 & \g{69.4}& \textbf{74.2} & 63.7 & \underline{72.5} & 67.2 & 66.5 & 69.2 & \g{68.9} \\
\hline
\multirow{5}{*}{\rotatebox{90}{\textbf{Interaction}}}
  & Navigation  & 68.0 & 66.0 & 70.3 & \textbf{72.0} & \underline{71.8} & 67.6 & 63.1 & 65.1 & 64.1 & \g{67.6}& \underline{79.8} & \textbf{87.5} & 72.1 & 72.8 & 67.7 & \g{76.0}& \textbf{85.1} & \underline{83.5} & 73.3 & 80.6 & 67.8 & 75.9 & \g{77.7} \\
  & Event Editing  & 80.4 & \textbf{84.0} & \underline{81.4} & 57.8 & 63.8 & 53.0 & 50.4 & 46.8 & 48.2 & \g{62.9}& -- & -- & -- & -- & -- & \g{--}& -- & -- & -- & -- & -- & -- & \g{--} \\
  & Subject Action & 80.0 & \underline{83.4} & \textbf{85.6} & 47.0 & 55.6 & 51.8 & 48.4 & 41.4 & 41.6 & \g{59.4}& -- & -- & -- & -- & -- & \g{--}& -- & -- & -- & -- & -- & -- & \g{--} \\
  & Persp. Switching & \underline{45.0} & \textbf{55.0} & \textbf{55.0} & 16.7 & 27.6 & 25.0 & 18.3 & 13.3 & 20.0 & \g{30.7}& -- & -- & -- & -- & -- & \g{--}& -- & -- & -- & -- & -- & -- & \g{--} \\
  & \textbf{Average} & 68.3 & \underline{72.1} & \textbf{73.1} & 48.4 & 54.7 & 49.3 & 45.1 & 41.6 & 43.5 & \g{55.1}& -- & -- & -- & -- & -- & \g{--}& -- & -- & -- & -- & -- & -- & \g{--} \\
\hline
\multirow{9}{*}{\rotatebox{90}{\textbf{Consistency}}}
  & Background  & 89.6 & 89.4 & \underline{92.3} & 90.3 & 92.1 & 88.3 & \textbf{95.1} & 91.1 & \underline{92.3} & \g{91.2}& \textbf{96.9} & 92.7 & 94.2 & \underline{95.0} & 85.3 & \g{92.8}& \textbf{91.4} & 85.7 & \underline{90.7} & 86.9 & 86.5 & 88.8 & \g{88.3} \\
  & Spatial     & 72.7 & 71.0 & 75.2 & 71.5 & \underline{79.2} & 70.2 & \textbf{83.3} & 76.8 & 78.1 & \g{75.3}& \underline{92.7} & 90.6 & 80.6 & \textbf{93.8} & 64.7 & \g{84.5}& 77.7 & \textbf{81.0} & \underline{79.9} & 64.5 & 60.5 & 74.9 & \g{73.1} \\
  & Gated Spatial & 72.4 & 71.0 & \textbf{75.1} & 71.4 & \textbf{75.1} & 70.2 & 66.2 & 62.0 & \underline{74.3} & \g{70.9}& \underline{67.1} & \textbf{84.9} & 64.2 & 66.5 & 63.3 & \g{69.2}& 75.8 & \textbf{80.4} & \underline{78.4} & 64.5 & 60.5 & 74.4 & \g{72.3} \\
  & Segment     & \underline{96.2} & 92.4 & 93.0 & \textbf{99.4} & \textbf{99.4} & 75.8 & \textbf{99.4} & 94.3 & 94.3 & \g{93.8}& \underline{99.4} & \textbf{100.0} & \textbf{100.0} & \textbf{100.0} & 86.6 & \g{97.2}& 96.2 & 89.8 & 93.6 & 21.0 & \underline{99.4} & \textbf{100.0} & \g{83.3} \\
  & Perspective & 70.5 & 78.2 & 76.8 & 48.0 & \textbf{86.6} & 69.8 & 81.5 & 76.3 & \underline{84.3} & \g{74.7}& \textbf{90.9} & 62.5 & \underline{79.8} & 72.5 & 30.0 & \g{67.1}& \textbf{75.0} & 13.3 & \underline{54.5} & 29.2 & 17.9 & 33.8 & \g{37.3} \\
  & Subject     & 90.1 & 90.7 & 88.5 & 88.8 & 91.6 & 87.2 & \textbf{93.4} & 90.8 & \underline{92.3} & \g{90.4}& \underline{93.5} & 89.1 & 92.5 & \textbf{94.4} & 83.5 & \g{90.6}& \textbf{91.5} & 83.0 & \underline{90.4} & 87.2 & 82.6 & 88.4 & \g{87.2} \\
  & Geometric   & 82.4 & 83.7 & 88.9 & 88.0 & \underline{94.6} & 76.9 & \textbf{95.4} & 89.0 & \underline{94.6} & \g{88.2}& \underline{95.4} & 92.0 & 95.3 & \textbf{97.3} & 85.6 & \g{93.1}& 87.2 & 87.6 & \underline{88.6} & 86.1 & 88.3 & \textbf{94.3} & \g{88.7} \\
  & Photometric & 76.8 & 76.4 & 79.9 & \textbf{83.3} & 80.3 & 79.2 & \underline{82.2} & 80.8 & 81.6 & \g{80.1}& 83.3 & 83.1 & 84.8 & \underline{87.4} & \textbf{87.5} & \g{85.2}& 79.8 & 75.3 & 84.5 & 81.3 & \underline{85.0} & \textbf{85.1} & \g{81.8} \\
  & \textbf{Average} & 81.3 & 81.6 & 83.7 & 80.1 & \textbf{87.4} & 77.2 & \underline{87.1} & 82.6 & 86.5 & \g{83.1}& \textbf{89.9} & 86.9 & 86.4 & \underline{88.4} & 73.3 & \g{85.0}& \textbf{84.3} & 74.5 & \underline{82.6} & 65.1 & 72.6 & 79.9 & \g{76.5} \\
\hline
\multirow{3}{*}{\rotatebox{90}{\textbf{Physical}}}
  & Causal Fidelity      & 76.0 & \textbf{83.3} & \underline{78.0} & 72.7 & 75.0 & 74.0 & 76.0 & 62.7 & 74.7 & \g{74.7}& \textbf{77.7} & \underline{74.0} & \underline{74.0} & 67.3 & 48.3 & \g{68.3}& \underline{69.3} & 64.7 & \textbf{71.7} & 59.3 & 68.3 & 67.0 & \g{66.7} \\
  & Visual Plausibility  & \underline{60.7} & 60.3 & \underline{60.7} & 57.7 & 59.7 & 55.7 & \textbf{61.8} & 58.0 & 60.1 & \g{59.4}& \textbf{64.8} & 58.6 & 59.7 & \underline{63.1} & 54.6 & \g{60.2}& \underline{57.6} & 54.0 & \textbf{59.7} & 55.0 & 56.5 & 57.2 & \g{56.7} \\
  & \textbf{Average} & 68.3 & \textbf{71.8} & \underline{69.3} & 65.2 & 67.3 & 64.8 & 68.9 & 60.4 & 67.4 & \g{67.0}& \textbf{71.2} & 66.3 & \underline{66.8} & 65.2 & 51.5 & \g{64.2}& \underline{63.5} & 59.4 & \textbf{65.7} & 57.1 & 62.4 & 62.1 & \g{61.7} \\
\thickhline
\end{tabular}
\end{table*}

\subsection{Evaluated Models and Protocol}
\label{sec:eval_protocol}

As shown in \cref{tab:full_results_transposed}, we evaluate \nummodel models spanning three paradigms: (1)~\textbf{text-driven} (9 models, \eg Seedance~1.5~\cite{gao2025seedance15}, Kling~3.0~\cite{kuaishou2025kling3}, Wan~2.7~\cite{wan2025wan27}), which accept all four interaction types and are evaluated on the full \numvideo-case test set via iterative last-frame forwarding,
(2)~\textbf{camera-controlled} (5 models, \eg HY-World~1.5~\cite{sun2025hyworldplay}, LingBot-World~\cite{gao2026lingbot}), and
(3)~\textbf{action-conditioned} (6 models, \eg Genie~3~\cite{ball2025genie3}, Matrix-Game~3.0~\cite{he2026matrixgame3}), both restricted to the navigation subset of \numnavi cases.
Each navigation action is defined as a canonical camera movement and is mapped to natural-language prompts, relative 6-DoF poses, or discrete keyboard commands depending on the paradigm, allowing fair cross-paradigm comparison.
Semantic interactions (event editing, subject action, perspective switching) are text-only and thus restricted to text-driven models. More details are in Appendix~\ref{app:model_details}.

\subsection{Per-Dimension Results}
\label{sec:per_dimension}

\cref{tab:full_results_transposed} presents a unified view across all five dimensions. We analyze each dimension below:

\myparagraph{$\bullet$ Video Quality.}
Video quality is the most mature dimension, with most sub-metrics (flickering, smoothness) near saturation across all paradigms. Text-driven models lead narrowly (Seedance~1.5 at 82.1, Wan~2.7 at 81.5), but world models such as LingBot-World (78.9) and Happy~Oyster (77.3) achieve competitive quality without sacrificing their control capabilities, suggesting that video quality is no longer the primary bottleneck differentiating paradigms.

\myparagraph{$\bullet$ Setting Adherence.}
Text-driven models dominate by a wide margin. Wan~2.7 (91.4) and Kling~3.0 (91.0) far exceed the best world models (Happy~Oyster 74.2, LingBot-World 72.6), with the gap concentrated in scene adherence rather than subject adherence, likely because world model training prioritizes navigation fidelity over the broad scene following capabilities of text-driven models.

\myparagraph{$\bullet$ Interaction.}
Navigation favors models with native control interfaces. Camera-controlled (76.0) and action-conditioned (77.7) models exceed text-driven ones (67.6) by approximately 10 points. Notably, YUME~1.5 achieves the highest navigation score (72.0) among text-driven models, likely benefiting from navigation-oriented fine-tuning, which suggests that targeted fine-tuning on navigation data can partially close the gap with dedicated world models. For semantic interactions exclusive to text-driven models, Kling~3.0 and Wan~2.7 dominate event editing and subject action, but perspective switching remains the hardest task (average 30.7).

\begin{figure}[t]
\centering
\includegraphics[width=\linewidth]{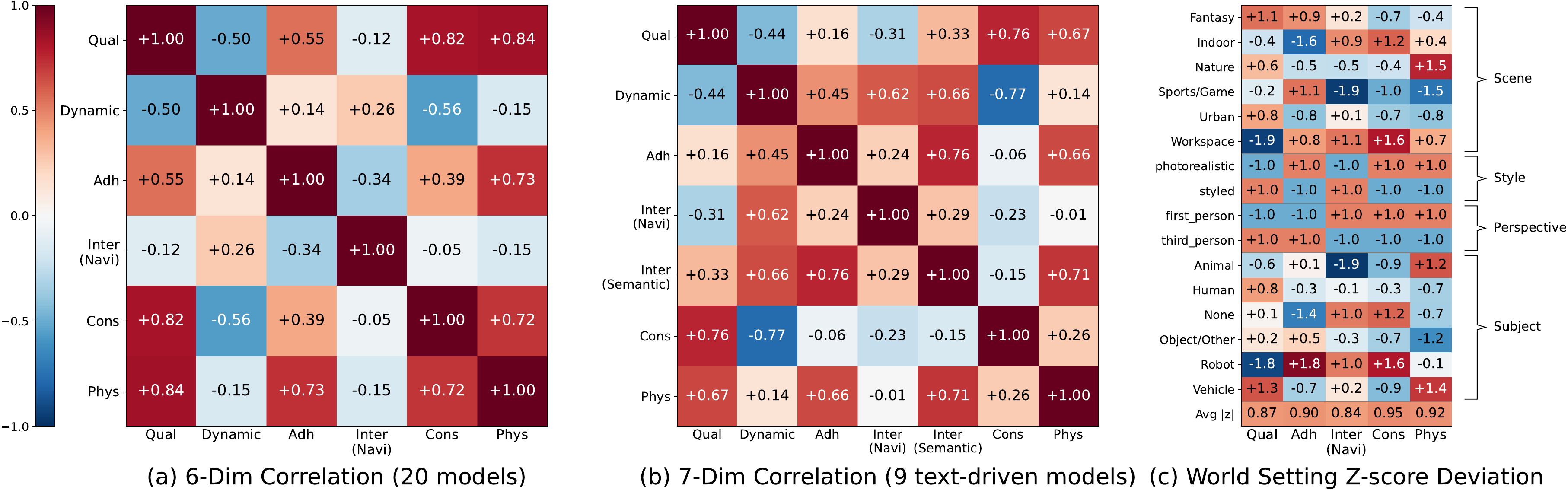}
\vspace{-5mm}
\caption[Cross-dimension correlation and per-setting deviation analysis]{Cross-dimension correlation and per-setting deviation analysis. \textbf{(a)} Pearson correlation among six dimensions ($n$=20 models, navi split). \textbf{(b)} seven-dimension correlation ($n$=9 text-conditioned models), with Interaction split into Navigation and Semantic.\textbf{(c)} per-setting Z-score deviation across five dimensions. Positive (red) = easier, negative (blue) = harder.}
\vspace{-2mm}
\label{fig:metric_correlation}
\end{figure}

\myparagraph{$\bullet$ Consistency.}
LingBot-World achieves the highest overall consistency (89.9), but consistency is multi-faceted. Camera-controlled models lead geometric consistency (93.1 \vs 88.2 for text-driven) thanks to explicit pose supervision, yet average only 67.1 on perspective consistency, underperforming text-driven models (74.7). As shown in \cref{fig:metric_correlation}~(a), dynamic degree is negatively correlated with consistency ($r{=}{-}0.56$), and camera-controlled models drop 15.3 points from spatial to gated spatial consistency (\vs 4.5 for text-driven), confirming that some high scores are inflated by scene stasis rather than maintaining genuine consistency through active motion.

\myparagraph{$\bullet$ Physical.}
Text-driven models (67.0) outperform camera-controlled (64.2) and action-conditioned (61.7) ones, suggesting that broad generative priors contribute more to physical correctness than specialized control training. Wan~2.7 leads (71.8), driven by exceptionally high causal fidelity (83.3) that likely benefits from diverse physical interaction data. LingBot-World leads camera models (71.2), where its strong consistency (89.9) helps maintain a stable context for physical events.

\subsection{Cross-Dimension Analysis}
\label{sec:cross_dimension}

The per-dimension results show that no single model or paradigm dominates uniformly. We now analyze the structural relationships between dimensions (\cref{fig:metric_correlation}) and identify deeper challenges.

\myparagraph{$\bullet$ Navigation Is Decoupled from Other Dimensions.}
The correlations in \cref{fig:metric_correlation}~(a) show that navigation is the most independent dimension, with near-zero correlation to video quality ($r{=}{-}0.12$), consistency ($r{=}{-}0.05$), and physical compliance ($r{=}{-}0.15$), suggesting that strong rendering, memory, or physics performance does not translate into controllable movement. In contrast, physical scores correlate strongly with video quality ($r{=}0.84$) and consistency ($r{=}0.72$) but not navigation ($r{=}{-}0.15$), indicating that physical plausibility is inherited from rich generative priors rather than control capabilities. Within text-driven models (\cref{fig:metric_correlation}~(b)), semantic interactions align with setting adherence ($r{=}0.76$) more than navigation ($r{=}0.29$), confirming that event editing and subject action depend on instruction grounding, while navigation relies on a separate spatial-state representation.

\myparagraph{$\bullet$ Camera Motion Control Does Not Guarantee Perspective Consistency.}
As shown in \cref{fig:metric_correlation}~(a), navigation also exhibits near-zero correlation with perspective consistency, revealing that camera control and subject control are separate capabilities. Several top-navigation models rank lowest in perspective consistency, notably HY-World~1.5 (navigation rank~1, perspective rank~8 among 11 world models) and Matrix-Game~3.0 (navigation rank~3, perspective rank~11). These models navigate accurately but fail to maintain coherent subject motion, especially in third-person cases.

\myparagraph{$\bullet$ Open-Source Models Are Competitive.}
Open-source world models achieve leading scores on multiple dimensions. HY-World~1.5 leads navigation among all models (87.5), LingBot-World leads consistency (89.9), and Matrix-Game~3.0 leads action-conditioned navigation (83.5). These results demonstrate that open-source systems can match or surpass closed-source alternatives on specific capabilities given appropriate architectural and training choices.

\begin{figure}[t]
\centering
\includegraphics[width=\linewidth]{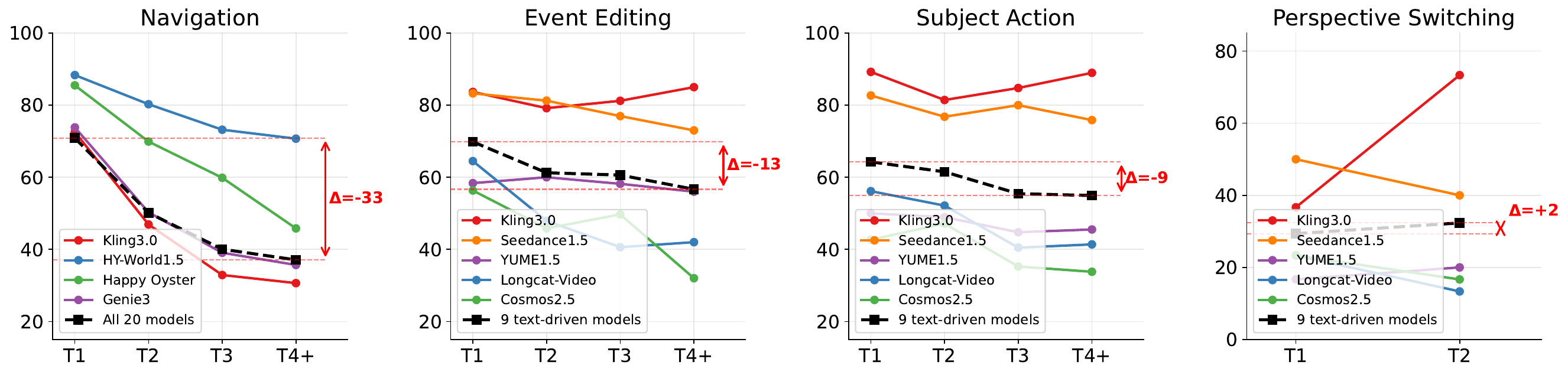}
\vspace{-3mm}
\vspace{-3mm}
\caption[Per-turn performance degradation]{Per-turn performance degradation. T4+ aggregates all turns from the 4$_{th}$ onward.}
\label{fig:degradation}
\vspace{-2mm}
\end{figure}

\myparagraph{$\bullet$ World Settings Induce Structured Difficulty.}
As shown in \cref{fig:metric_correlation}~(c), first-person perspective makes navigation easier ($z{=}{+}1.0$) thanks to the direct action-to-camera mapping, while third-person adds geometric complexity from joint subject-camera control. Sports/game scenes are hardest for navigation ($z{=}{-}1.9$) and animal subjects likewise ($z{=}{-}1.9$) due to fast dynamics and complex non-rigid motion, while workspace scenes ($z{=}{+}1.6$) and robot subjects ($z{=}{+}1.0$) are easiest given their static geometry and rigid-body motion. Benchmark difficulty is thus governed by control-mapping complexity, scene dynamism, and subject rigidity.

\myparagraph{$\bullet$ Navigation Breaks Down Over Turns.}
As shown in \cref{fig:degradation}, navigation degrades much faster than other interaction types ($-$33 points from turn~1 to turn~4+), because it requires maintaining a spatial reference frame where pose errors compound across steps.
Dedicated world models are more robust: HY-World~1.5 degrades much less than Kling~3.0, suggesting that explicit geometric control better preserves spatial state than text-based prompting.
Event editing and subject action degrade moderately ($-$13 and $-$9), driven by accumulated visual artifacts along the interaction chain, though flagship models such as Kling~3.0 and Wan~2.7 remain stable. Perspective switching stays nearly flat (+2), largely because models already perform poorly (average 30.7) with little room to drop further. The slight upward trend in Kling~3.0 is likely because some cases involve symmetric transitions where earlier turns introduce visual priors that benefit later switches.

\subsection{Human Preference Alignment}
\label{sec:human_validation}

\begin{figure}[t]
\centering
\includegraphics[width=0.95\linewidth]{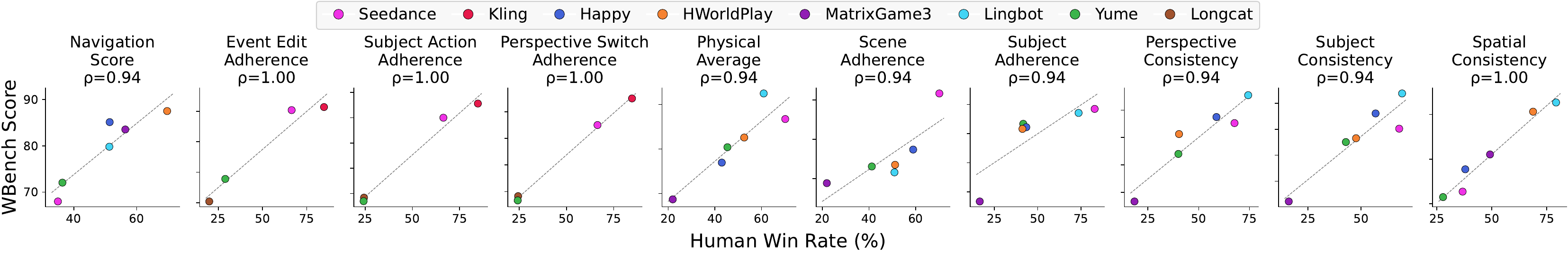}
\vspace{-2.5mm}
\caption[Human-auto alignment across ten evaluation aspects]{Spearman $\rho$ between per-model human win rates ($x$-axis) and automated \benchmark scores ($y$-axis) across ten evaluation aspects. All aspects achieve $\rho \geq 0.94$, with four reaching $\rho = 1.00$.}
\vspace{-4mm}
\label{fig:human_alignment}
\end{figure}

Following the human alignment protocol of VBench~\cite{huang2024vbench}, we validate automated metrics by computing the Spearman rank correlation between per-model human win rates and \benchmark scores. We recruit 400 crowdsourced annotators to perform blind pairwise comparisons across ten evaluation aspects, covering four to six models per aspect. For each comparison, annotators select \textit{A wins}, \textit{B wins}, or \textit{Tie} based on dimension-specific criteria. We aggregate preferences into per-model win rates (Tie counts as 0.5 for each side) and correlate against the corresponding automated score. As shown in \cref{fig:human_alignment}, all ten aspects achieve Spearman $\rho \geq 0.94$, with four (event editing, subject action, perspective switching, and spatial consistency) reaching $\rho = 1.00$, confirming that \benchmark metrics reliably reflect human preference at model-ranking granularity. Details are provided in Appendix~\ref{app:human_annotation}.

\section{Conclusion}
\label{sec:conclusion}

We introduce \benchmark, a benchmark for evaluating interactive world models across five complementary dimensions through explicit world settings and multi-turn interactions.
Experiments on \nummodel models reveal that no model dominates all dimensions, navigation is largely independent of other capabilities, camera control does not imply subject control, and physical correctness follows rendering quality rather than control ability.
These findings confirm that current world models have not yet unified high-fidelity rendering with reliable controllability, consistency, and physics.

\myparagraph{Limitations.}
The current test set focuses on discrete action sequences rather than continuous control. The physical dimension relies partly on LMM-based evaluation whose reliability may degrade for subtle effects. Expanding to additional domains and real-time evaluation are promising extensions.

{
\small
\bibliographystyle{unsrtnat}
\bibliography{neurips_2026}

\begin{thebibliography}{84}
\providecommand{\natexlab}[1]{#1}
\providecommand{\url}[1]{\texttt{#1}}
\expandafter\ifx\csname urlstyle\endcsname\relax
  \providecommand{\doi}[1]{doi: #1}\else
  \providecommand{\doi}{doi: \begingroup \urlstyle{rm}\Url}\fi

\bibitem[Ho et~al.(2020)Ho, Jain, and Abbeel]{ho2020denoising}
Jonathan Ho, Ajay Jain, and Pieter Abbeel.
\newblock Denoising diffusion probabilistic models.
\newblock \emph{Advances in neural information processing systems}, 33:\penalty0 6840--6851, 2020.

\bibitem[Kong et~al.(2024)Kong, Tian, Zhang, Min, Dai, Zhou, Xiong, Li, Wu, Zhang, et~al.]{kong2024hunyuanvideo}
Weijie Kong, Qi~Tian, Zijian Zhang, Rox Min, Zuozhuo Dai, Jin Zhou, Jiangfeng Xiong, Xin Li, Bo~Wu, Jianwei Zhang, et~al.
\newblock Hunyuanvideo: A systematic framework for large video generative models.
\newblock \emph{arXiv preprint arXiv:2412.03603}, 2024.

\bibitem[Wan et~al.(2025{\natexlab{a}})Wan, Wang, Ai, Wen, Mao, Xie, Chen, Yu, Zhao, Yang, et~al.]{wan2025wan}
Team Wan, Ang Wang, Baole Ai, Bin Wen, Chaojie Mao, Chen-Wei Xie, Di~Chen, Feiwu Yu, Haiming Zhao, Jianxiao Yang, et~al.
\newblock Wan: Open and advanced large-scale video generative models.
\newblock \emph{arXiv preprint arXiv:2503.20314}, 2025{\natexlab{a}}.

\bibitem[Polyak et~al.(2024)Polyak, Zohar, Brown, Tjandra, Sinha, Lee, Vyas, Shi, Ma, Chuang, et~al.]{polyak2024moviegen}
Adam Polyak, Amit Zohar, Andrew Brown, Andros Tjandra, Animesh Sinha, Ann Lee, Apoorv Vyas, Bowen Shi, Chih-Yao Ma, Ching-Yao Chuang, et~al.
\newblock Movie gen: A cast of media foundation models.
\newblock \emph{arXiv preprint arXiv:2410.13720}, 2024.

\bibitem[Brooks et~al.(2024)Brooks, Peebles, Holmes, DePue, Guo, Jing, Schnurr, Taylor, Luhman, Luhman, et~al.]{brooks2024sora}
Tim Brooks, Bill Peebles, Connor Holmes, Will DePue, Yufei Guo, Leo Jing, David Schnurr, Joe Taylor, Troy Luhman, Eric Luhman, et~al.
\newblock Video generation models as world simulators.
\newblock \emph{OpenAI Blog}, 1\penalty0 (8):\penalty0 1, 2024.

\bibitem[Bruce et~al.(2024)Bruce, Dennis, Edwards, Parker-Holder, Shi, Hughes, Lai, Mavalankar, Steigerwald, Apps, et~al.]{bruce2024genie}
Jake Bruce, Michael~D Dennis, Ashley Edwards, Jack Parker-Holder, Yuge Shi, Edward Hughes, Matthew Lai, Aditi Mavalankar, Richie Steigerwald, Chris Apps, et~al.
\newblock Genie: Generative interactive environments.
\newblock In \emph{Forty-first International Conference on Machine Learning}, 2024.

\bibitem[Parker-Holder et~al.(2024)Parker-Holder, Ball, Bruce, Dasagi, Holsheimer, Kaplanis, Moufarek, Scully, Shar, Shi, et~al.]{deepmind2024genie2}
Jack Parker-Holder, Philip Ball, Jake Bruce, Vibhavari Dasagi, Kristian Holsheimer, Christos Kaplanis, Alexandre Moufarek, Guy Scully, Jeremy Shar, Jimmy Shi, et~al.
\newblock Genie 2: A large-scale foundation world model.
\newblock \url{https://deepmind.google/discover/blog/genie-2-a-large-scale-foundation-world-model/}, 2024.

\bibitem[Ball et~al.(2025)Ball, Bauer, Belletti, Brownfield, Ephrat, Fruchter, Gupta, Holsheimer, Holynski, Hron, Kaplanis, Limont, McGill, Oliveira, Parker-Holder, Perbet, Scully, Shar, Spencer, Tov, Villegas, Wang, Yung, Baetu, Berbel, Bridson, Bruce, Buttimore, Chakera, Chandra, Collins, Cullum, Damoc, Dasagi, Gazeau, Gbadamosi, Han, Hirst, Kachra, Kerley, Kjems, Knoepfel, Koriakin, Lo, Lu, Mehring, Moufarek, Nandwani, Oliveira, Pardo, Park, Pierson, Poole, Ran, Salimans, Sanchez, Saprykin, Shen, Sidhwani, Smith, Stanton, Tomlinson, Vijaykumar, Wang, Wingfield, Wong, Xu, Yew, Young, Zubov, Eck, Erhan, Kavukcuoglu, Hassabis, Gharamani, Hadsell, van~den Oord, Mosseri, Bolton, Singh, and Rockt{\"a}schel]{ball2025genie3}
Philip~J. Ball, Jakob Bauer, Frank Belletti, Bethanie Brownfield, Ariel Ephrat, Shlomi Fruchter, Agrim Gupta, Kristian Holsheimer, Aleksander Holynski, Jiri Hron, Christos Kaplanis, Marjorie Limont, Matt McGill, Yanko Oliveira, Jack Parker-Holder, Frank Perbet, Guy Scully, Jeremy Shar, Stephen Spencer, Omer Tov, Ruben Villegas, Emma Wang, Jessica Yung, Cip Baetu, Jordi Berbel, David Bridson, Jake Bruce, Gavin Buttimore, Sarah Chakera, Bilva Chandra, Paul Collins, Alex Cullum, Bogdan Damoc, Vibha Dasagi, Maxime Gazeau, Charles Gbadamosi, Woohyun Han, Ed~Hirst, Ashyana Kachra, Lucie Kerley, Kristian Kjems, Eva Knoepfel, Vika Koriakin, Jessica Lo, Cong Lu, Zeb Mehring, Alex Moufarek, Henna Nandwani, Valeria Oliveira, Fabio Pardo, Jane Park, Andrew Pierson, Ben Poole, Helen Ran, Tim Salimans, Manuel Sanchez, Igor Saprykin, Amy Shen, Sailesh Sidhwani, Duncan Smith, Joe Stanton, Hamish Tomlinson, Dimple Vijaykumar, Luyu Wang, Piers Wingfield, Nat Wong, Keyang Xu, Christopher Yew, Nick Young, Vadim Zubov, Douglas Eck, Dumitru Erhan, Koray Kavukcuoglu, Demis Hassabis, Zoubin Gharamani, Raia Hadsell, A{\"a}ron van~den Oord, Inbar Mosseri, Adrian Bolton, Satinder Singh, and Tim Rockt{\"a}schel.
\newblock Genie 3: A new frontier for world models.
\newblock \url{https://deepmind.google/discover/blog/genie-3-a-new-frontier-for-world-models/}, 2025.

\bibitem[He et~al.(2025)He, Peng, Liu, Wang, Zhang, Cui, Kang, Jiang, An, Ren, et~al.]{he2025matrixgame2}
Xianglong He, Chunli Peng, Zexiang Liu, Boyang Wang, Yifan Zhang, Qi~Cui, Fei Kang, Biao Jiang, Mengyin An, Yangyang Ren, et~al.
\newblock Matrix-game 2.0: An open-source real-time and streaming interactive world model.
\newblock \emph{arXiv preprint arXiv:2508.13009}, 2025.

\bibitem[Wang et~al.(2026)Wang, Liu, Li, Huang, Xu, Kang, An, Wang, Jiang, Wei, et~al.]{he2026matrixgame3}
Zile Wang, Zexiang Liu, Jaixing Li, Kaichen Huang, Baixin Xu, Fei Kang, Mengyin An, Peiyu Wang, Biao Jiang, Yichen Wei, et~al.
\newblock Matrix-game 3.0: Real-time and streaming interactive world model with long-horizon memory.
\newblock \emph{arXiv preprint arXiv:2604.08995}, 2026.

\bibitem[Li et~al.(2025{\natexlab{a}})Li, Tang, Xu, Wu, Zhou, Shao, Yu, Cao, and Lu]{li2025hunyuangamecraft}
Jiaqi Li, Junshu Tang, Zhiyong Xu, Longhuang Wu, Yuan Zhou, Shuai Shao, Tianbao Yu, Zhiguo Cao, and Qinglin Lu.
\newblock Hunyuan-gamecraft: High-dynamic interactive game video generation with hybrid history condition.
\newblock \emph{arXiv preprint arXiv:2506.17201}, 2\penalty0 (3):\penalty0 6, 2025{\natexlab{a}}.

\bibitem[Tang et~al.(2025)Tang, Liu, Li, Wu, Yang, Zhao, Gong, Yuan, Shao, Zhang, et~al.]{tang2025interbench}
Junshu Tang, Jiacheng Liu, Jiaqi Li, Longhuang Wu, Haoyu Yang, Penghao Zhao, Siruis Gong, Xiang Yuan, Shuai Shao, Linfeng Zhang, et~al.
\newblock Hunyuan-gamecraft-2: Instruction-following interactive game world model.
\newblock \emph{arXiv preprint arXiv:2511.23429}, 2025.

\bibitem[Hu et~al.(2023)Hu, Russell, Yeo, Murez, Fedoseev, Kendall, Shotton, and Corrado]{hu2023gaia1}
Anthony Hu, Lloyd Russell, Hudson Yeo, Zak Murez, George Fedoseev, Alex Kendall, Jamie Shotton, and Gianluca Corrado.
\newblock Gaia-1: A generative world model for autonomous driving.
\newblock \emph{arXiv preprint arXiv:2309.17080}, 2023.

\bibitem[Gao et~al.(2024)Gao, Yang, Chen, Chitta, Qiu, Geiger, Zhang, and Li]{gao2024vista}
Shenyuan Gao, Jiazhi Yang, Li~Chen, Kashyap Chitta, Yihang Qiu, Andreas Geiger, Jun Zhang, and Hongyang Li.
\newblock Vista: A generalizable driving world model with high fidelity and versatile controllability.
\newblock \emph{Advances in Neural Information Processing Systems}, 37:\penalty0 91560--91596, 2024.

\bibitem[Yang et~al.(2023)Yang, Du, Ghasemipour, Tompson, Kaelbling, Schuurmans, and Abbeel]{yang2023unisim}
Sherry Yang, Yilun Du, Kamyar Ghasemipour, Jonathan Tompson, Leslie Kaelbling, Dale Schuurmans, and Pieter Abbeel.
\newblock Learning interactive real-world simulators.
\newblock \emph{arXiv preprint arXiv:2310.06114}, 2023.

\bibitem[Zhu et~al.(2024)Zhu, Wu, Guo, Liu, Cheang, and Kong]{zhu2024irasim}
Fangqi Zhu, Hongtao Wu, Song Guo, Yuxiao Liu, Chilam Cheang, and Tao Kong.
\newblock Irasim: Learning interactive real-robot action simulators.
\newblock \emph{arXiv preprint arXiv:2406.14540}, 1\penalty0 (2):\penalty0 3, 2024.

\bibitem[HunyuanWorld(2025)]{sun2025hyworldplay}
Team HunyuanWorld.
\newblock Hy-world 1.5: A systematic framework for interactive world modeling with real-time latency and geometric consistency.
\newblock \emph{arXiv preprint}, 2025.

\bibitem[Mao et~al.(2025{\natexlab{a}})Mao, Lin, Li, Li, Peng, He, Pang, Chi, Qiao, and Zhang]{chen2025yume}
Xiaofeng Mao, Shaoheng Lin, Zhen Li, Chuanhao Li, Wenshuo Peng, Tong He, Jiangmiao Pang, Mingmin Chi, Yu~Qiao, and Kaipeng Zhang.
\newblock Yume: An interactive world generation model.
\newblock \emph{arXiv preprint arXiv:2507.17744}, 2025{\natexlab{a}}.

\bibitem[Mao et~al.(2025{\natexlab{b}})Mao, Li, Li, Xu, Ying, He, Pang, Qiao, and Zhang]{mao2025yume15}
Xiaofeng Mao, Zhen Li, Chuanhao Li, Xiaojie Xu, Kaining Ying, Tong He, Jiangmiao Pang, Yu~Qiao, and Kaipeng Zhang.
\newblock Yume-1.5: A text-controlled interactive world generation model.
\newblock \emph{arXiv preprint arXiv:2512.22096}, 2025{\natexlab{b}}.

\bibitem[Team et~al.(2026{\natexlab{a}})Team, Gao, Wang, Zeng, Zhu, Cheng, Li, Wang, Xu, Ma, et~al.]{gao2026lingbot}
Robbyant Team, Zelin Gao, Qiuyu Wang, Yanhong Zeng, Jiapeng Zhu, Ka~Leong Cheng, Yixuan Li, Hanlin Wang, Yinghao Xu, Shuailei Ma, et~al.
\newblock Advancing open-source world models.
\newblock \emph{arXiv preprint arXiv:2601.20540}, 2026{\natexlab{a}}.

\bibitem[Huang et~al.(2024)Huang, He, Yu, Zhang, Si, Jiang, Zhang, Wu, Jin, Chanpaisit, et~al.]{huang2024vbench}
Ziqi Huang, Yinan He, Jiashuo Yu, Fan Zhang, Chenyang Si, Yuming Jiang, Yuanhan Zhang, Tianxing Wu, Qingyang Jin, Nattapol Chanpaisit, et~al.
\newblock Vbench: Comprehensive benchmark suite for video generative models.
\newblock In \emph{Proceedings of the IEEE/CVF Conference on Computer Vision and Pattern Recognition}, pages 21807--21818, 2024.

\bibitem[Zheng et~al.(2025)Zheng, Huang, Liu, Zou, He, Zhang, Gu, Zhang, He, Zheng, et~al.]{zheng2025vbench2}
Dian Zheng, Ziqi Huang, Hongbo Liu, Kai Zou, Yinan He, Fan Zhang, Lulu Gu, Yuanhan Zhang, Jingwen He, Wei-Shi Zheng, et~al.
\newblock Vbench-2.0: Advancing video generation benchmark suite for intrinsic faithfulness.
\newblock \emph{arXiv preprint arXiv:2503.21755}, 2025.

\bibitem[Xu et~al.(2026)Xu, Lin, He, Feng, Mao, Yin, Zhang, and Ge]{xu2026worldmark}
Xiaojie Xu, Zhengyuan Lin, Kang He, Yukang Feng, Xiaofeng Mao, Yuanyang Yin, Kaipeng Zhang, and Yongtao Ge.
\newblock Worldmark: A unified benchmark suite for interactive video world models.
\newblock \emph{arXiv preprint arXiv:2604.21686}, 2026.

\bibitem[Ye et~al.(2026)Ye, Lu, Jiang, Gu, Zhao, Liang, Pan, Zhang, Wu, and Wang]{ye2026mind}
Yixuan Ye, Xuanyu Lu, Yuxin Jiang, Yuchao Gu, Rui Zhao, Qiwei Liang, Jiachun Pan, Fengda Zhang, Weijia Wu, and Alex~Jinpeng Wang.
\newblock Mind: Benchmarking memory consistency and action control in world models.
\newblock \emph{arXiv preprint arXiv:2602.08025}, 2026.

\bibitem[Wu et~al.(2026{\natexlab{a}})Wu, Cai, Zhao, Feng, Dang, Song, Tian, Zhu, Lei, Dou, et~al.]{wu2026omniworldbench}
Meiqi Wu, Zhixin Cai, Fufangchen Zhao, Xiaokun Feng, Rujing Dang, Bingze Song, Ruitian Tian, Jiashu Zhu, Jiachen Lei, Hao Dou, et~al.
\newblock Omni-worldbench: Towards a comprehensive interaction-centric evaluation for world models.
\newblock \emph{arXiv preprint arXiv:2603.22212}, 2026{\natexlab{a}}.

\bibitem[Liang et~al.(2025)Liang, Kong, Yan, Liu, Yang, Huang, Yin, Zuo, Hu, Zhu, et~al.]{worldlens2025}
Ao~Liang, Lingdong Kong, Tianyi Yan, Hongsi Liu, Wesley Yang, Ziqi Huang, Wei Yin, Jialong Zuo, Yixuan Hu, Dekai Zhu, et~al.
\newblock Worldlens: Full-spectrum evaluations of driving world models in real world.
\newblock \emph{arXiv preprint arXiv:2512.10958}, 2025.

\bibitem[Ho et~al.(2022)Ho, Salimans, Gritsenko, Chan, Norouzi, and Fleet]{ho2022video}
Jonathan Ho, Tim Salimans, Alexey Gritsenko, William Chan, Mohammad Norouzi, and David~J Fleet.
\newblock Video diffusion models.
\newblock \emph{Advances in neural information processing systems}, 35:\penalty0 8633--8646, 2022.

\bibitem[Blattmann et~al.(2023)Blattmann, Rombach, Ling, Dockhorn, Kim, Fidler, and Kreis]{blattmann2023align}
Andreas Blattmann, Robin Rombach, Huan Ling, Tim Dockhorn, Seung~Wook Kim, Sanja Fidler, and Karsten Kreis.
\newblock Align your latents: High-resolution video synthesis with latent diffusion models.
\newblock In \emph{Proceedings of the IEEE/CVF conference on computer vision and pattern recognition}, pages 22563--22575, 2023.

\bibitem[Yang et~al.(2024)Yang, Teng, Zheng, Ding, Huang, Xu, Yang, Hong, Zhang, Feng, et~al.]{yang2024cogvideox}
Zhuoyi Yang, Jiayan Teng, Wendi Zheng, Ming Ding, Shiyu Huang, Jiazheng Xu, Yuanming Yang, Wenyi Hong, Xiaohan Zhang, Guanyu Feng, et~al.
\newblock Cogvideox: Text-to-video diffusion models with an expert transformer.
\newblock \emph{arXiv preprint arXiv:2408.06072}, 2024.

\bibitem[{OpenAI}(2025{\natexlab{a}})]{openai2025sora2}
{OpenAI}.
\newblock {Sora 2}.
\newblock \url{https://openai.com/zh-Hans-CN/index/sora-2/}, 2025{\natexlab{a}}.

\bibitem[{Kuaishou Technology}(2025)]{kuaishou2025kling3}
{Kuaishou Technology}.
\newblock Kling 3.0 pro.
\newblock \url{https://klingai.com}, 2025.

\bibitem[{Google DeepMind}(2025)]{deepmind2025veo3}
{Google DeepMind}.
\newblock {Veo 3}: State-of-the-art video generation with audio.
\newblock \url{https://deepmind.google/models/veo/}, 2025.

\bibitem[Wan et~al.(2025{\natexlab{b}})Wan, Wang, Ai, Wen, Mao, Xie, Chen, Yu, Zhao, Yang, et~al.]{wan2025wan27}
Team Wan, Ang Wang, Baole Ai, Bin Wen, Chaojie Mao, Chen-Wei Xie, Di~Chen, Feiwu Yu, Haiming Zhao, Jianxiao Yang, et~al.
\newblock Wan: Open and advanced large-scale video generative models.
\newblock \emph{arXiv preprint arXiv:2503.20314}, 2025{\natexlab{b}}.

\bibitem[Seedance et~al.(2026)Seedance, Chen, Chen, Chen, Chen, Chen, Chen, Cheng, Cheng, Cheng, et~al.]{gao2026seedance2}
Team Seedance, De~Chen, Liyang Chen, Xin Chen, Ying Chen, Zhuo Chen, Zhuowei Chen, Feng Cheng, Tianheng Cheng, Yufeng Cheng, et~al.
\newblock Seedance 2.0: Advancing video generation for world complexity.
\newblock \emph{arXiv preprint arXiv:2604.14148}, 2026.

\bibitem[{Shengshu Technology}(2025)]{shengshu2025viduq3}
{Shengshu Technology}.
\newblock Vidu q3 pro.
\newblock \url{https://www.vidu.com}, 2025.

\bibitem[HaCohen et~al.(2024)HaCohen, Chiprut, Brazowski, Shalem, Moshe, Richardson, Levin, Shiran, Zabari, Gordon, et~al.]{lightricks2025ltx}
Yoav HaCohen, Nisan Chiprut, Benny Brazowski, Daniel Shalem, Dudu Moshe, Eitan Richardson, Eran Levin, Guy Shiran, Nir Zabari, Ori Gordon, et~al.
\newblock Ltx-video: Realtime video latent diffusion.
\newblock \emph{arXiv preprint arXiv:2501.00103}, 2024.

\bibitem[Gu(2025)]{nvidia2025cosmos}
Jinwei Gu.
\newblock Cosmos world foundation models for physical ai.
\newblock In \emph{Proceedings of the 3rd International Workshop on Rich Media With Generative AI}, pages 39--39, 2025.

\bibitem[Team et~al.(2025)Team, Cai, Huang, Kang, Li, Liang, Ma, Ren, Wei, Xie, et~al.]{cai2025longcat}
Meituan~LongCat Team, Xunliang Cai, Qilong Huang, Zhuoliang Kang, Hongyu Li, Shijun Liang, Liya Ma, Siyu Ren, Xiaoming Wei, Rixu Xie, et~al.
\newblock Longcat-video technical report.
\newblock \emph{arXiv preprint arXiv:2510.22200}, 2025.

\bibitem[Heusel et~al.(2017)Heusel, Ramsauer, Unterthiner, Nessler, and Hochreiter]{heusel2017fid}
Martin Heusel, Hubert Ramsauer, Thomas Unterthiner, Bernhard Nessler, and Sepp Hochreiter.
\newblock Gans trained by a two time-scale update rule converge to a local nash equilibrium.
\newblock \emph{Advances in neural information processing systems}, 30, 2017.

\bibitem[Unterthiner et~al.(2018)Unterthiner, Van~Steenkiste, Kurach, Marinier, Michalski, and Gelly]{unterthiner2019fvd}
Thomas Unterthiner, Sjoerd Van~Steenkiste, Karol Kurach, Raphael Marinier, Marcin Michalski, and Sylvain Gelly.
\newblock Towards accurate generative models of video: A new metric \& challenges.
\newblock \emph{arXiv preprint arXiv:1812.01717}, 2018.

\bibitem[Ha and Schmidhuber(2018)]{ha2018world}
David Ha and J{\"u}rgen Schmidhuber.
\newblock World models.
\newblock \emph{arXiv preprint arXiv:1803.10122}, 2\penalty0 (3):\penalty0 440, 2018.

\bibitem[LeCun et~al.(2022)]{lecun2022path}
Yann LeCun et~al.
\newblock A path towards autonomous machine intelligence version 0.9. 2, 2022-06-27.
\newblock \emph{Open Review}, 62\penalty0 (1):\penalty0 1--62, 2022.

\bibitem[{Alibaba Token Hub}(2026)]{alibaba2026happyoyster}
{Alibaba Token Hub}.
\newblock {Happy Oyster}: An open-ended world model for real-time world creation and interaction.
\newblock \url{https://happyoyster.cn/}, 2026.

\bibitem[{World Labs}(2025)]{worldlabs2025marble}
{World Labs}.
\newblock Marble: A multimodal world model.
\newblock \url{https://www.worldlabs.ai/blog/marble-world-model}, 2025.

\bibitem[Huang et~al.(2025)Huang, Zhang, Xu, He, Yu, Dong, Ma, Chanpaisit, Si, Jiang, et~al.]{huang2024vbenchplus}
Ziqi Huang, Fan Zhang, Xiaojie Xu, Yinan He, Jiashuo Yu, Ziyue Dong, Qianli Ma, Nattapol Chanpaisit, Chenyang Si, Yuming Jiang, et~al.
\newblock Vbench++: Comprehensive and versatile benchmark suite for video generative models.
\newblock \emph{IEEE Transactions on Pattern Analysis and Machine Intelligence}, 2025.

\bibitem[Liu et~al.(2024)Liu, Cun, Liu, Wang, Zhang, Chen, Liu, Zeng, Chan, and Shan]{liu2024evalcrafter}
Yaofang Liu, Xiaodong Cun, Xuebo Liu, Xintao Wang, Yong Zhang, Haoxin Chen, Yang Liu, Tieyong Zeng, Raymond Chan, and Ying Shan.
\newblock Evalcrafter: Benchmarking and evaluating large video generation models.
\newblock In \emph{Proceedings of the IEEE/CVF conference on computer vision and pattern recognition}, pages 22139--22149, 2024.

\bibitem[Bansal et~al.(2024)Bansal, Lin, Xie, Zong, Yarom, Bitton, Jiang, Sun, Chang, and Grover]{bansal2024videophy}
Hritik Bansal, Zongyu Lin, Tianyi Xie, Zeshun Zong, Michal Yarom, Yonatan Bitton, Chenfanfu Jiang, Yizhou Sun, Kai-Wei Chang, and Aditya Grover.
\newblock Videophy: Evaluating physical commonsense for video generation.
\newblock \emph{arXiv preprint arXiv:2406.03520}, 2024.

\bibitem[Meng et~al.(2024)Meng, Liao, Tan, Shao, Lu, Zhang, Cheng, Li, Qiao, and Luo]{meng2024phygenbench}
Fanqing Meng, Jiaqi Liao, Xinyu Tan, Wenqi Shao, Quanfeng Lu, Kaipeng Zhang, Yu~Cheng, Dianqi Li, Yu~Qiao, and Ping Luo.
\newblock Towards world simulator: Crafting physical commonsense-based benchmark for video generation.
\newblock \emph{arXiv preprint arXiv:2410.05363}, 2024.

\bibitem[Duan et~al.(2025)Duan, Yu, Chen, Fei-Fei, and Wu]{duan2024worldscore}
Haoyi Duan, Hong-Xing Yu, Sirui Chen, Li~Fei-Fei, and Jiajun Wu.
\newblock Worldscore: A unified evaluation benchmark for world generation.
\newblock In \emph{Proceedings of the IEEE/CVF International Conference on Computer Vision}, pages 27713--27724, 2025.

\bibitem[Li et~al.(2025{\natexlab{b}})Li, Fang, Chen, Yang, Cao, Wong, Luo, Wang, Yin, Gonzalez, et~al.]{wang2024worldmodelbench}
Dacheng Li, Yunhao Fang, Yukang Chen, Shuo Yang, Shiyi Cao, Justin Wong, Michael Luo, Xiaolong Wang, Hongxu Yin, Joseph~E Gonzalez, et~al.
\newblock Worldmodelbench: Judging video generation models as world models.
\newblock \emph{arXiv preprint arXiv:2502.20694}, 2025{\natexlab{b}}.

\bibitem[Shang et~al.(2026)Shang, Li, Ma, Su, Jin, Wang, Jin, Zhang, Tang, Su, et~al.]{shang2026worldarena}
Yu~Shang, Zhuohang Li, Yiding Ma, Weikang Su, Xin Jin, Ziyou Wang, Lei Jin, Xin Zhang, Yinzhou Tang, Haisheng Su, et~al.
\newblock Worldarena: A unified benchmark for evaluating perception and functional utility of embodied world models.
\newblock \emph{arXiv preprint arXiv:2602.08971}, 2026.

\bibitem[Han et~al.(2025)Han, Li, Chen, Yuan, Wu, Deng, Leong, Du, Fu, Li, et~al.]{han2025videobench}
Hui Han, Siyuan Li, Jiaqi Chen, Yiwen Yuan, Yuling Wu, Yufan Deng, Chak~Tou Leong, Hanwen Du, Junchen Fu, Youhua Li, et~al.
\newblock Video-bench: Human-aligned video generation benchmark.
\newblock In \emph{Proceedings of the Computer Vision and Pattern Recognition Conference}, pages 18858--18868, 2025.

\bibitem[Liu et~al.(2023)Liu, Li, Ren, Gao, Li, Chen, Sun, and Hou]{liu2024fetv}
Yuanxin Liu, Lei Li, Shuhuai Ren, Rundong Gao, Shicheng Li, Sishuo Chen, Xu~Sun, and Lu~Hou.
\newblock Fetv: A benchmark for fine-grained evaluation of open-domain text-to-video generation.
\newblock \emph{Advances in Neural Information Processing Systems}, 36:\penalty0 62352--62387, 2023.

\bibitem[Gu et~al.(2026)Gu, Liu, Zeng, Nagarajan, Zhu, Hong, Fan, Yan, Zhou, Liu, and Wang]{gu2025phyworldbench}
Jing Gu, Xian Liu, Yu~Zeng, Ashwin Nagarajan, Fangrui Zhu, Daniel Hong, Yue Fan, Qianqi Yan, Kaiwen Zhou, Ming-Yu Liu, and Xin~Eric Wang.
\newblock "phyworldbench": A comprehensive evaluation of physical realism in text-to-video models, 2026.
\newblock URL \url{https://arxiv.org/abs/2507.13428}.

\bibitem[Cai et~al.(2025)Cai, Qiu, Ma, Zhao, Zhou, Huang, Kordjamshidi, Zhang, Xiao, Gu, et~al.]{cai2025mmgr}
Zefan Cai, Haoyi Qiu, Tianyi Ma, Haozhe Zhao, Gengze Zhou, Kung-Hsiang Huang, Parisa Kordjamshidi, Minjia Zhang, Wen Xiao, Jiuxiang Gu, et~al.
\newblock Mmgr: Multi-modal generative reasoning.
\newblock \emph{arXiv preprint arXiv:2512.14691}, 2025.

\bibitem[Upadhyay et~al.(2026)Upadhyay, Zhang, Solomon, Agrawal, Boreddy, Narayana, Ba, Wong, de~Melo, and Kadambi]{upadhyay2026worldbench}
Rishi Upadhyay, Howard Zhang, Jim Solomon, Ayush Agrawal, Pranay Boreddy, Shruti~Satya Narayana, Yunhao Ba, Alex Wong, Celso~M de~Melo, and Achuta Kadambi.
\newblock Worldbench: Disambiguating physics for diagnostic evaluation of world models.
\newblock \emph{arXiv preprint arXiv:2601.21282}, 2026.

\bibitem[Lu et~al.(2025)Lu, Luo, Tu, Li, Zhu, Yu, Wang, Chen, Peng, Li, et~al.]{lu2025_4dworldbench}
Yiting Lu, Wei Luo, Peiyan Tu, Haoran Li, Hanxin Zhu, Zihao Yu, Xingrui Wang, Xinyi Chen, Xinge Peng, Xin Li, et~al.
\newblock 4dworldbench: A comprehensive evaluation framework for 3d/4d world generation models.
\newblock \emph{arXiv preprint arXiv:2511.19836}, 2025.

\bibitem[Team et~al.(2026{\natexlab{b}})Team, Gao, Zhou, Xiang, Liu, Yang, Chen, Ahmad, Zeng, Bannur, Huang, Liu, Gu, Yang, Liu, Hu, Liu, and Xing]{gao2026wrena}
PAN Team, Qiyue Gao, Kun Zhou, Jiannan Xiang, Zihan Liu, Dequan Yang, Junrong Chen, Arif Ahmad, Cong Zeng, Ganesh Bannur, Xinqi Huang, Zheqi Liu, Yi~Gu, Yichi Yang, Guangyi Liu, Zhiting Hu, Zhengzhong Liu, and Eric Xing.
\newblock World reasoning arena, 2026{\natexlab{b}}.

\bibitem[Qin et~al.(2024)Qin, Shi, Yu, Wang, Zhou, Li, Yin, Liu, Sheng, Shao, et~al.]{lian2024worldsimbench}
Yiran Qin, Zhelun Shi, Jiwen Yu, Xijun Wang, Enshen Zhou, Lijun Li, Zhenfei Yin, Xihui Liu, Lu~Sheng, Jing Shao, et~al.
\newblock Worldsimbench: Towards video generation models as world simulators.
\newblock \emph{arXiv preprint arXiv:2410.18072}, 2024.

\bibitem[Yue et~al.(2025)Yue, Huang, Liao, Chen, Zhou, Chen, Yao, and Ren]{yue2025ewmbench}
Hu~Yue, Siyuan Huang, Yue Liao, Shengcong Chen, Pengfei Zhou, Liliang Chen, Maoqing Yao, and Guanghui Ren.
\newblock Ewmbench: Evaluating scene, motion, and semantic quality in embodied world models.
\newblock \emph{arXiv preprint arXiv:2505.09694}, 2025.

\bibitem[Zhang et~al.(2025)Zhang, Jiang, Dai, Lu, Uzunoglu, Zhang, Wei, Wang, Patel, Liang, et~al.]{zhang2025worldinworld}
Jiahan Zhang, Muqing Jiang, Nanru Dai, Taiming Lu, Arda Uzunoglu, Shunchi Zhang, Yana Wei, Jiahao Wang, Vishal~M Patel, Paul~Pu Liang, et~al.
\newblock World-in-world: World models in a closed-loop world.
\newblock \emph{arXiv preprint arXiv:2510.18135}, 2025.

\bibitem[Zhou et~al.(2026)Zhou, Shao, Wang, Zong, Li, and Waslander]{zhou2026drivinggen}
Yang Zhou, Hao Shao, Letian Wang, Zhuofan Zong, Hongsheng Li, and Steven~L Waslander.
\newblock Drivinggen: A comprehensive benchmark for generative video world models in autonomous driving.
\newblock \emph{arXiv preprint arXiv:2601.01528}, 2026.

\bibitem[{Google}(2025)]{google2025nanobanana2}
{Google}.
\newblock Nano banana 2.
\newblock \url{https://blog.google/innovation-and-ai/technology/ai/nano-banana-2/}, 2025.

\bibitem[{OpenAI}(2025{\natexlab{b}})]{openai2025gptimage}
{OpenAI}.
\newblock {GPT-Image-1.5}.
\newblock \url{https://openai.com/zh-Hans-CN/index/new-chatgpt-images-is-here/}, 2025{\natexlab{b}}.

\bibitem[Ma et~al.(2025)Ma, Wu, Sun, and Li]{hpsv3}
Yuhang Ma, Xiaoshi Wu, Keqiang Sun, and Hongsheng Li.
\newblock Hpsv3: Towards wide-spectrum human preference score, 2025.
\newblock URL \url{https://arxiv.org/abs/2508.03789}.

\bibitem[Li et~al.(2025{\natexlab{c}})Li, Tucker, Cole, Wang, Jin, Ye, Kanazawa, Holynski, and Snavely]{li2024megasam}
Zhengqi Li, Richard Tucker, Forrester Cole, Qianqian Wang, Linyi Jin, Vickie Ye, Angjoo Kanazawa, Aleksander Holynski, and Noah Snavely.
\newblock Megasam: Accurate, fast and robust structure and motion from casual dynamic videos.
\newblock In \emph{Proceedings of the IEEE/CVF Conference on Computer Vision and Pattern Recognition}, pages 10486--10496, 2025{\natexlab{c}}.

\bibitem[Fu et~al.(2023)Fu, Tamir, Sundaram, Chai, Zhang, Dekel, and Isola]{fu2023dreamsim}
Stephanie Fu, Netanel Tamir, Shobhita Sundaram, Lucy Chai, Richard Zhang, Tali Dekel, and Phillip Isola.
\newblock Dreamsim: Learning new dimensions of human visual similarity using synthetic data.
\newblock \emph{arXiv preprint arXiv:2306.09344}, 2023.

\bibitem[Soucek and Lokoc(2024)]{soucek2020transnet}
Tom{\'a}s Soucek and Jakub Lokoc.
\newblock Transnet v2: An effective deep network architecture for fast shot transition detection.
\newblock In \emph{Proceedings of the 32nd ACM International Conference on Multimedia}, pages 11218--11221, 2024.

\bibitem[Ravi et~al.(2024)Ravi, Gabeur, Hu, Hu, Ryali, Ma, Khedr, R{\"a}dle, Rolland, Gustafson, et~al.]{ravi2024sam2}
Nikhila Ravi, Valentin Gabeur, Yuan-Ting Hu, Ronghang Hu, Chaitanya Ryali, Tengyu Ma, Haitham Khedr, Roman R{\"a}dle, Chloe Rolland, Laura Gustafson, et~al.
\newblock Sam 2: Segment anything in images and videos.
\newblock \emph{arXiv preprint arXiv:2408.00714}, 2024.

\bibitem[Lin et~al.(2025)Lin, Chen, Liew, Chen, Li, Shi, Feng, and Kang]{lin2025depthanything3}
Haotong Lin, Sili Chen, Junhao Liew, Donny~Y Chen, Zhenyu Li, Guang Shi, Jiashi Feng, and Bingyi Kang.
\newblock Depth anything 3: Recovering the visual space from any views.
\newblock \emph{arXiv preprint arXiv:2511.10647}, 2025.

\bibitem[An et~al.(2026)An, Kupyn, Uscidda, Colaco, Ahuja, Belongie, Gonzalez-Franco, and Gazulla]{an2026vggrpo}
Zhaochong An, Orest Kupyn, Th{\'e}o Uscidda, Andrea Colaco, Karan Ahuja, Serge Belongie, Mar Gonzalez-Franco, and Marta~Tintore Gazulla.
\newblock Vggrpo: Towards world-consistent video generation with 4d latent reward.
\newblock \emph{arXiv preprint arXiv:2603.26599}, 2026.

\bibitem[Du et~al.(2026)Du, Ye, Cong, Li, Ni, Agarwal, Zhou, Li, Balestriero, and Wang]{du2026videogpa}
Hongyang Du, Junjie Ye, Xiaoyan Cong, Runhao Li, Jingcheng Ni, Aman Agarwal, Zeqi Zhou, Zekun Li, Randall Balestriero, and Yue Wang.
\newblock Videogpa: Distilling geometry priors for 3d-consistent video generation.
\newblock \emph{arXiv preprint arXiv:2601.23286}, 2026.

\bibitem[Oquab et~al.(2023)Oquab, Darcet, Moutakanni, Vo, Szafraniec, Khalidov, Fernandez, Haziza, Massa, El-Nouby, et~al.]{oquab2024dinov2}
Maxime Oquab, Timoth{\'e}e Darcet, Th{\'e}o Moutakanni, Huy Vo, Marc Szafraniec, Vasil Khalidov, Pierre Fernandez, Daniel Haziza, Francisco Massa, Alaaeldin El-Nouby, et~al.
\newblock Dinov2: Learning robust visual features without supervision.
\newblock \emph{arXiv preprint arXiv:2304.07193}, 2023.

\bibitem[Bai et~al.(2025)Bai, Cai, Chen, Chen, Chen, Cheng, Deng, Ding, Gao, Ge, et~al.]{Qwen3-VL}
Shuai Bai, Yuxuan Cai, Ruizhe Chen, Keqin Chen, Xionghui Chen, Zesen Cheng, Lianghao Deng, Wei Ding, Chang Gao, Chunjiang Ge, et~al.
\newblock Qwen3-vl technical report.
\newblock \emph{arXiv preprint arXiv:2511.21631}, 2025.

\bibitem[Seedance et~al.(2025)Seedance, Chen, Chen, Chen, Chen, Chen, Chen, Cheng, Cheng, Cheng, et~al.]{gao2025seedance15}
Team Seedance, Heyi Chen, Siyan Chen, Xin Chen, Yanfei Chen, Ying Chen, Zhuo Chen, Feng Cheng, Tianheng Cheng, Xinqi Cheng, et~al.
\newblock Seedance 1.5 pro: A native audio-visual joint generation foundation model.
\newblock \emph{arXiv preprint arXiv:2512.13507}, 2025.

\bibitem[Bansal et~al.(2025)Bansal, Peng, Bitton, Goldenberg, Grover, and Chang]{bansal2025videophy2}
Hritik Bansal, Clark Peng, Yonatan Bitton, Roman Goldenberg, Aditya Grover, and Kai-Wei Chang.
\newblock Videophy-2: A challenging action-centric physical commonsense evaluation in video generation.
\newblock \emph{arXiv preprint arXiv:2503.06800}, 2025.

\bibitem[{ACE Robotics}(2026)]{ace2026kairos}
{ACE Robotics}.
\newblock Kairos 3.0-4b: Real-time generative world model for embodied intelligence.
\newblock \url{https://github.com/kairos-agi/kairos-sensenova/tree/main}, 2026.

\bibitem[Dai et~al.(2025)Dai, Jiang, Wang, Xu, and Qi]{fantasyworld2025}
Yixiang Dai, Fan Jiang, Chiyu Wang, Mu~Xu, and Yonggang Qi.
\newblock Fantasyworld: Geometry-consistent world modeling via unified video and 3d prediction.
\newblock \emph{arXiv preprint arXiv:2509.21657}, 2025.

\bibitem[Team et~al.(2026{\natexlab{c}})Team, Shen, Zhang, Liu, Ji, Bao, Zhai, Liu, Guo, Wang, et~al.]{inspatio2026}
InSpatio Team, Donghui Shen, Guofeng Zhang, Haomin Liu, Haoyu Ji, Hujun Bao, Hongjia Zhai, Jialin Liu, Jing Guo, Nan Wang, et~al.
\newblock Inspatio-world: A real-time 4d world simulator via spatiotemporal autoregressive modeling.
\newblock \emph{arXiv preprint arXiv:2604.07209}, 2026{\natexlab{c}}.

\bibitem[Zhu et~al.(2025)Zhu, Feng, Zheng, Gao, Tao, Wan, Zhou, and Lu]{zhu2025astra}
Yixuan Zhu, Jiaqi Feng, Wenzhao Zheng, Yuan Gao, Xin Tao, Pengfei Wan, Jie Zhou, and Jiwen Lu.
\newblock Astra: General interactive world model with autoregressive denoising.
\newblock \emph{arXiv preprint arXiv:2512.08931}, 2025.

\bibitem[Wu et~al.(2026{\natexlab{b}})Wu, He, Cheng, Yang, Zhang, Kang, Cai, Wei, Guo, Li, et~al.]{wu2026infiniteworld}
Ruiqi Wu, Xuanhua He, Meng Cheng, Tianyu Yang, Yong Zhang, Zhuoliang Kang, Xunliang Cai, Xiaoming Wei, Chunle Guo, Chongyi Li, et~al.
\newblock Infinite-world: Scaling interactive world models to 1000-frame horizons via pose-free hierarchical memory.
\newblock \emph{arXiv preprint arXiv:2602.02393}, 2026{\natexlab{b}}.

\bibitem[Schuhmann et~al.(2022)Schuhmann, Beaumont, Vencu, Gordon, Wightman, Cherti, Coombes, Katta, Mullis, Wortsman, et~al.]{schuhmann2022laion5b}
Christoph Schuhmann, Romain Beaumont, Richard Vencu, Cade Gordon, Ross Wightman, Mehdi Cherti, Theo Coombes, Aarush Katta, Clayton Mullis, Mitchell Wortsman, et~al.
\newblock Laion-5b: An open large-scale dataset for training next generation image-text models.
\newblock \emph{Advances in neural information processing systems}, 35:\penalty0 25278--25294, 2022.

\bibitem[Ke et~al.(2021)Ke, Wang, Wang, Milanfar, and Yang]{ke2021musiq}
Junjie Ke, Qifei Wang, Yilin Wang, Peyman Milanfar, and Feng Yang.
\newblock Musiq: Multi-scale image quality transformer.
\newblock In \emph{Proceedings of the IEEE/CVF international conference on computer vision}, pages 5148--5157, 2021.

\bibitem[Li et~al.(2023)Li, Zhu, Han, Hou, Guo, and Cheng]{li2023amt}
Zhen Li, Zuo-Liang Zhu, Ling-Hao Han, Qibin Hou, Chun-Le Guo, and Ming-Ming Cheng.
\newblock Amt: All-pairs multi-field transforms for efficient frame interpolation.
\newblock In \emph{Proceedings of the IEEE/CVF Conference on Computer Vision and Pattern Recognition}, pages 9801--9810, 2023.

\end{thebibliography}
}

\appendix
\crefalias{section}{appendix}
\crefalias{subsection}{appendix}
\crefalias{subsubsection}{appendix}
\clearpage

\DoAppendixToC

\listoffigures
\listoftables

\clearpage

\section{Additional Dataset Statistics and Analysis}
\label{app:dataset}

\subsection{Extended Benchmark Comparison}
\label{app:benchmark_comparison_full}
\begin{table}[H]
\caption[Extended benchmark comparison]{Extended comparison of \benchmark with all surveyed benchmarks. FPP\thinslash TPP denotes first-\thinslash third-person perspective. Navi, SA, EE, and PS denote navigation, subject action, event editing, and perspective switching. Qual, Adh, Inter, Cons, and Phys denote video quality, setting adherence, interaction adherence, consistency, and physics compliance.}
\vspace{+3mm}
\label{tab:benchmark_comparison_full}
\centering
\renewcommand{\arraystretch}{1.3}
\footnotesize
\setlength{\tabcolsep}{3.5pt}
\begin{tabular}{@{}l c cc cccc ccccc rr@{}}
\toprule
\multirow{2.5}{*}{\textbf{Benchmark}} & \multirow{2.5}{*}{\textbf{Venue}} & \multicolumn{2}{c}{\textbf{Persp.}} & \multicolumn{4}{c}{\textbf{Interaction Type}} & \multicolumn{5}{c}{\textbf{Evaluation Dimension}} & \multicolumn{2}{c}{\textbf{Scale}} \\
\cmidrule(lr){3-4} \cmidrule(lr){5-8} \cmidrule(lr){9-13} \cmidrule(l){14-15}
& & \textbf{FPP} & \textbf{TPP} & \textbf{Navi} & \textbf{SA} & \textbf{EE} & \textbf{PS} & \textbf{Qual} & \textbf{Adh} & \textbf{Inter} & \textbf{Cons} & \textbf{Phys} & \textbf{Cases} & \textbf{Turns} \\
\midrule
VBench~\cite{huang2024vbench}                     & \pub{CVPR'24}    & - & - & \xmark & \xmark & \xmark & \xmark & \cmark & \xmark & \xmark & \xmark & \xmark & 946   & 946   \\
EvalCrafter~\cite{liu2024evalcrafter}             & \pub{CVPR'24}    & - & - & \xmark & \xmark & \xmark & \xmark & \cmark & \xmark & \xmark & \xmark & \xmark & 700   & 700   \\
WorldSimBench~\cite{lian2024worldsimbench}        & \pub{arXiv'24}   & \cmark & \xmark & \cmark & \cmark & \xmark & \xmark & \xmark & \xmark & \cmark & \cmark & \xmark & 35k   & 35k   \\
PhyGenBench~\cite{meng2024phygenbench}            & \pub{ICML'25}    & - & - & \xmark & \xmark & \xmark & \xmark & \cmark & \xmark & \xmark & \xmark & \cmark & 160   & 160   \\
VBench++~\cite{huang2024vbenchplus}               & \pub{TPAMI'25}   & - & - & \xmark & \xmark & \xmark & \xmark & \cmark & \xmark & \xmark & \xmark & \cmark & 1,260 & 1,260 \\
Video-Bench~\cite{han2025videobench}              & \pub{arXiv'25}   & - & - & \xmark & \xmark & \xmark & \xmark & \cmark & \xmark & \xmark & \xmark & \xmark & 419   & 419   \\
VideoPhy-2~\cite{bansal2025videophy2}             & \pub{arXiv'25}   & - & - & \xmark & \xmark & \xmark & \xmark & \cmark & \xmark & \xmark & \xmark & \cmark & 590   & 590   \\
WorldScore~\cite{duan2024worldscore}              & \pub{ICCV'25}    & \cmark & \xmark & \cmark & \xmark & \cmark & \xmark & \cmark & \cmark & \xmark & \xmark & \xmark & 3,000 & 3,000 \\
WorldModelBench~\cite{wang2024worldmodelbench}    & \pub{NeurIPS'25} & \cmark & - & \cmark & \cmark & \xmark & \xmark & \xmark & \cmark & \cmark & \xmark & \cmark & 350   & 350   \\
4DWorldBench~\cite{lu2025_4dworldbench}           & \pub{arXiv'25}   & - & - & \xmark & \xmark & \xmark & \xmark & \cmark & \xmark & \xmark & \xmark & \cmark & 333   & 333   \\
InterBench~\cite{tang2025interbench}              & \pub{arXiv'25}   & \cmark & \cmark & \xmark & \cmark & \cmark & \xmark & \cmark & \xmark & \cmark & \cmark & \cmark & 920   & 920   \\
MIND~\cite{ye2026mind}                            & \pub{arXiv'26}   & \cmark & - & \cmark & \xmark & \xmark & \xmark & \cmark & \xmark & \cmark & \cmark & \xmark & 250   & --    \\
WorldArena~\cite{shang2026worldarena}             & \pub{arXiv'26}   & \cmark & \xmark & \cmark & \cmark & \xmark & \xmark & \cmark & \xmark & \cmark & \cmark & \cmark & 500   & 500   \\
Omni-WorldBench~\cite{wu2026omniworldbench}       & \pub{arXiv'26}   & \cmark & - & \cmark & \cmark & \cmark & \xmark & \cmark & \xmark & \cmark & \cmark & \cmark & 1,068 & 1,068 \\
WR-Arena~\cite{gao2026wrena}                      & \pub{arXiv'26}   & \cmark & \xmark & \cmark & \cmark & \xmark & \xmark & \xmark & \xmark & \cmark & \xmark & \cmark & 62    & $\leq$558 \\
WorldLens~\cite{worldlens2025}                    & \pub{CVPR'26}    & \cmark & \xmark & \cmark & \xmark & \xmark & \xmark & \cmark & \cmark & \cmark & \cmark & \cmark & 26k   & --   \\
World-in-World~\cite{zhang2025worldinworld}       & \pub{ICLR'26}    & \cmark & \xmark & \cmark & \cmark & \xmark & \xmark & \cmark & \xmark & \cmark & \xmark & \xmark & 1,079 & --    \\
WorldMark~\cite{xu2026worldmark}                  & \pub{arXiv'26}   & \cmark & \cmark & \cmark & \xmark & \xmark & \xmark & \cmark & \xmark & \cmark & \cmark & \xmark & 500   & --    \\
\midrule
\textbf{\benchmark (Ours)}                        & --        & \cmark & \cmark & \cmark & \cmark & \cmark & \cmark & \cmark & \cmark & \cmark & \cmark & \cmark & \numvideo  & \numturn \\
\bottomrule
\end{tabular}
\end{table}

\cref{tab:benchmark_comparison_full} summarizes the full set of surveyed benchmarks. Representative examples span quality-oriented suites such as VBench~\cite{huang2024vbench}, which dissects text-to-video quality into 16 hierarchical dimensions with tailored prompts, and EvalCrafter~\cite{liu2024evalcrafter}, which scores around 700 prompts with 17 objective metrics calibrated against user preferences. Physics-focused work includes PhyGenBench~\cite{meng2024phygenbench}, with 160 prompts across 27 physical laws in four domains, and VideoPhy-2~\cite{bansal2025videophy2}, which targets action-centric physics over 200 real-world actions. WorldScore~\cite{duan2024worldscore} casts world generation as a sequence of next-scene tasks along explicit camera trajectories on 3,000 cases, while WorldModelBench~\cite{wang2024worldmodelbench} evaluates instruction following and physics adherence using 67K human labels in robotics and driving settings. Among interactive efforts, WorldSimBench~\cite{lian2024worldsimbench} pairs perceptual evaluation with a manipulative evaluation that checks whether generated videos translate into correct control signals, World-in-World~\cite{zhang2025worldinworld} builds a closed-loop platform with a unified planner and standardized action API that reports task success as the primary metric, and WorldMark~\cite{xu2026worldmark} introduces a WASD-style action-mapping layer to enable apples-to-apples comparison across interactive video world models under identical scenes and trajectories.

\subsection{Dataset Gallery}
\label{app:visual_showcase}

\begin{figure}[H]
    \centering
    \includegraphics[width=\linewidth,height=0.95\textheight,keepaspectratio]{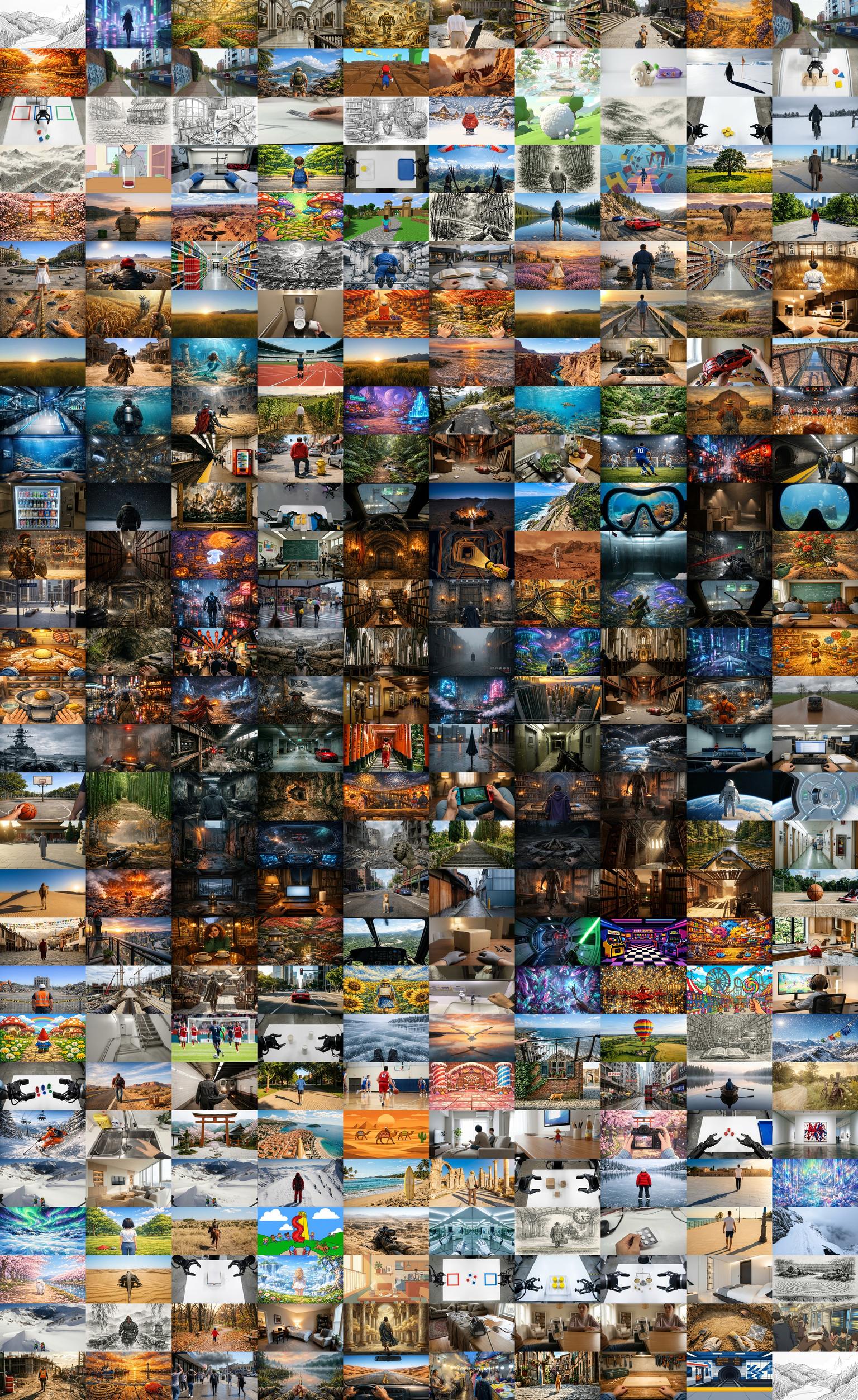}
    \caption[Thumbnail gallery of all cases]{Thumbnail gallery of all cases in \benchmark.}
    \label{fig:all_gallery}
\end{figure}

This section provides qualitative showcases of \benchmark along the taxonomy axes introduced in Section~\ref{sec:ontology}. We show the thumbnails of all the initial frames of the \benchmark in \cref{fig:all_gallery}.

\subsubsection{Scene and Style Gallery}
\label{app:scene_style_gallery}

\begin{figure}[H]
\centering
\setlength{\fboxsep}{0pt}
\includegraphics[width=0.98\textwidth]{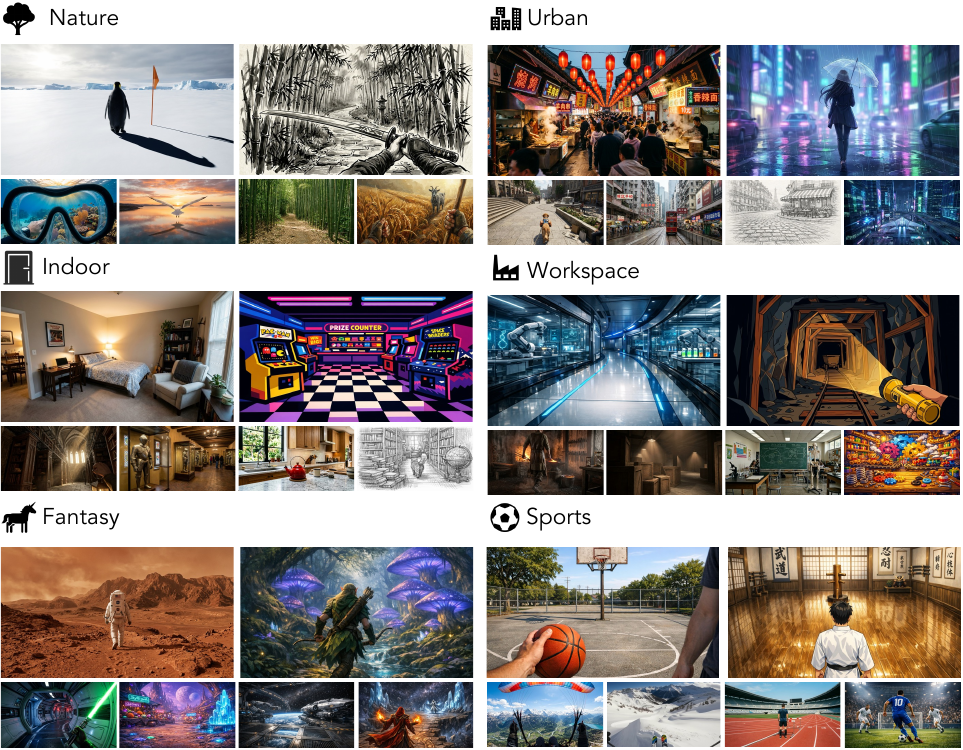}
\caption[Scene and style coverage]{Scene and style coverage. Two categories per row are presented, each shown as a photorealistic/stylized pair that shares the same underlying scene specification. The six categories cover nature, urban, indoor, workspace, fantasy, and sports/game.}
\label{fig:scene_style_matrix}
\end{figure}

\begin{figure}[H]
\centering
\setlength{\fboxsep}{0pt}
\includegraphics[width=0.98\textwidth]{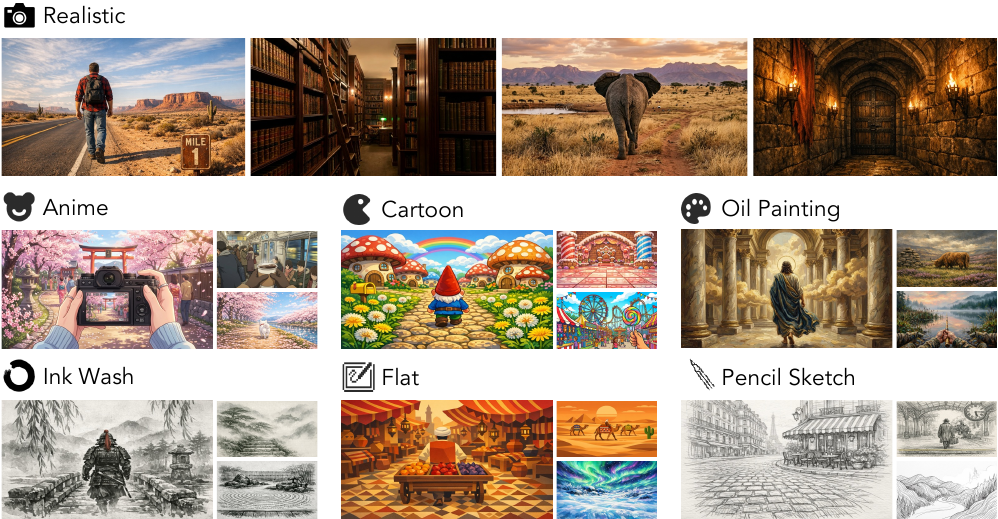}
\caption[Style gallery]{Style gallery. Representative initial frames spanning the rendering styles covered by \benchmark: realistic, anime, cartoon, oil painting, ink wash, flat, and pencil sketch.}
\label{fig:style_gallery}
\end{figure}

\cref{fig:scene_style_matrix} displays two photorealistic and stylized pairs per row, covering all six scene categories in a compact three-row layout. The six categories, nature, urban, indoor, workspace, fantasy, and sports or game, span both naturalistic settings (outdoor landscapes, cityscapes, living spaces) and more constructed or speculative environments (office or workshop, surreal or magical worlds, and curated activity spaces), and each category is shown through a photorealistic and a stylized rendering of the same underlying scene specification, so that readers can compare how the rendering pipeline preserves the scene semantics while varying the visual appearance. \cref{fig:style_gallery} further enlarges the stylized rendering to show the full range of visual styles supported by \benchmark, including realistic, anime, cartoon, oil painting, ink wash, flat, and pencil sketch, which together cover photographic, illustrative, painterly, and sketch-like appearances used by practical content-creation workflows.

\subsubsection{Perspective and Subject Gallery}
\label{app:persp_subject_gallery}

\begin{figure}[H]
\centering
\setlength{\fboxsep}{0pt}
\includegraphics[width=0.996\textwidth]{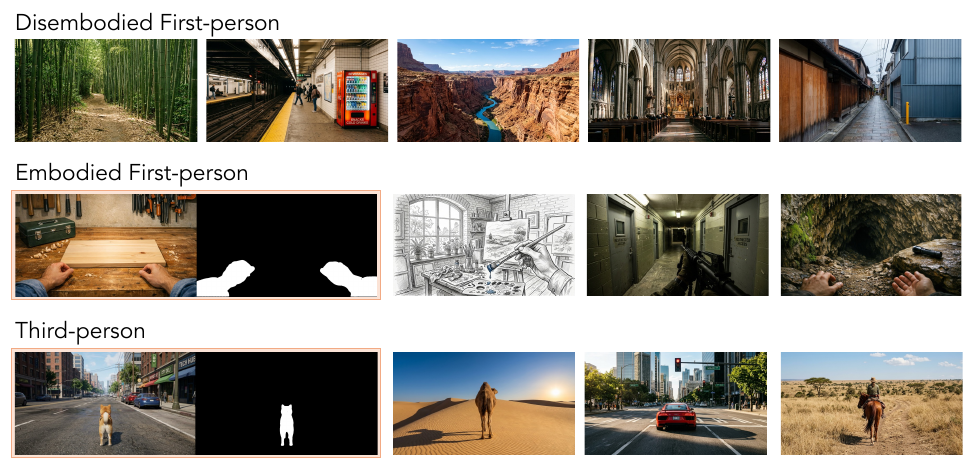}
\caption[Perspective gallery]{Perspective gallery. Cases are grouped into three rows by perspective type: disembodied first-person with no visible agent, embodied first-person with a visible body part such as hands or weapon, and third-person with the controlled subject in view.}
\label{fig:perspective_gallery}
\end{figure}

\begin{figure}[H]
\centering
\setlength{\fboxsep}{0pt}
\includegraphics[width=0.996\textwidth]{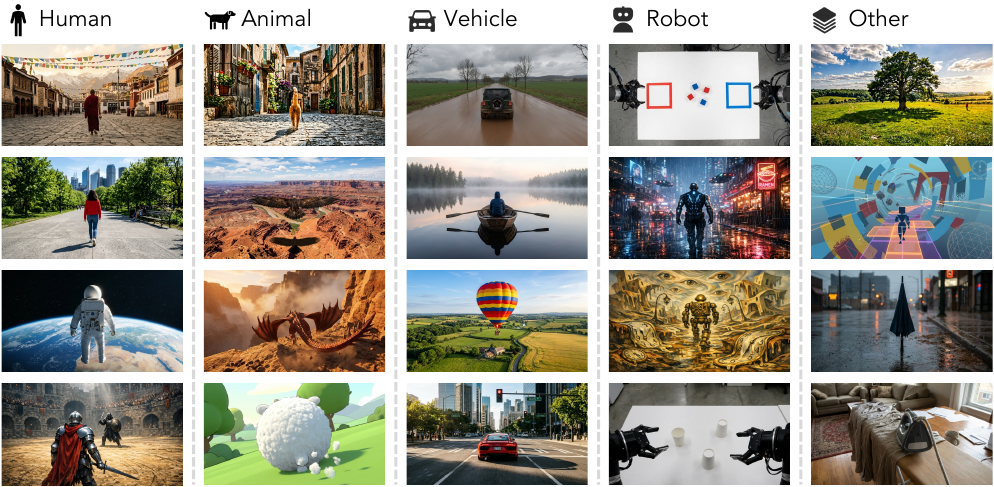}
\caption[Subject gallery]{Subject gallery. One column per subject category, covering human, animal, vehicle, robot, and other objects. Each column lists multiple representative cases that span different scenes and styles.}
\label{fig:subject_gallery}
\end{figure}

\cref{fig:perspective_gallery} separates the perspective axis into three rows. The top row shows \textit{disembodied first-person} cases, where the camera is attached to an implicit agent and no body part is visible, so that perspective is defined purely by the camera's position and motion. The middle row shows \textit{embodied first-person} cases, where a body part (typically hands, held tools, or a mounted weapon) is visible at the bottom of the frame and grounds the perspective in a specific agent. The bottom row shows \textit{third-person} cases, where the controlled subject is visible inside the frame and the camera follows it externally. This three-way split matters because embodiment changes how navigation and subject-action prompts should be interpreted: in disembodied first-person, instructions like ``walk forward'' collapse to camera translation, whereas in embodied first- and third-person settings the same instruction must be grounded in a specific agent's motion. \cref{fig:subject_gallery} further enumerates the five controllable-subject categories on the third-person axis, namely human, animal, vehicle, robot, and other objects, each illustrated by multiple representative cases that vary scene and style, so that subject diversity is shown to be orthogonal to scene and style diversity rather than confounded with them.

\subsubsection{Subject Action and Event Editing Examples}
\label{app:sa_ee_examples}

Unlike scenes, styles, perspectives, and subjects, which are naturally illustrated by static initial frames, subject action and event editing are fundamentally temporal phenomena and are therefore hard to render compactly as image galleries. In place of such galleries, \cref{tab:sa_ee_examples} lists condensed example phrases abstracted from the released \benchmark cases, so that reviewers can directly read off the semantic breadth each sub-type aims to cover. On the Subject Action side, the five sub-types range from fine-grained hand-object interaction (manipulation, tool use) to whole-body movement and agent--environment physics (locomotion, combat), and further to socially grounded behaviors that carry communicative intent rather than physical effect (gestural interaction). On the Event Editing side, the six sub-types progressively move outward from the subject: appearance or state changes that attach to existing objects, environmental changes and natural phenomena that reshape global scene attributes, NPC motion that introduces new dynamic agents, mechanical transitions that reconfigure scene geometry, and physical effects that trigger irreversible state changes such as explosions or collapses. Together the eleven sub-types are designed to stress-test different aspects of world-model behavior, from local controllability over a subject to global responsiveness of the surrounding world.

\begin{table}[H]
\centering
\small
\renewcommand{\arraystretch}{1.2}
\setlength{\tabcolsep}{4pt}
\caption[Subject action and event editing example phrases]{Condensed example phrases for each Subject Action and Event Editing sub-type, abstracted from released \benchmark cases. Each cell lists three semicolon-separated phrases that cover the typical semantic range of the subtype.}
\vspace{+1mm}
\begin{tabular}{@{}p{0.22\linewidth}p{0.74\linewidth}@{}}
\toprule
\textbf{Sub-type} & \textbf{Example phrases} \\
\midrule
\multicolumn{2}{@{}l}{\textit{Subject Action}} \\
\midrule
Manipulation           & open a box lid; knead dough; sweep leaves into a pile. \\
Tool use               & saw a plank along a line; light a candle with a match; trim branches with shears. \\
Locomotion             & glide on ice; dive and descend underwater; squeeze through a narrow cave. \\
Combat                 & downward sword slash; straight punch at a dummy; axe strike on a door. \\
Gestural interaction   & expressive hand-gesture speech; kneel before a herald; pat a horse. \\
\midrule
\multicolumn{2}{@{}l}{\textit{Event Editing}} \\
\midrule
Environment change        & rain begins; sunset turns to night; fog rolls in. \\
Appearance state change   & runes glow cyan; armor visor lights up; window fogs from steam. \\
NPC motion                & a passer-by walks by; a fish school drifts past; a dog dashes out. \\
Mechanical transition     & drawbridge lowers; gate slides up; elevator doors open. \\
Physical effect           & canister explodes; building collapses with dust; ice cracks and water floods in. \\
Natural phenomenon        & volcano erupts; sandstorm approaches; hot-air balloons rise at sunset. \\
\bottomrule
\end{tabular}

\label{tab:sa_ee_examples}
\end{table}

\begin{figure}[t]
\centering
\setlength{\fboxsep}{0pt}
\includegraphics[width=0.8\textwidth]{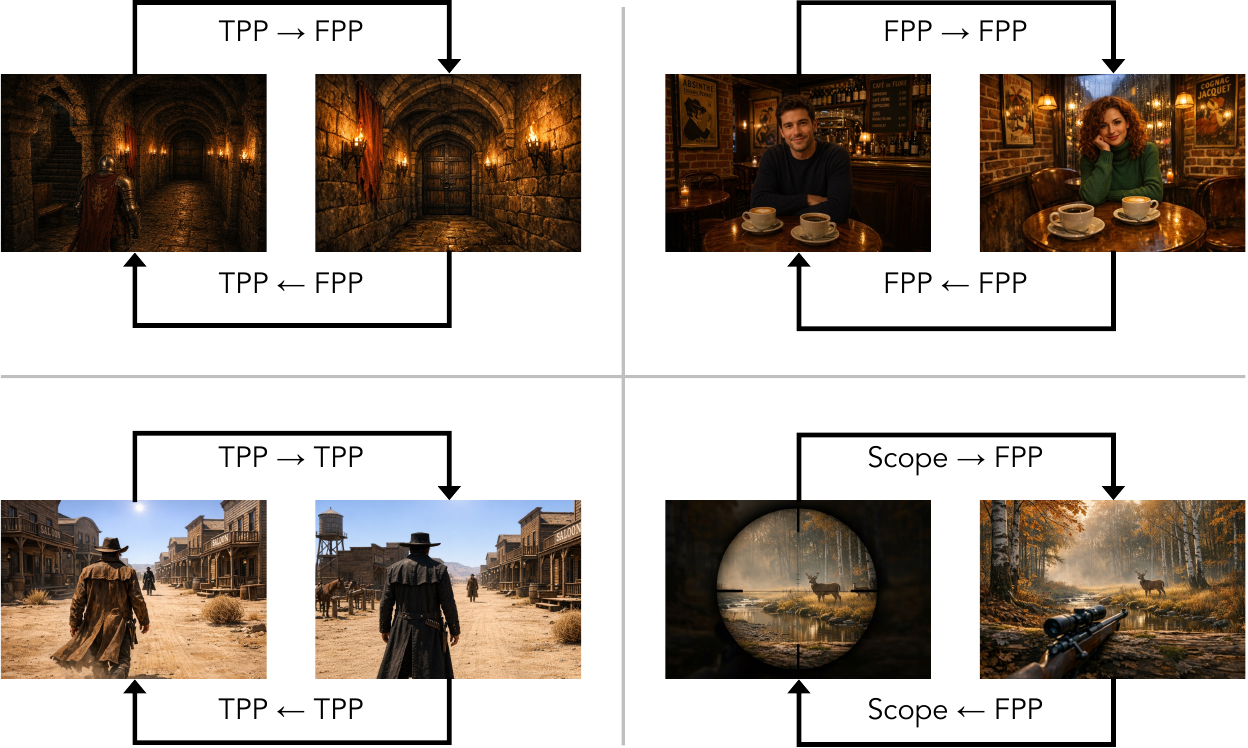}
\caption[Perspective-switching taxonomy showcase]{Perspective-switching taxonomy showcase. One representative case per sub-type, covering same-subject switches, multi-subject switches, and scope-mode transitions. The source perspective and the switching prompt are rendered inside each frame.}
\label{fig:persp_switch_showcase}
\end{figure}

\subsubsection{Perspective Switching Prompts Gallery}
\label{app:persp_switch_gallery}

\cref{fig:persp_switch_showcase} illustrates the three perspective-switching sub-types in a single row. Each panel shows the source perspective as the initial frame, with the released switching prompt baked in to describe the intended target perspective, so that the interaction shown is exactly the one the model is asked to execute at evaluation time. \textit{Same-subject} switches change the camera relative to the same controlled subject, for example from a third-person follow shot to an embodied first-person view of the same character, probing whether the model can re-anchor the camera without losing identity or pose. \textit{Multi-subject} switches hand control over from one subject to another within the same scene, so that the new perspective is attached to a different agent while the surrounding scene must remain consistent. \textit{Scope mode} transitions change the spatial scope of the perspective itself, for example zooming from a narrow first-person view out to a broader third-person shot or vice versa, testing whether the model maintains a coherent world across scale changes. The three sub-types together cover the principal ways a user can reconfigure perspective at run time without regenerating the world from scratch.

\subsection{Navigation Design and Distribution}
\label{app:nav_coverage}

Navigation is the most common interaction type in \benchmark and receives the most systematic treatment. This section details the action definition, distribution statistics, and trajectory design.

\begin{figure}[t]
\centering
\setlength{\fboxsep}{0pt}
\includegraphics[width=0.996\textwidth]{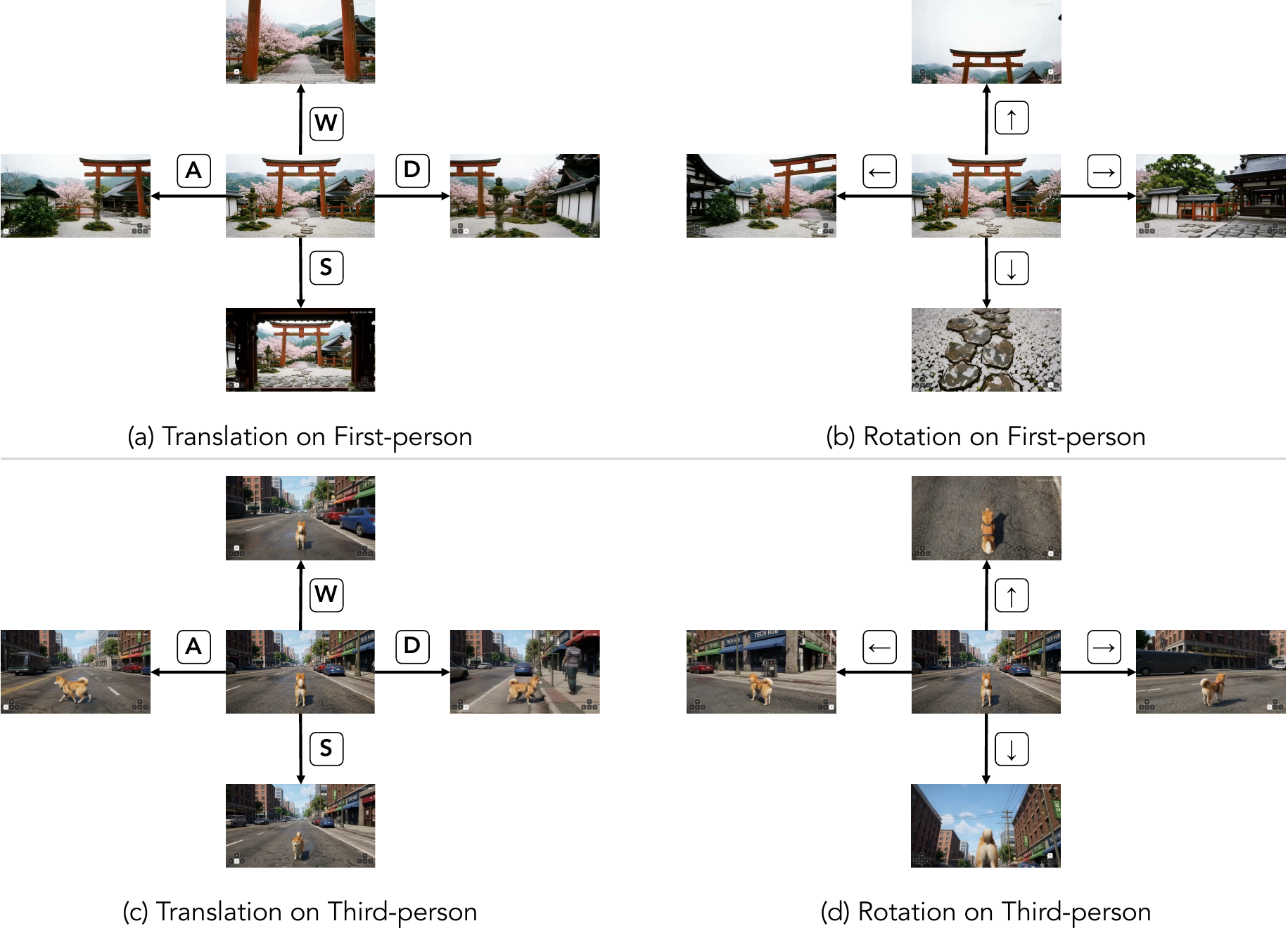}
\caption[Navigation action definition]{Navigation action definition. Illustration of how each WASD and arrow key maps to a physical motion under first-person and third-person perspectives. Frames following the navigation definition panel are from Happy Oyster~\cite{alibaba2026happyoyster}.}
\label{fig:nav_define}
\end{figure}

\paragraph{Action definition.}
\benchmark adopts a WASD plus arrow-key scheme for discrete navigation control. As shown in \cref{fig:nav_define} and \cref{tab:nav_actions}, the same key triggers different physical motions depending on the perspective. Under first-person view, W/S/A/D translate the camera forward, backward, left, and right, while arrow keys rotate the viewpoint. Under third-person view, W/S/A/D move the subject, while arrow keys orbit the camera around the subject. This perspective-dependent mapping mirrors how game engines handle first- and third-person control, and ensures that each key carries a clear, unambiguous spatial semantics for evaluation.

\begin{figure}[H]
\centering
\includegraphics[width=0.7\linewidth]{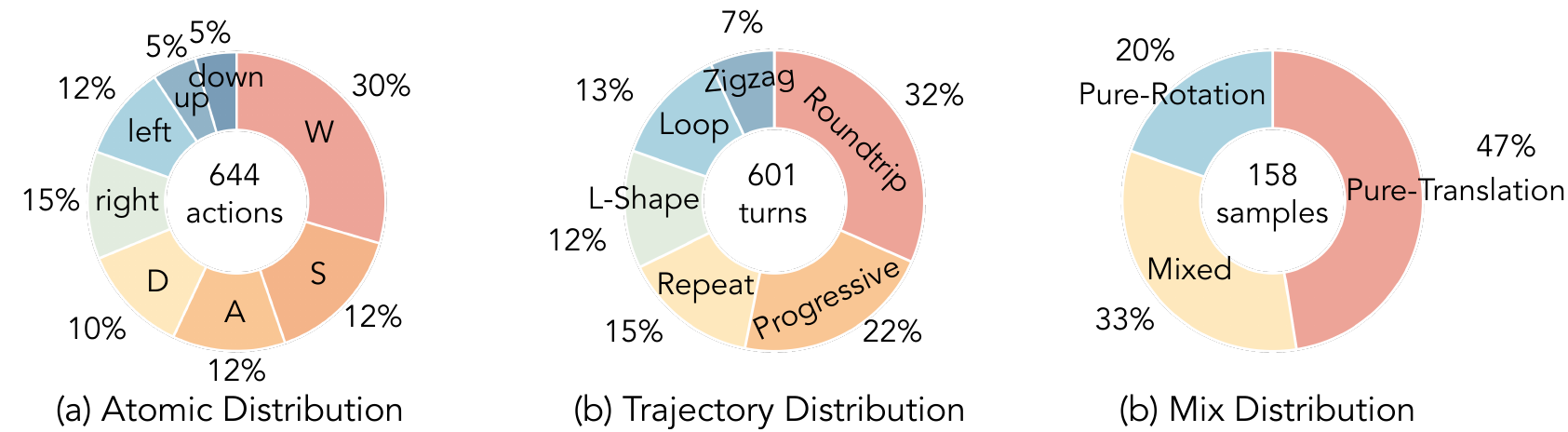}
\caption[Navigation test case distribution]{Distribution of navigation test cases across direction, scene type, and control interface.}
\label{fig:navi_distribute}
\end{figure}

\paragraph{Atomic action distribution.}
\cref{fig:navi_distribute} shows the atomic action distribution across all navigation sequences. Translational actions (W/S/A/D) account for 62.8\% and rotational actions (arrow keys) for 37.0\%, ensuring balanced coverage of both movement families. Among translational actions, forward motion (W, 29\%) is the most frequent, reflecting its prevalence in natural navigation, while lateral (A/D, 12\% each) and backward (S, 11\%) motions are evenly represented. Rotational actions are distributed across left (10\%), right (12\%), up (10\%), and down (6\%), with downward rotation being the least common as it is rarely the primary navigation intent.

\paragraph{Trajectory type design.}
We categorize navigation sequences into six trajectory types, each targeting different aspects of spatial understanding. \cref{tab:trajectory_examples} lists representative action sequences for each type.
\textbf{Round-trip} trajectories move forward and then retrace the path, directly testing whether the model can maintain spatial consistency when revisiting previously seen regions.
\textbf{Progressive} trajectories combine multiple distinct directions without returning, requiring the model to coherently extend the scene into new areas.
\textbf{Repeat} trajectories apply the same action consecutively, testing whether the model sustains smooth and consistent motion over extended sequences.
\textbf{L-shape} trajectories execute a single sharp turn, probing the model's ability to handle abrupt direction changes while preserving scene geometry.
\textbf{Loop} trajectories traverse a closed path that returns to the starting area via a different route, demanding global spatial coherence across the entire sequence.
\textbf{Zigzag} trajectories alternate between opposing directions, challenging the model to maintain stable scene structure under rapid, repeated directional switches.
Among these, round-trip trajectories are the most frequent (32\%) as they directly enable spatial consistency evaluation through revisitation, followed by Progressive (22\%), Repeat (15\%), and Loop (13\%). L-Shape and Zigzag make up the remaining.

\paragraph{Overall navigation distribution.}
\cref{fig:navi_distribute} shows the distribution of navigation test cases across direction, scene type, and control interface. The direction distribution covers all eight cardinal and diagonal directions, ensuring that the benchmark does not bias toward any particular movement axis. Scene types span indoor, outdoor, and fantasy environments, providing diverse spatial contexts. Control interfaces include text, 6DoF-pose, and discrete-action inputs, enabling evaluation of models with different native navigation interfaces under comparable conditions.

\begin{table}[H]
\caption[Navigation action semantics]{Navigation action semantics under different perspectives. The same key triggers different physical motions depending on whether the case is first-person or third-person.}
\vspace{+1mm}
\label{tab:nav_actions}
\centering
\small
\renewcommand{\arraystretch}{1.1}
\setlength{\tabcolsep}{6pt}
\begin{tabular}{@{}llll@{}}
\toprule
\textbf{Type} & \textbf{Key} & \textbf{First-Person} & \textbf{Third-Person} \\
\midrule
\multirow{4}{*}{Translation}
  & W & Camera pushes forward  & Subject walks forward \\
  & S & Camera pulls backward  & Subject steps backward \\
  & A & Camera strafes left    & Subject moves left \\
  & D & Camera strafes right   & Subject moves right \\
\midrule
\multirow{4}{*}{Rotation}
  & $\leftarrow$  & View turns left  & Camera orbits left \\
  & $\rightarrow$ & View turns right & Camera orbits right \\
  & $\uparrow$    & View tilts up    & Camera elevates \\
  & $\downarrow$  & View tilts down  & Camera descends \\
\bottomrule
\end{tabular}
\end{table}

\begin{table}[H]
\centering
\caption[Trajectory type examples]{Trajectory type examples in navigation test cases.}
\label{tab:trajectory_examples}
\renewcommand{\arraystretch}{1.05}
\vspace{+1mm}
\begin{tabular}{ll}
\toprule
Trajectory Type & Examples \\
\midrule
Round-trip & {[A, A, D, D]}, {[left, left, right, right]}, {[W, W, W, S, S, S]} \\
Progressive & {[W, right, right, left]}, {[W, A, W, D]}, {[W, W, left, left, right, right]} \\
Repeat & {[W, W, W, W]}, {[W, W, W]} \\
L-shape & {[W, W, A, A]} \\
Loop & {[W, A, S, D]}, {[A, A, W, S, D, D]} \\
Zigzag & {[right, W, right, W]}, {[W+A, W+D, W+A, W+D]}, {[up, down, up, down]} \\
\bottomrule
\end{tabular}
\renewcommand{\arraystretch}{1}
\end{table}

\clearpage
\section{Evaluated Models}
\label{app:models}

\subsection{Per-Model Configuration}
\label{app:model_details}

\begin{table}[H]
\caption[Detailed overview of evaluated models]{Detailed overview of all \nummodel evaluated models. ``Params'' = parameter count (--- if undisclosed). ``Frames'' = number of frames generated per interaction turn (RT = real-time streaming). ``Inference : Video'' is the ratio of wall-clock generation time to the duration of the produced video, lower is faster.}
\vspace{+1mm}
\label{tab:evaluated_models}
\centering
\renewcommand{\arraystretch}{1.1}
\footnotesize
\setlength{\tabcolsep}{3pt}
\begin{tabular}{@{}lclclrcl@{}}
\toprule
\textbf{Model} & \textbf{Access} & \textbf{Architecture} & \textbf{Params} & \textbf{Resolution} & \textbf{Frames} & \textbf{Inference : Video} & \textbf{Hardware / Note} \\
\midrule

\multicolumn{8}{@{}l}{\textit{Text-driven models}} \\
\addlinespace[2pt]
Seedance 1.5~\cite{gao2025seedance15}        & API    & -             & ---  & 1280$\times$720  & 120 & ---     & API \\
Wan 2.7~\cite{wan2025wan27}                  & API    & -             & ---  & 1920$\times$1080 & 80  & ---     & API \\
Kling 3.0~\cite{kuaishou2025kling3}          & API    & -             & ---  & 1280$\times$720  & 120 & ---     & API \\
YUME 1.5~\cite{mao2025yume15}                & Open   & DiT             & 5B   & 1280$\times$704  & 58  & $1.7$   & A100 80G \\
HY-Video 1.5~\cite{kong2024hunyuanvideo}     & Open   & DiT             & 8B   & 848$\times$480   & 120 & ---     & A100 80G \\
LTX 2.3~\cite{lightricks2025ltx}             & Open   & DiT      & 22B  & 1536$\times$1024 & 120 & ---     & A100 80G \\
LongCat~\cite{cai2025longcat}                & Open   & DiT             & 14B  & 832$\times$480   & 48  & $621.8$ & A100 80G \\
Kairos 3.0~\cite{ace2026kairos}              & Open   & DiT             & 4B   & 832$\times$480   & 96  & ---     & A100 80G \\
Cosmos 2.5~\cite{nvidia2025cosmos}           & Open   & DiT             & 2B   & 1280$\times$704  & 93  & ---     & A100 80G \\

\addlinespace[4pt]
\multicolumn{8}{@{}l}{\textit{Camera-controlled models}} \\
\addlinespace[2pt]
LingBot-World~\cite{gao2026lingbot}          & Open   & Multi-stage DiT & 14B  & 1280$\times$704  & 32  & $268.8$ & A100 80G \\
HY-World 1.5~\cite{sun2025hyworldplay}       & Open   & Streaming DiT   & 8B   & 832$\times$480   & 96  & $18.8$  & A100 80G \\
Fantasy-World~\cite{fantasyworld2025}        & Open   & DiT + 3D head   & 14B  & 592$\times$336   & 80  & $2320$  & A100 80G \\
InSpatio-World~\cite{inspatio2026}           & Open   & V2V AR DiT      & 1.3B & 832$\times$480   & 96  & ---     & A100 80G \\
Astra~\cite{zhu2025astra}                    & Open   & AR DiT  & 2B   & 832$\times$480   & 80  & $30.0$  & A100 80G \\

\addlinespace[4pt]
\multicolumn{8}{@{}l}{\textit{Action-conditioned models}} \\
\addlinespace[2pt]
Happy Oyster~\cite{alibaba2026happyoyster}   & Closed & ---             & ---  & 1280$\times$720  & RT  & ---     & Web interface \\
Matrix-Game 3.0~\cite{he2026matrixgame3}     & Open   & AR Diffusion    & 6B   & 1280$\times$704  & 70  & $0.5$   & A100 80G \\
Genie 3~\cite{ball2025genie3}                & Closed & ---  & ---  & 1280$\times$720  & RT  & ---     & Web interface \\
Matrix-Game 2.0~\cite{he2025matrixgame2}     & Open   & AR Diffusion    & 2B   & 640$\times$352   & 48  & $5.6$   & A100 80G \\
HY-GameCraft~\cite{li2025hunyuangamecraft}   & Open   & Hist-Cond DiT   & 13B  & 832$\times$480   & 132 & $2.5$   & A100 80G \\
Infinite-World~\cite{wu2026infiniteworld}    & Open   & Hierarchical DiT & 1.5B & 896$\times$448   & 160 & $49.2$  & A100 80G \\

\bottomrule
\end{tabular}
\end{table}

\subsubsection{Text-driven Models}

Since these models do not natively accept action signals, we adopt an iterative I2V protocol: for each interaction turn, we extract the last frame of the previous clip, construct a text prompt that combines the world-setting description with the current interaction instruction, and generate the next video segment.

\myparagraph{Seedance~1.5~\cite{gao2025seedance15}.}
Seedance~1.5 is ByteDance's commercial video generation model, accessed through a closed-source API.
It supports image-conditioned generation at up to 1280$\times$720 resolution and 120 frames per clip.
In \benchmark, it is evaluated under the iterative I2V protocol, with interaction instructions injected as text prompts.

\myparagraph{Wan~2.7~\cite{wan2025wan27}.}
Wan~2.7 is a publicly released checkpoint in the open-source Wan model family, developed by Alibaba.
It supports multi-resolution video synthesis at up to 1280$\times$720 and 80 frames per clip, with full open weights available for reproducible research.
In \benchmark, it is evaluated under the same iterative I2V protocol.

\myparagraph{Kling~3.0~\cite{kuaishou2025kling3}.}
Kling~3.0 is Kuaishou's commercial video generation model, accessed through a closed-source API.
It supports image-conditioned generation with interaction instructions provided as natural-language prompts.
In \benchmark, it is evaluated under the iterative I2V protocol.

\myparagraph{YUME~1.5~\cite{mao2025yume15}.}
YUME~1.5 is the first fully open-source interactive world model to support the combined capabilities of text-to-world generation, image-to-world initialization, and text-driven event editing, all within a single Diffusion Transformer.
For \benchmark, we use the released \texttt{Yume-5B-720P} checkpoint, which contains approximately $5.2$B parameters.
Unlike keyboard-driven world models, YUME~1.5 uses natural language to specify both the scene content and the desired interaction, making it the only IWM in this evaluation that shares the same text-based control interface as the general I2V models.
Its training combines large-scale text-video paired data with curated interactive gameplay footage, enabling the model to ground language descriptions in plausible world dynamics.
Each turn is assembled by concatenating fixed-size $29$-frame iterations, with the number of iterations per turn chosen to best approximate the target turn duration.

\myparagraph{HunyuanVideo~1.5~\cite{kong2024hunyuanvideo}.}
HunyuanVideo~1.5 is Tencent's open-source video generation model based on a causal 3D full-attention DiT, pre-trained on a large-scale video and image corpus.
The released 480p I2V checkpoint contains roughly $8.3$B transformer parameters.
Its architecture jointly models spatial and temporal attention without factorized approximations, enabling high visual fidelity and strong temporal coherence across long clips.
The model supports image-conditioned generation at 848$\times$480 and 120 frames, making it well-suited for multi-turn evaluation where each turn requires consistent subject appearance and background layout.

\myparagraph{LTX~2.3~\cite{lightricks2025ltx}.}
LTX~2.3 is a lightweight 2B-parameter latent Diffusion Transformer developed by Lightricks, optimized for real-time or near-real-time inference on consumer hardware.
It leverages a highly compressed latent space with a spatiotemporal VAE and a shallow attention stack to achieve fast sampling with minimal quality degradation.
While its parameter count is an order of magnitude smaller than other I2V models in this evaluation, LTX~2.3 represents an important operating point for deployment-constrained scenarios, and we include it to assess the trade-off between model scale and interaction adherence.

\myparagraph{LongCat-Video~\cite{cai2025longcat}.}
LongCat-Video is a DiT model developed by Meituan, explicitly designed for long-video generation via a native video-continuation mechanism.
Its released DiT transformer has approximately $13.6$B parameters.
Unlike standard I2V models that generate each clip independently, LongCat-Video conditions each new segment on a sliding window of the most recent frames encoded into a latent prefix, which provides short-horizon temporal context beyond the last conditioning frame.
This architectural choice makes it a natural baseline for multi-turn evaluation, as segment-level continuity is directly supported by the model design rather than enforced only by the evaluation protocol.

\myparagraph{Kairos~3.0~\cite{ace2026kairos}.}
Kairos is a DiT-based video generation model from ACE Robotics, originally developed to support embodied intelligence applications that require egocentric video synthesis conditioned on navigation commands.
For \benchmark, we use the officially released \texttt{kairos-common-4B-720P-16fps} checkpoint, which contains roughly $4.05$B parameters.
Its training pipeline incorporates large volumes of first-person footage collected from both simulated and real-world environments, endowing the model with a strong prior over forward-facing camera dynamics.
We evaluate Kairos under the standard iterative I2V protocol using text prompts, treating it as a general I2V baseline that benefits from egocentric pre-training.

\myparagraph{Cosmos-Predict~2.5~\cite{nvidia2025cosmos}.}
Cosmos~2.5 is NVIDIA's world foundation model, pre-trained on a large-scale mixture of real-world video, physical simulation output, and synthetic robotics demonstrations.
For \benchmark, we evaluate the Cosmos-Predict2.5 2B post-trained variant (\texttt{2B/post-trained}) in image-to-world mode, which generates $93$ frames per turn at $1280\times704$ resolution.
The Cosmos platform is designed as a general-purpose foundation for physical AI, with a Diffusion Transformer backbone that supports world-consistent video generation conditioned on text, images, or structured action specifications.
Although Cosmos is primarily positioned for robotics and autonomous-driving applications, we include it as a general-purpose I2V baseline given its large-scale pre-training and publicly available weights.

\subsubsection{Camera-controlled Models}

These models natively accept camera-control inputs in the form of 6DoF camera parameter matrices.
For each interaction turn, we provide the model with the last generated frame and a camera trajectory, represented as a sequence of 6DoF camera matrices derived from the current navigation instruction.
The model then generates the next video segment conditioned on this prescribed viewpoint change.

\myparagraph{LingBot-World~\cite{gao2026lingbot}.}
LingBot-World is an interactive world model targeting minute-level generation horizons with sub-second latency at 8~fps, designed for long-context robotic simulation.
It is built on a Wan2.2 14B I2V backbone and consumes camera-pose sequences as the control signal: the given navigation action is first converted into a deterministic pose sequence by a geometric mapping and then fed to the backbone for frame generation, with no learned planner involved.
For \benchmark, we use the officially released \texttt{lingbot-world-base-cam} checkpoint.
Training data is a hybrid mixture of real-world footage, game-engine recordings, and synthetic renderings, providing broad coverage of camera dynamics and scene diversity.

\myparagraph{HY-World~1.5~\cite{sun2025hyworldplay}.}
HY-World~1.5 is Tencent Hunyuan's streaming Diffusion Transformer for real-time interactive world generation, built on the 480p I2V branch of HunyuanVideo~1.5.
The deployed transformer contains roughly $8.3$B parameters.
It introduces a dual action representation that encodes both discrete keyboard tokens and continuous camera deltas, together with a reconstituted context memory module that compresses long-horizon history into a fixed-size token buffer to maintain geometric consistency across hundreds of generation steps.
This design allows the model to sustain coherent spatial layouts and stable subject appearances over extended multi-turn trajectories, which is directly reflected in its leading navigation adherence scores on \benchmark.

\myparagraph{Fantasy-World~\cite{fantasyworld2025}.}
Fantasy-World achieves geometry-consistent interactive world modeling by augmenting a standard video Diffusion Transformer with a parallel geometric prediction branch that jointly estimates depth, point, and camera trajectory maps for each generated frame.
For \benchmark, we use the officially released \texttt{acvlab/FantasyWorld-Wan2.1-I2V-14B-480P} adapter.
Within each generation segment, the estimated geometric representations are fed back as conditioning signals for subsequent frames via IRG blocks, creating an implicit 3D consistency loop that constrains the generated video to respect plausible spatial structure.
Across segments, only the last RGB frame is carried forward as the next conditioning input, so the geometric feedback operates at the segment level rather than globally across the whole video.
The model is trained on a curated dataset of game-engine renderings with dense geometric annotations, and it is evaluated on navigation cases in \benchmark where the 3D geometric consistency signal is most informative.

\myparagraph{InSpatio-World~\cite{inspatio2026}.}
InSpatio-World introduces a state-anchored world modeling paradigm for 4D exploration, in which the world state is represented as a set of spatiotemporal anchor tokens that persist across steps and are updated incrementally as the user navigates.
For \benchmark, we use the officially released $1.3$B distilled configuration, \texttt{InSpatio-World-1.3B}.
Generation is performed via a causal block-wise autoregressive pipeline with KV caching, where each new latent block is conditioned on the current action and the cached history of previously generated blocks, rather than in a single-shot V2V pass.
Since the V2V interface expects a fixed-length video input, we construct it by replicating the initial conditioning frame up to the required number of frames.
InSpatio-World achieves strong spatial consistency scores due to its explicit state-maintenance mechanism, despite operating at a higher dynamic degree than most camera-controlled baselines.

\myparagraph{Astra~\cite{zhu2025astra}.}
Astra is a general interactive world model based on an autoregressive denoising framework that jointly predicts the next-frame latent and its denoising trajectory conditioned on an action token sequence.
It is built on a Wan2.1-T2V-1.3B base with an action-conditioned adapter, and the combined model contains roughly $2$B parameters.
Operating at 832$\times$480 resolution, Astra prioritizes temporal coherence and action responsiveness over raw visual fidelity, and its open-source architecture makes it accessible for ablation studies and downstream fine-tuning.
The model is evaluated at its native resolution, and scores are reported without upsampling to ensure fair comparison.

\subsubsection{Action-conditioned Models}

These models natively accept discrete or continuous action signals and generate subsequent observations conditioned on the full action history.

\myparagraph{Happy~Oyster~\cite{alibaba2026happyoyster}.}
Happy~Oyster is Alibaba's real-time interactive world creation system, supporting two distinct interaction paradigms: a \emph{directing} mode in which the user issues discrete navigation commands, such as W/A/S/D and rotation keys, and a \emph{wandering} mode in which the model autonomously explores an initialized world.
The model is built on a streaming generative architecture that renders video frames at native resolution with low latency, allowing real-time human-in-the-loop interaction through its official web interface.
Like Genie~3, it is evaluated following the standardized web-based protocol in Appendix~\ref{app:genie3_eval}.

\myparagraph{Matrix-Game~2.0~\cite{he2025matrixgame2} and Matrix-Game~3.0~\cite{he2026matrixgame3}.}
Matrix-Game is a series of autoregressive diffusion models for interactive world generation developed by the Skywork team.
Matrix-Game~2.0 introduces a few-step AR diffusion framework that decouples temporal prediction from per-frame quality refinement, enabling real-time streaming generation with competitive visual quality; the released distilled transformer has approximately $1.6$B parameters.
Matrix-Game~3.0 scales the architecture and adds a camera-aware memory retrieval module that indexes previously generated keyframes for spatial consistency, and achieves 40~fps at 720p; the released base transformer has approximately $6.5$B parameters.
For \benchmark, Matrix-Game~3.0 is evaluated using the web-interface protocol described in Appendix~\ref{app:genie3_eval}, where a human operator issues keyboard commands and the resulting video stream is screen-captured for offline metric computation.
Matrix-Game~2.0 generates each rollout through a global iterative schedule of fixed-size $40$-frame chunks, with $57$ frames for the first chunk, and the resulting sequence is partitioned into per-turn clips.
We evaluate both versions to trace the progression of the Matrix-Game design across generations.

\myparagraph{Genie~3~\cite{ball2025genie3}.}
Genie~3 is Google DeepMind's third-generation interactive world model, generating diverse and visually rich playable environments at 24~fps conditioned on keyboard and mouse inputs.
The model is built on a large-scale autoregressive Transformer trained on internet-scale gameplay video, enabling multi-minute visual memory and zero-shot generalization to novel scene descriptions from a single conditioning image.
Because Genie~3 does not expose a public API, evaluation on \benchmark follows the web-interface protocol described in Appendix~\ref{app:genie3_eval}, where a human operator issues keyboard commands and the resulting video stream is screen-captured for offline metric computation.

\myparagraph{Hunyuan-GameCraft~\cite{li2025hunyuangamecraft}.}
Hunyuan-GameCraft is Tencent's interactive game video generation model, employing a history-conditioned DiT architecture that conditions each new frame on a rolling window of past frames and control tokens.
The deployed model is built on the full HunyuanVideo dual-stream Transformer, and the distilled checkpoint we evaluate contains roughly $13$B parameters.
The model is trained on a large-scale dataset of over one million gameplay recordings across more than 100 AAA games with paired keyboard and mouse annotations, specializing it for high-dynamic game-engine-style interactions where rapid camera movement, character animation, and environmental changes must be rendered consistently.
It supports real-time generation via model distillation, taking keyboard and mouse as native input modalities.
Each turn is assembled by concatenating fixed-size $33$-frame segments, with the number of segments per turn chosen as $\lceil T_{\text{turn}} / T_{\text{seg}} \rceil$ to best approximate the target turn duration $T_{\text{turn}}$.

\myparagraph{Infinite-World~\cite{wu2026infiniteworld}.}
Infinite-World is an interactive world model developed at Nankai University and Meituan, designed to scale generation horizons to 1000+ frames without explicit geometric priors.
It is built on a Wan2.1-T2V-1.3B base with an additional action encoder, and the combined checkpoint contains roughly $1.5$B parameters.
The model generates video autoregressively, conditioning each new chunk on the latent representation of previously generated frames together with the current discrete keyboard action, which allows it to sustain coherent spatial layouts over very long rollouts.
Each turn is assembled by concatenating fixed-size $80$-frame iterations, with the number of iterations per turn chosen to best approximate the target turn duration.

\subsubsection{Inference Speed Analysis}

\cref{tab:evaluated_models} also reports the generation efficiency of some evaluated models, measured by the ratio between wall-clock generation time and the duration of the produced video. This value can be interpreted as seconds of computation per second of output video; equivalently, for frame-based generation, it corresponds to $\mathrm{fps}\times s_{\text{frame}}$, where $s_{\text{frame}}$ denotes the average generation time per frame. Lower values indicate faster generation, and values below $1$ indicate faster-than-real-time generation.

For text-driven models, the reported value corresponds to the time required for a single API call or local inference pass to generate one video clip. For camera-controlled and action-conditioned models, we measure the wall-clock time required to generate the number of frames used for one evaluation turn and divide it by the corresponding video duration. All locally deployed models are evaluated on a single NVIDIA A100 80GB GPU to ensure comparable efficiency measurements, while API- or web-only models are marked separately when their underlying hardware or runtime is not accessible.

\subsection{Web-Based Evaluation Protocol (Genie~3 \& Happy~Oyster)}
\label{app:genie3_eval}

\begin{figure}[H]
\centering
\includegraphics[width=\linewidth]{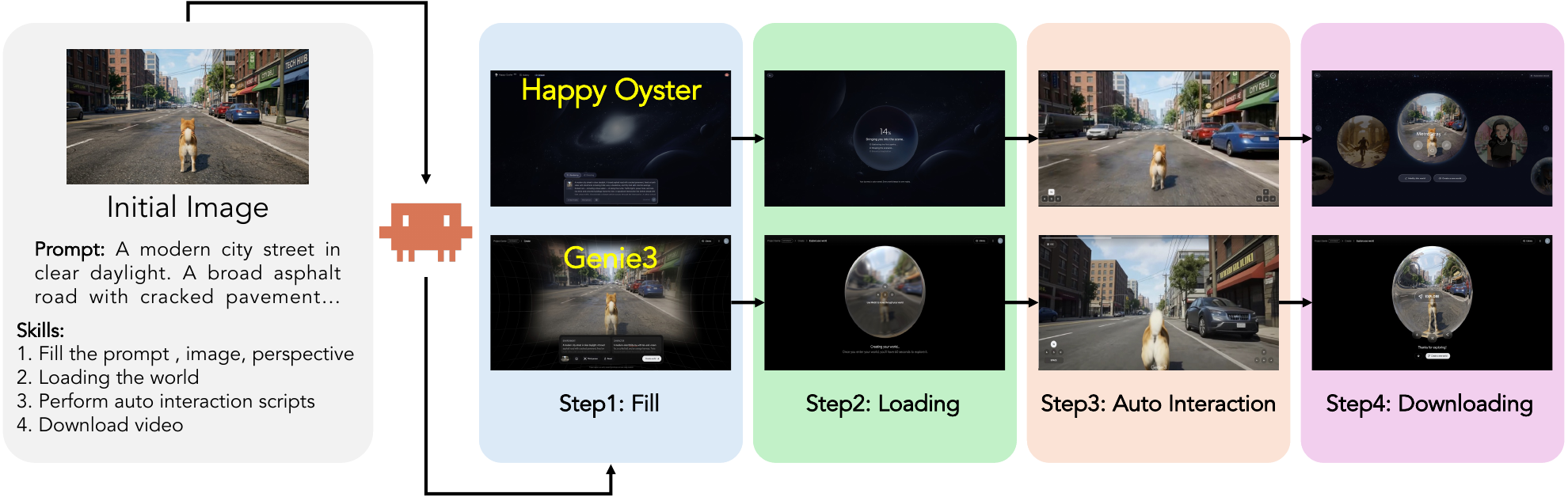}
\caption[Automated web-based evaluation pipeline]{Automated web-based evaluation pipeline for Genie~3 and Happy~Oyster. Given an initial image and a text prompt, the browser-use agent follows four steps: \textbf{Step~1} fills in the prompt, image, and perspective; \textbf{Step~2} waits for world loading to complete; \textbf{Step~3} executes the interaction script with each turn lasting 5 seconds; \textbf{Step~4} downloads the recorded video.}
\label{fig:genie_eval_pipeline}
\end{figure}

Genie~3~\cite{ball2025genie3} and Happy~Oyster~\cite{alibaba2026happyoyster} are only accessible through preview web interfaces and do not expose public APIs or downloadable weights.
To enable programmatic evaluation at scale, we automate the entire interaction pipeline using Claude Code's browser-use capability, which controls Chrome via structured commands.
As illustrated in \cref{fig:genie_eval_pipeline}, the process is decomposed into four steps:
\textbf{(1)~Fill}: the script fills in the text prompt, uploads the initial image, and configures the perspective setting on the model's web page;
\textbf{(2)~Loading}: the script waits for the model to finish loading the world;
\textbf{(3)~Auto Interaction}: once world loading is detected as complete, the script automatically executes the pre-defined interaction sequence, holding each navigation key for 5 seconds per turn;
\textbf{(4)~Downloading}: the generated video is downloaded for offline evaluation.

\section{Per-Metric Evaluation Details}
\label{app:metric_details}

\subsection{Metric Quick Reference}
\label{app:metric_quickref}

\cref{tab:metric_quickref} provides a unified index of all \numsubmetric evaluation sub-metrics. Entries in the \textbf{Detail} column are hyperlinked to the corresponding methodology description.

\begin{table}[H]
\caption[Metric quick-reference]{Metric quick-reference table. ``EM'' = Expert Model; ``VLM'' = vision-language model judge.}
\vspace{+1mm}
\label{tab:metric_quickref}
\centering
\renewcommand{\arraystretch}{1.15}
\scriptsize
\setlength{\tabcolsep}{3pt}
\begin{tabular}{@{}l l l l c@{}}
\toprule
\textbf{Dimension} & \textbf{Sub-Metric} & \textbf{Method} & \textbf{Tool / Model} & \textbf{Detail} \\
\midrule
\multirow{6}{*}{Video Quality}
  & Aesthetic Quality    & EM  & CLIP + LAION head     & \hyperref[app:aesthetic_detail]{\S\ref*{app:aesthetic_detail}} \\
  & Imaging Quality      & EM  & MUSIQ                 & \hyperref[app:imaging_detail]{\S\ref*{app:imaging_detail}} \\
  & Temporal Flickering  & EM  & Pixel MAE             & \hyperref[app:flickering_detail]{\S\ref*{app:flickering_detail}} \\
  & Dynamic Degree       & EM  & RAFT optical flow     & \hyperref[app:dynamic_detail]{\S\ref*{app:dynamic_detail}} \\
  & Motion Smoothness    & EM  & AMT-S interpolation   & \hyperref[app:smoothness_detail]{\S\ref*{app:smoothness_detail}} \\
  & HPSv3-Norm           & EM  & HPSv3                 & \hyperref[app:hps_detail]{\S\ref*{app:hps_detail}} \\
\midrule
\multirow{2}{*}{Setting Adh.}
  & Scene Adherence      & VLM & Doubao-Seed-2.0-lite  & \hyperref[app:scene_adherence_detail]{\S\ref*{app:scene_adherence_detail}} \\
  & Subject Adherence    & VLM & Doubao-Seed-2.0-lite  & \hyperref[app:subject_adherence_detail]{\S\ref*{app:subject_adherence_detail}} \\
\midrule
\multirow{4}{*}{Interaction Adh.}
  & NavScore             & EM  & MegaSaM pose est.     & \hyperref[app:nav_detail]{\S\ref*{app:nav_detail}} \\
  & Event Editing Adh.   & VLM & Doubao-Seed-2.0-lite  & \hyperref[app:event_edit_detail]{\S\ref*{app:event_edit_detail}} \\
  & Subject Action Adh.  & VLM & Doubao-Seed-2.0-lite  & \hyperref[app:subject_action_detail]{\S\ref*{app:subject_action_detail}} \\
  & Persp.\ Switching Adh.  & VLM & Doubao-Seed-2.0-lite  & \hyperref[app:persp_switch_detail]{\S\ref*{app:persp_switch_detail}} \\
\midrule
\multirow{6}{*}{Consistency}
  & Subject Consistency  & EM  & DINOv2 + CLIP + SAM2  & \hyperref[app:subject_consistency_detail]{\S\ref*{app:subject_consistency_detail}} \\
  & Background Cons.     & EM  & CLIP ViT-B/32         & \hyperref[app:background_consistency_detail]{\S\ref*{app:background_consistency_detail}} \\
  & Spatial Consistency  & EM  & MegaSaM + DreamSim    & \hyperref[app:spatial_detail]{\S\ref*{app:spatial_detail}} \\
  & Segment Continuity   & EM  & TransNetV2            & \hyperref[app:segment_continuity_detail]{\S\ref*{app:segment_continuity_detail}} \\
  & Perspective Cons.    & EM  & SAM2   & \hyperref[app:perspective_consistency]{\S\ref*{app:perspective_consistency}} \\
  & Reconstruction Cons. & EM  & Depth Anything 3 reproj. & \hyperref[app:reconstruction_detail]{\S\ref*{app:reconstruction_detail}} \\
\midrule
\multirow{2}{*}{Physical}
  & Causal Fidelity     & VLM & Doubao-Seed-2.0-lite  & \hyperref[app:causal_fidelity_detail]{\S\ref*{app:causal_fidelity_detail}} \\
  & Visual Plausibility & EM & Qwen3VL-35B (ft.)     & \hyperref[app:visual_plausibility_detail]{\S\ref*{app:visual_plausibility_detail}} \\
\bottomrule
\end{tabular}
\end{table}

\tcbset{
  d2box/.style={
    enhanced,
    colback=teal!8,
    colframe=teal!80,
    boxrule=0.5pt,
    arc=3pt,
    left=6pt, right=6pt, top=5pt, bottom=5pt,
    fonttitle=\bfseries\small,
    colbacktitle=teal!35,
    coltitle=black,
    attach boxed title to top left={yshift=-2mm, xshift=4mm},
    boxed title style={arc=2pt, boxrule=0.3pt},
    breakable,
  },
  d3box/.style={
    enhanced,
    colback=orange!8,
    colframe=orange!80,
    boxrule=0.5pt,
    arc=3pt,
    left=6pt, right=6pt, top=5pt, bottom=5pt,
    fonttitle=\bfseries\small,
    colbacktitle=orange!35,
    coltitle=black,
    attach boxed title to top left={yshift=-2mm, xshift=4mm},
    boxed title style={arc=2pt, boxrule=0.3pt},
    breakable,
  },
  d5box/.style={
    enhanced,
    colback=red!8,
    colframe=red!70,
    boxrule=0.5pt,
    arc=3pt,
    left=6pt, right=6pt, top=5pt, bottom=5pt,
    fonttitle=\bfseries\small,
    colbacktitle=red!30,
    coltitle=black,
    attach boxed title to top left={yshift=-2mm, xshift=4mm},
    boxed title style={arc=2pt, boxrule=0.3pt},
    breakable,
  },
  systemprompt/.style={
    colback=gray!10,
    colframe=gray!60,
    boxrule=0.5pt,
    arc=2pt,
    fontupper=\ttfamily\scriptsize,
    left=4pt, right=4pt, top=3pt, bottom=3pt,
    title={\scriptsize\textbf{System Prompt}},
  },
  userprompt/.style={
    colback=blue!8,
    colframe=blue!50,
    boxrule=0.5pt,
    arc=2pt,
    fontupper=\ttfamily\scriptsize,
    left=4pt, right=4pt, top=3pt, bottom=3pt,
    title={\scriptsize\textbf{User Prompt}},
  },
}

This appendix provides per-sub-metric evaluation details organized by dimension. Each sub-metric includes its definition and evaluation examples (VLM prompts or qualitative cases where applicable). The full list of all \numsubmetric sub-metrics is summarized in the unified metric quick-reference table (\cref{tab:metric_quickref} in \cref{app:metric_quickref}).

\subsection{Video Quality}
\label{app:vq_detail}

All six Video Quality sub-metrics are computed by expert models and produce scores in $[0, 100]$ (higher is better).

\subsubsection{Aesthetic Quality}
\label{app:aesthetic_detail}

\myparagraph{Definition.}
Frames are sampled at 2\,FPS. Each frame is encoded by CLIP ViT-L/14, and a LAION Aesthetic linear head~\cite{schuhmann2022laion5b} predicts an aesthetic score in $[0, 10]$. The final score is the mean over all sampled frames, rescaled to $[0, 100]$:
\begin{equation}
  S_{\text{aesth}} = \frac{1}{N}\sum_{i=1}^{N} f_{\text{LAION}}(\text{CLIP}(x_i)) \times 10.
\end{equation}

\subsubsection{Imaging Quality}
\label{app:imaging_detail}

\myparagraph{Definition.}
Frames are sampled at 2\,FPS and resized such that the longer edge $\leq 512$. MUSIQ~\cite{ke2021musiq} predicts a per-frame quality score in $[0, 100]$:
\begin{equation}
  S_{\text{imag}} = \frac{1}{N}\sum_{i=1}^{N} \text{MUSIQ}(x_i).
\end{equation}

\subsubsection{Temporal Flickering}
\label{app:flickering_detail}

\myparagraph{Definition.}
All consecutive frame pairs are compared via pixel-level mean absolute error (MAE). The score is inverted so that lower flickering yields a higher score:
\begin{equation}
  S_{\text{flick}} = \frac{255 - \frac{1}{N-1}\sum_{i=1}^{N-1}\text{MAE}(x_i, x_{i+1})}{255} \times 100.
\end{equation}

\subsubsection{Dynamic Degree}
\label{app:dynamic_detail}

\myparagraph{Definition.}
Frames are sampled at 8\,FPS. RAFT optical flow is computed for each consecutive pair, and the top 5\% magnitude values are averaged per pair:
\begin{equation}
  m_i = \mathrm{mean}\!\bigl(\mathrm{top}_{5\%}\lVert \mathbf{f}_i \rVert\bigr), \quad
  S_{\mathrm{dyn}} =
  \begin{cases}
    100 & \text{if } \sum_{i} \mathbf{1}[m_i > \tau] \geq N_{\min}, \\
    0   & \text{otherwise},
  \end{cases}
\end{equation}
where $\mathbf{f}_i$ is the optical flow field for frame pair $i$, $\tau$ is the dynamic threshold, and $N_{\min}$ is the minimum count of dynamic pairs required.

\subsubsection{Motion Smoothness}
\label{app:smoothness_detail}

\myparagraph{Definition.}
Following VBench, we use AMT-S~\cite{li2023amt} to interpolate even-indexed frames from odd-indexed neighbors, then compute pixel MAE between predicted and actual even frames:
\begin{equation}
  S_{\mathrm{app}} = \frac{255 - \frac{1}{M}\sum_{j=1}^{M}\text{MAE}(\hat{x}_{2j}, x_{2j})}{255} \times 100.
\end{equation}

\subsubsection{HPSv3-Norm}
\label{app:hps_detail}

\myparagraph{Definition.}
We compute the raw HPSv3 reward for each frame and apply linear percentile normalization. Let $p_1$ and $p_{99}$ denote the 1st and 99th percentiles of all per-video mean scores across the evaluated models:
\begin{equation}
  S_{\text{HPS}} = \text{clip}\!\left(\frac{\bar{r} - p_1}{p_{99} - p_1} \times 100,\; 0,\; 100\right),
\end{equation}
where $\bar{r}$ is the per-video mean raw reward. In our evaluation, $p_1 = 5.21$ and $p_{99} = 8.66$.

\subsection{Setting Adherence}
\label{app:setting_detail}

Setting adherence evaluates whether the generated video faithfully realizes the declared world settings.
It comprises two VLM-based sub-metrics: \emph{scene adherence} (environment and offscreen elements) and \emph{subject adherence} (appearance and motion style).
All prompts are issued to \texttt{Doubao-Seed-2.0-lite} with video frames sampled at 2\,fps from the first interaction turn.

\myparagraph{Pre-processing: world-setting decomposition.}
Before evaluation, a VLM automatically decomposes each case's world-setting description into a \emph{visible part} (elements present in the initial frame) and an \emph{offscreen part} (elements expected to appear only after camera movement).
For subject adherence, the subject description is further decomposed into an \emph{appearance part} (visual attributes) and an \emph{action part} (expected movement pattern).
The decomposition is precomputed once per case and shared across all models.

\subsubsection{Scene Adherence}
\label{app:scene_adherence_detail}

\myparagraph{Definition.}
Scene adherence combines two complementary signals:
(i)~\emph{environment maintenance} ($s_{\mathrm{maint}} \in [1,5]$): how well the initially visible scene elements and visual style are preserved across the generated segment;
(ii)~\emph{offscreen content appearance} ($s_{\mathrm{off}} \in \{0,1\}$): whether elements described as offscreen in the initial frame become visible after the prescribed camera movement.
The per-case scene adherence score is $(s_{\mathrm{maint}}/5 + s_{\mathrm{off}}) / 2$.

\myparagraph{Prompt.}
Both aspects are evaluated in a single VLM call with structured JSON output:

\begin{tcolorbox}[d2box, title={Scene Adherence Prompt}]
\ttfamily\scriptsize
You are a video analysis expert. You will see video frames from Turn 1 of a generated video (sampled at 2fps).\\
\\
Evaluate the video on two aspects:\\
\\
**1. Environment Maintenance (1-5)**\\
How well does the video maintain these scene elements that were visible in the initial frame? Also consider whether the visual style remains consistent.\\
Scene description: [VISIBLE\_PART]\\
Visual style: [STYLE]\\
\\
Scoring:\\
- 5: All elements perfectly maintained, style fully consistent\\
- 4: Most elements maintained, style mostly consistent, minor issues\\
- 3: Some elements maintained or style partially drifts\\
- 2: Few elements maintained, significant changes or style mismatch\\
- 1: Scene barely matches the description or style completely wrong\\
\\
**2. Offscreen Content Appearance (0 or 1)**\\
The following elements were described as existing OFF-SCREEN (not visible in the initial frame). Did ANY of them appear in the video as the camera moved?\\
Description: [OFFSCREEN\_PART]\\
\\
- 1: At least one offscreen element appeared\\
- 0: None of the offscreen elements appeared\\
\\
Answer in JSON format ONLY. Each reason must be within 20 words.\\
\{``maintenance'': <1-5>, ``maintenance\_reason'': ``...'', ``offscreen'': <0 or 1>, ``offscreen\_reason'': ``...''\}
\end{tcolorbox}

\subsubsection{Subject Adherence}
\label{app:subject_adherence_detail}

\myparagraph{Definition.}
Subject adherence (third-person cases only) combines:
(i)~\emph{appearance maintenance} ($s_{\mathrm{app}} \in [1,5]$): how well the subject's visual appearance (shape, color, clothing, equipment) is maintained throughout the segment;
(ii)~\emph{action match} ($s_{\mathrm{act}} \in \{0,1\}$): whether the subject exhibits the described movement or action pattern.
The per-case subject adherence score is $(s_{\mathrm{app}}/5 + s_{\mathrm{act}}) / 2$.

\myparagraph{Prompt.}
Both aspects are evaluated in a single VLM call with structured JSON output:

\begin{tcolorbox}[d2box, title={Subject Adherence Prompt}]
\ttfamily\scriptsize
You are a video analysis expert. You will see video frames from Turn 1 of a generated video (sampled at 2fps).\\
\\
Evaluate the video on two aspects:\\
\\
**1. Subject Appearance Maintenance (1-5)**\\
How well does the video maintain the subject's visual appearance (shape, color, clothing, equipment, category)?\\
Subject appearance: [APPEARANCE\_PART]\\
\\
Scoring:\\
- 5: Subject appearance perfectly maintained throughout\\
- 4: Mostly maintained, minor visual changes\\
- 3: Some features maintained, noticeable changes\\
- 2: Few features maintained, significant appearance drift\\
- 1: Subject barely recognizable or completely changed\\
\\
**2. Subject Action (0 or 1)**\\
Does the subject exhibit the described movement/action pattern in the video?\\
Expected action: [ACTION\_PART]\\
\\
- 1: The described action/movement pattern is clearly visible\\
- 0: The action/movement pattern is not visible or completely different\\
\\
Answer in JSON format ONLY. Each reason must be within 20 words.\\
\{``maintenance'': <1-5>, ``maintenance\_reason'': ``...'', ``action'': <0 or 1>, ``action\_reason'': ``...''\}
\end{tcolorbox}

\subsection{Interaction Adherence}
\label{app:interaction_detail}

Interaction adherence comprises navigation trajectory evaluation (expert model) and three VLM-based sub-metrics for event editing, subject action, and perspective switching.

\subsubsection{NavScore}
\label{app:nav_detail}

\myparagraph{Definition.}
NavScore evaluates navigation accuracy by comparing predicted camera trajectories against synthetic ground-truth (GT) trajectories constructed from the action sequence.

\myparagraph{Camera pose estimation.}
We extract per-frame camera-to-world poses using MegaSaM~\cite{li2024megasam}, a monocular structure-from-motion method that produces dense 4$\times$4 pose matrices from video.
Each multi-turn video yields a single globally consistent pose sequence, which is then split at turn boundaries for per-turn and global analysis.

\myparagraph{Ground-truth trajectory construction.}
GT trajectories are built adaptively per turn based on the action type:
\begin{itemize}[leftmargin=1.5em, nosep]
  \item \emph{Pure translation} (W/S/A/D): a straight-line trajectory in the action direction under the initial camera orientation, with length matched to the predicted displacement. If the predicted displacement is below a minimum threshold ($0.1$ units), a fallback length of $1.0$ is used to penalize non-responsive models.
  \item \emph{Rotation} (left/right/up/down): an adaptive orbital trajectory with radius $R$ and angle $\theta$ estimated from the predicted trajectory via $R = \text{chord} / (2\sin(\theta/2))$. For third-person perspectives, $R$ has a minimum of $1.0$ to prevent degenerate orbits. If the predicted rotation is below $3^\circ$, fallback parameters ($\theta = 30^\circ$, $R = 1.0$) are applied.
\end{itemize}
The GT is then aligned to the predicted coordinate frame (translation offset + rotation alignment), while the predicted trajectory remains unchanged.

\myparagraph{Arc-length resampling.}
To ensure equal weighting across turns regardless of frame count or motion speed, both GT and predicted trajectories are resampled to a fixed number of points ($K = 20$) uniformly along their respective arc lengths.
Position is linearly interpolated and rotation is interpolated via Slerp along the arc-length parameter.

\myparagraph{Evaluation metrics.}
We evaluate navigation accuracy using normalized Absolute Trajectory Error (ATE), which measures the global shape agreement between predicted and ground-truth trajectories after per-turn arc-length resampling and concatenation:
\begin{align}
  \mathrm{nATE}_t &= \min\!\left(\frac{\mathrm{ATE}_t}{\max(L_{\mathrm{pred}},\; 0.5)},\; 1\right), \label{eq:nate_t} \\
  \mathrm{nATE}_r &= \min\!\left(\frac{\mathrm{ATE}_r}{\max(\Theta_{\mathrm{pred}},\; 10^\circ)},\; 1\right),
\end{align}
where $L_{\mathrm{pred}}$ and $\Theta_{\mathrm{pred}}$ are the predicted path length and total rotation.
The minimum denominators ($0.5$ for translation, $10^\circ$ for rotation) prevent instability when the model produces near-zero motion.
We also compute Relative Pose Error (RPE) and report it as an auxiliary diagnostic, but it is excluded from the final score because action-conditioned world models often exhibit non-uniform motion profiles, making frame-level velocity-matched GT trajectories unreliable.
ATE therefore serves as the primary signal for trajectory-shape correctness.

\myparagraph{Trajectory consistency.}
For repeated or symmetric actions, we compare all valid same-group turn pairs, including W$\leftrightarrow$S, A$\leftrightarrow$D, left$\leftrightarrow$right, and up$\leftrightarrow$down.
Each turn is first normalized to its own start pose, and symmetric pairs are mirrored into a shared canonical frame when needed.
We then compute pairwise normalized ATE between the two turns and average over all valid pairs:
\begin{align}
  \mathrm{cnATE}_t &= \frac{1}{P}\sum_{p=1}^{P} \mathrm{nATE}^{(p)}_t, \\
  \mathrm{cnATE}_r &= \frac{1}{P}\sum_{p=1}^{P} \mathrm{nATE}^{(p)}_r,
\end{align}
where $P$ is the number of valid same-group turn pairs.
The trajectory consistency term is then defined as
\begin{equation}
  \mathrm{Cons} = 1 - \frac{\mathrm{cnATE}_t + \mathrm{cnATE}_r}{2}.
\end{equation}
If no valid same-group pair exists, we set $\mathrm{cnATE}_t = \mathrm{cnATE}_r = 0$, which yields $\mathrm{Cons} = 1$.

\myparagraph{NavScore aggregation.}
The final NavScore for each video is computed as
\begin{align}
  \mathrm{Acc} &= 1 - \frac{\mathrm{nATE}_t + \mathrm{nATE}_r}{2}, \\
  \mathrm{NavScore} &= \frac{\mathrm{Acc} + \mathrm{Cons}}{2}.
\end{align}
Model-level scores are the mean across all navigation cases, rescaled to $[0, 100]$.

\begin{figure}[H]
\centering
\includegraphics[width=\linewidth]{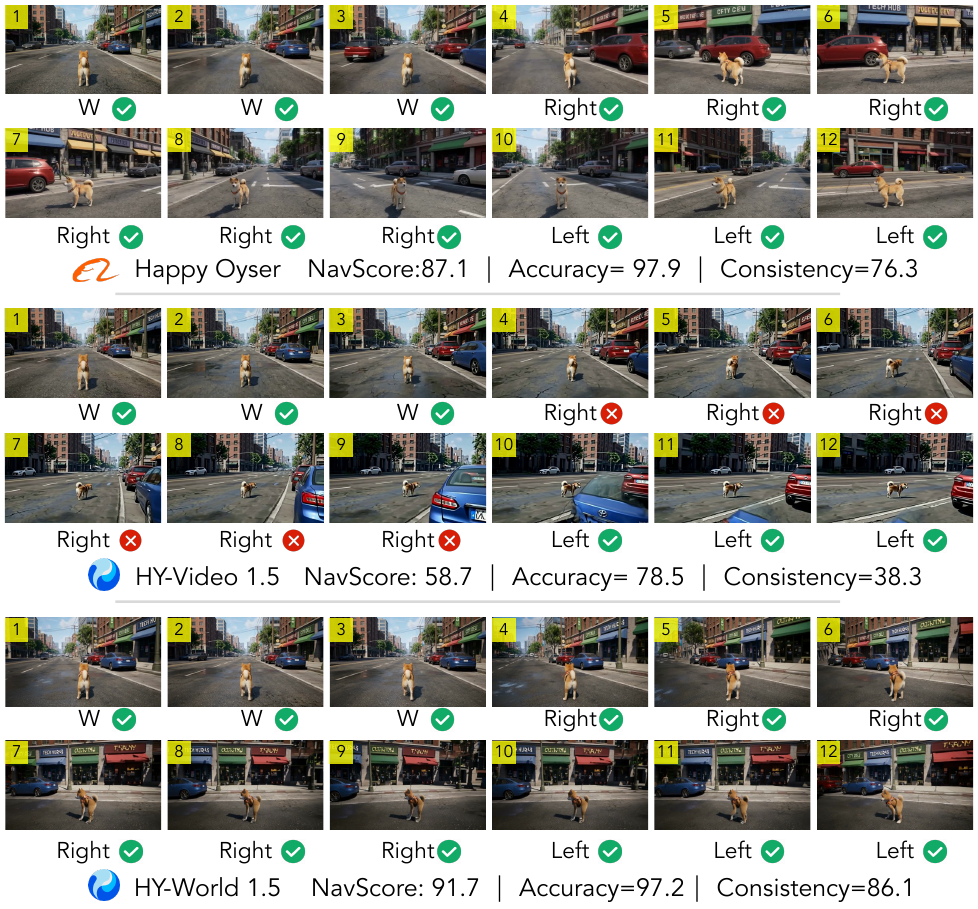}
\caption[Navigation qualitative comparison]{A four-turn navigation case (T1: \texttt{W}, T2: $\rightarrow$, T3: $\rightarrow$, T4: $\leftarrow$) across three models. Frames 1--3 correspond to T1, 4--6 to T2, 7--9 to T3, and 10--12 to T4. Happy Oyster and HY-World~1.5 follow the instructed directions correctly, while HY-Video~1.5 reverses the rotation in T2 and T3.}
\label{fig:navi_visualize}
\end{figure}

\begin{figure}[H]
\centering
\includegraphics[width=\linewidth]{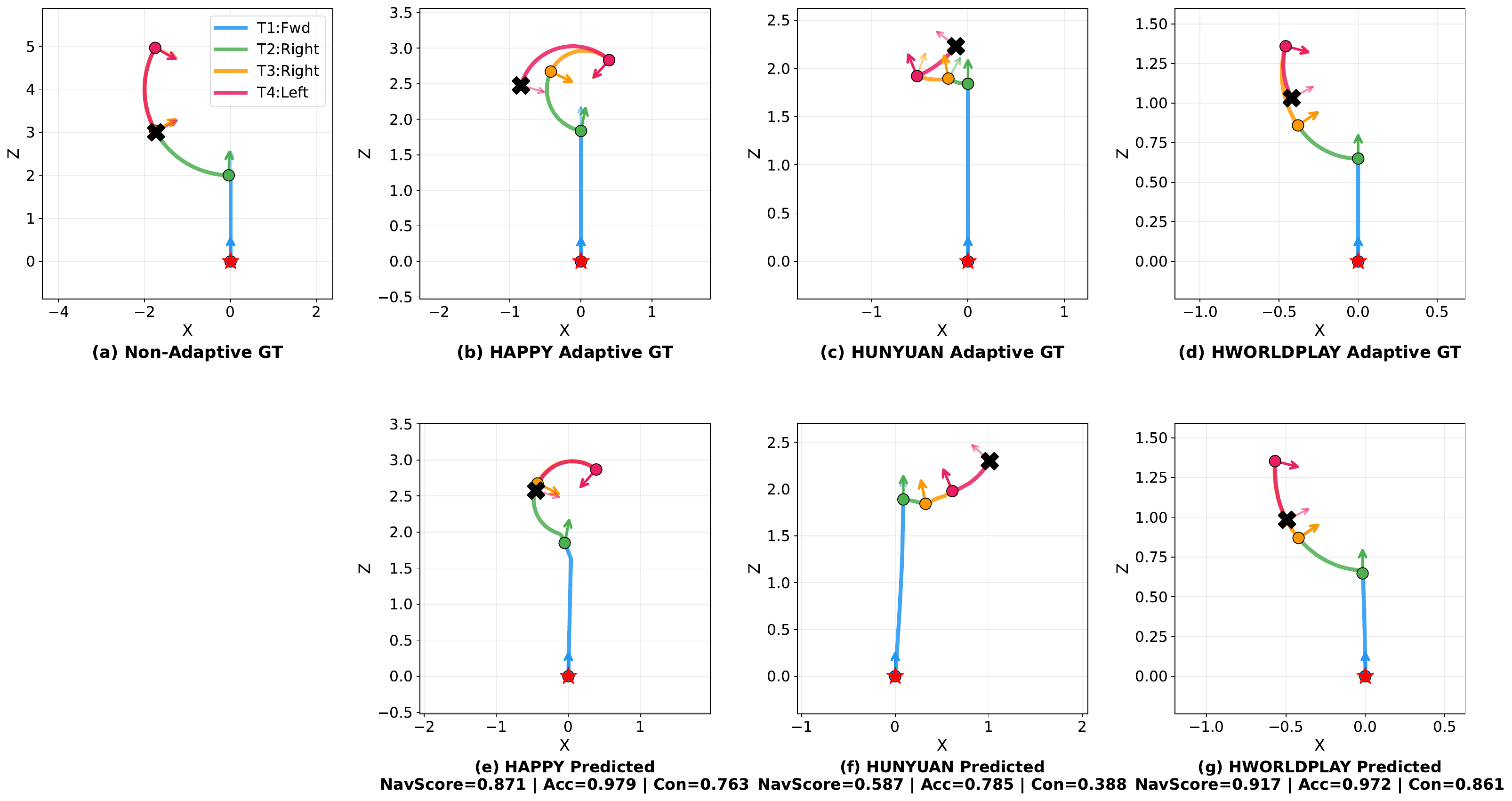}
\caption[Adaptive ground-truth construction]{Adaptive GT construction for the same case. The GT trajectory adapts to each model's predicted motion magnitude, so correctly navigating models with different amplitudes (Happy Oyster vs.\ HY-World~1.5) both receive fair evaluation. For HY-Video~1.5, the wrong-direction motion is captured as trajectory error rather than a scale mismatch.}
\label{fig:7panels_gt_pred}
\end{figure}

\myparagraph{Qualitative Results.}
\cref{fig:navi_visualize} shows a four-turn navigation case (T1: \texttt{W}, T2: $\rightarrow$, T3: $\rightarrow$, T4: $\leftarrow$) evaluated on three models. Happy Oyster and HY-World~1.5 produce trajectories that consistently follow the instructed directions across all turns, while HY-Video~1.5 reverses the rotation direction in T2 and T3, resulting in clear trajectory deviation.
\cref{fig:7panels_gt_pred} illustrates how the adaptive ground-truth is constructed per model. Because different models accept different input interfaces and exhibit varying motion magnitudes, the GT trajectory adapts its scale to the predicted path length. Both Happy Oyster and HY-World~1.5 navigate correctly but with different amplitudes, so their adaptive GTs differ in scale while both yielding high NavScores. In contrast, HY-Video~1.5 moves in the wrong direction, and the adaptive GT faithfully reflects this directional error, penalizing it appropriately.

\subsubsection{Event Editing Adherence}
\label{app:event_edit_detail}

\myparagraph{Definition.}
Event editing adherence uses \texttt{Doubao-Seed-2.0-lite} with video frames sampled at 3\,fps. Five progressive binary questions are asked per turn, each with an \emph{expected} answer. The turn receives one point per question whose response matches the expected answer, yielding a raw score in $[0,5]$. The case score is the mean over all event-edit turns, rescaled to $[0, 100]$ by multiplying by $20$.

\myparagraph{Prompt.}
Event editing and subject action share the following system prompt:

\begin{tcolorbox}[d3box, title={Shared System Prompt (event editing \& subject action)}]
\begin{tcolorbox}[systemprompt]
You are evaluating AI-generated video segments from interactive world models. Each video segment corresponds to one interaction turn where a specific event was instructed to happen.\\
\\
Your task: answer with ONLY `Yes' or `No' based on what you observe in the provided frames. Follow these guidelines:\\
- These are AI-generated videos, not real footage. Minor visual artifacts, slight inconsistencies, or imperfect rendering should NOT cause a `No' if the overall intent is recognizable.\\
- Focus on whether the CORE semantic meaning of the described event is visually present, not whether every literal detail matches perfectly.\\
- A long or detailed event description may only be partially depicted. Judge based on whether the main action or change is recognizable.\\
- When in doubt between Yes and No, lean toward the interpretation that better matches what you actually see, not what you expect to see.
\end{tcolorbox}
\vspace{2pt}
\textit{\small Each response is followed by a one-line \texttt{Reason:} field (at most 50 words) that names the specific issue whenever the point is not awarded.}
\end{tcolorbox}

\vspace{6pt}

\begin{tcolorbox}[d3box, title={Event Editing Adherence (Q1--Q5)}]
\ttfamily\scriptsize
World setting: [SCENE\_DESCRIPTION]\\
Interaction instruction for this turn: [ACTION]\\
\\
The following frames are uniformly sampled from the generated video segment for this turn.\\
\mbox{[FRAME\_SEQUENCE]}\\
\\
Q1 (Static Scene Check): Does the scene remain largely unchanged, with no clear indication of `[ACTION]' occurring? \emph{Answer No if there is ANY recognizable change related to this event.} Expected: \texttt{No}.\\
Q2 (Event Occurrence): Does the video show something resembling or consistent with the described event `[ACTION]'? \emph{Partial depiction counts as Yes.} Expected: \texttt{Yes}.\\
Q3 (Event Completion): By the end of this segment, has the described event `[ACTION]' reached a clear conclusion or outcome? \emph{Ongoing events without a visible end state receive No.} Expected: \texttt{Yes}.\\
Q4 (Detail Accuracy): Are the key details of `[ACTION]' accurately depicted, such as the correct objects, agents, directions, and quantities? Expected: \texttt{Yes}.\\
Q5 (Anomaly Absence): Does any unexpected object, entity, or visual anomaly unrelated to `[ACTION]' appear in this segment? \emph{Natural consequences of the event and minor AI rendering artifacts do not count.} Expected: \texttt{No}.
\end{tcolorbox}

\begin{figure}[H]
\centering
\includegraphics[width=\linewidth]{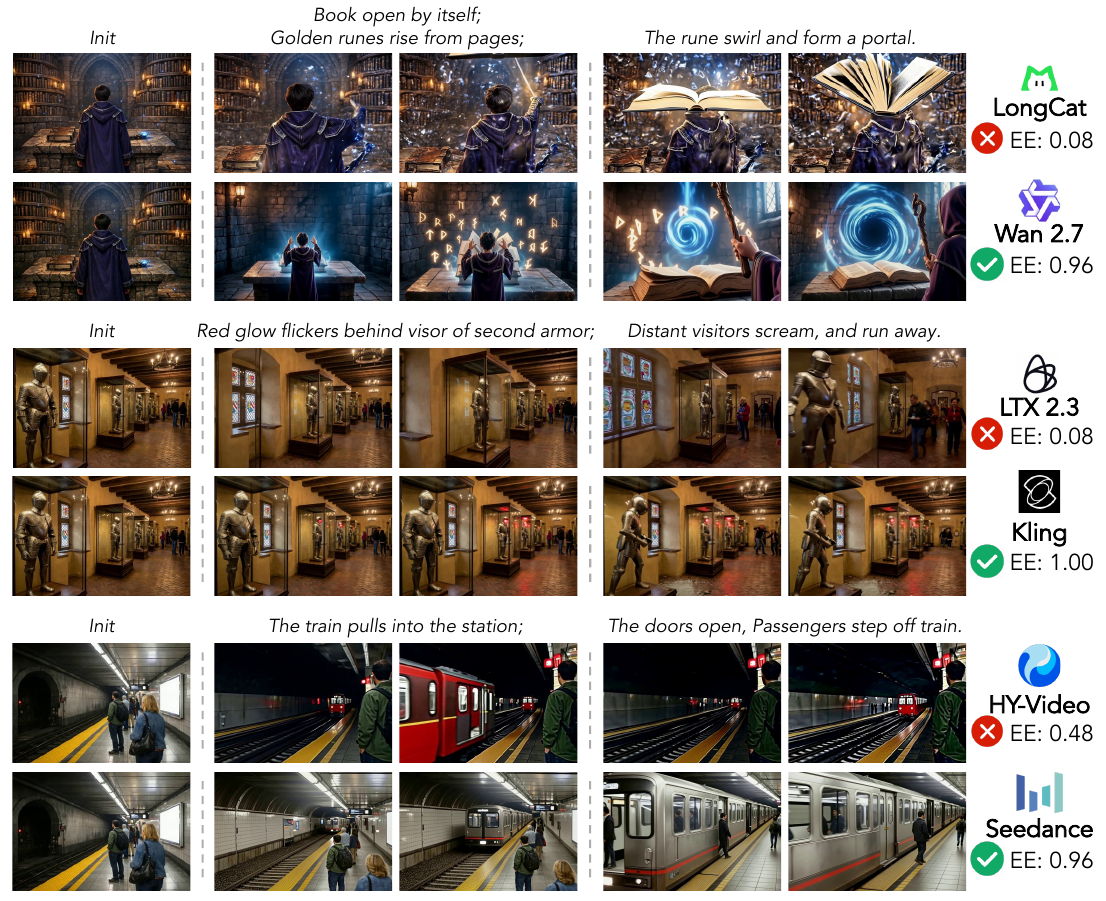}
\caption[Qualitative comparisons on event editing]{Qualitative comparisons on event editing.}
\label{fig:qual_ee}
\end{figure}

\myparagraph{Qualitative Results.}
\cref{fig:qual_ee} contrasts representative outputs across models on two event-edit turns. High-scoring cases exhibit a clear state transition aligned with the instructed event (matching Q2--Q3), while low-scoring cases either leave the scene untouched (Q1 fires) or introduce unrelated entities that trigger the anomaly check (Q5).

\subsubsection{Subject Action Adherence}
\label{app:subject_action_detail}

\myparagraph{Definition.}
Subject action adherence uses the same shared system prompt and five-question scoring rule as event editing adherence, but with question templates tuned to whether the subject performs the instructed action. The case score is the mean over all subject-action turns.

\myparagraph{Prompt.} We show the prompt below.

\begin{tcolorbox}[d3box, title={Subject Action Adherence (Q1--Q5)}]
\ttfamily\scriptsize
World setting: [SCENE\_DESCRIPTION]\\
Interaction instruction for this turn: [ACTION]\\
\\
The following frames are uniformly sampled from the generated video segment for this turn.\\
\mbox{[FRAME\_SEQUENCE]}\\
\\
Q1 (Idle Subject Check): Does the subject remain idle or stationary, with no attempt to perform `[ACTION]'? \emph{Answer No if the subject shows ANY movement or gesture related to this action.} Expected: \texttt{No}.\\
Q2 (Action Occurrence): Does the video show the subject performing something resembling or consistent with `[ACTION]'? \emph{Partial execution counts as Yes.} Expected: \texttt{Yes}.\\
Q3 (Action Completion): By the end of this segment, has the subject's action `[ACTION]' reached a clear conclusion or outcome? Expected: \texttt{Yes}.\\
Q4 (Detail Accuracy): Are the key details of `[ACTION]' accurately depicted, such as the correct subject, target objects, and manner of the action? Expected: \texttt{Yes}.\\
Q5 (Unnatural Motion Check): Does the subject exhibit any unnatural body movement, impossible pose, or physically implausible interaction while performing `[ACTION]'? \emph{Only clearly unnatural movements or impossible poses count; slight jitter or minor hand artifacts do not.} Expected: \texttt{No}.
\end{tcolorbox}

\begin{figure}[H]
\centering
\includegraphics[width=\linewidth]{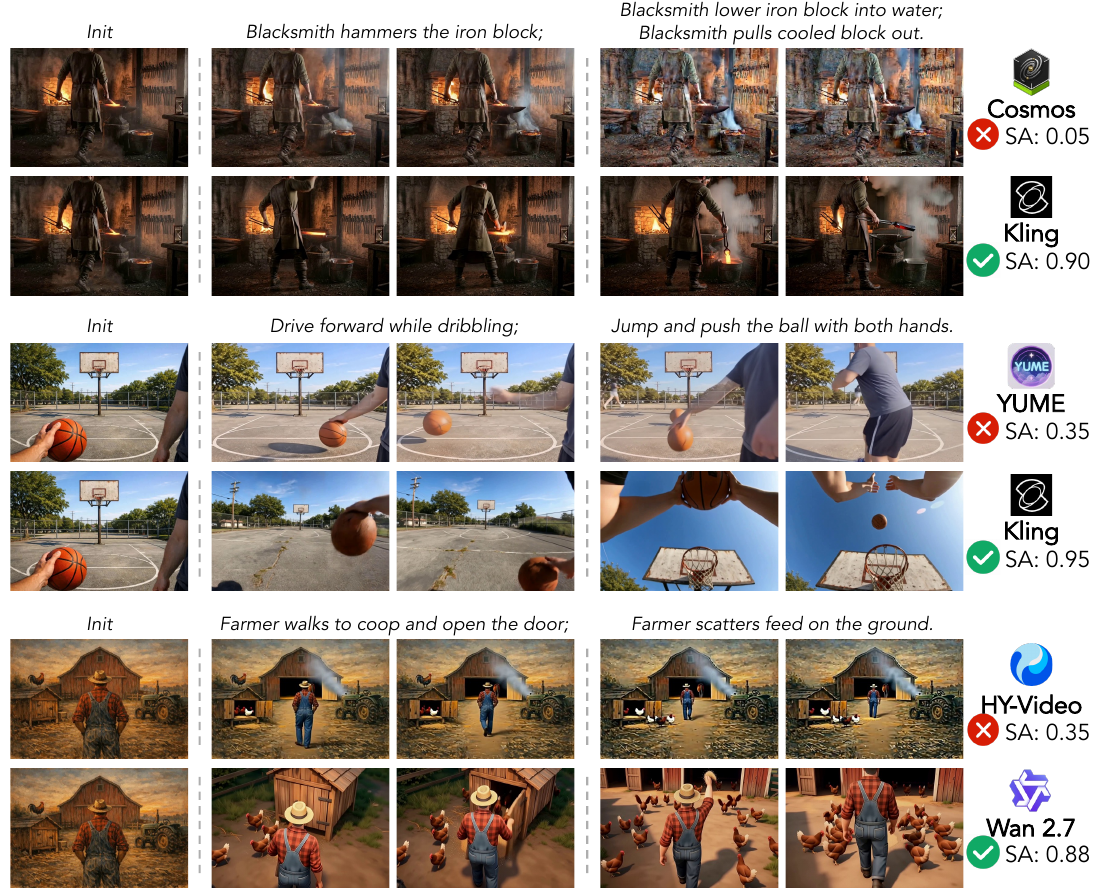}
\caption[Qualitative comparisons on subject action]{Qualitative comparisons on subject action.}
\label{fig:qual_sa}
\end{figure}

\myparagraph{Qualitative Results.}
\cref{fig:qual_sa} illustrates how subject-action adherence differentiates models on the same instruction. Successful generations show the subject initiating and completing the prescribed action successfully, whereas failure cases leave the subject idle or execute only a partial gesture that never reaches a clear end state.

\subsubsection{Perspective Switching Adherence}
\label{app:persp_switch_detail}

\myparagraph{Definition.}
Perspective switching adherence evaluates whether the model can transition between first-person and third-person views. Three binary questions are asked per turn. A turn receives a score of 1 only if all three answers are \texttt{Yes} ($Q1 \wedge Q2 \wedge Q3$), and 0 otherwise.

\myparagraph{Prompt.} We provide the prompt below. We provide the prompt below.

\begin{tcolorbox}[d3box, title={Perspective Switching Adherence (Q1--Q3)}]
\textbf{System:}
\begin{tcolorbox}[systemprompt]
You are a precise video evaluator. You will be given frames from the beginning and end of a generated video turn and a perspective-switching instruction. Answer each question with Yes or No only.
\end{tcolorbox}
\vspace{4pt}
\textbf{User:}
\begin{tcolorbox}[userprompt]
Perspective switching instruction: [SWITCH\_INSTRUCTION]\\
(e.g., ``Switch from first-person to third-person view following the character'')\\
Source perspective: [SOURCE\_PERSPECTIVE] (first-person / third-person)\\
Target perspective: [TARGET\_PERSPECTIVE] (first-person / third-person)\\
\\
Early frames (beginning of the turn segment):\\
\mbox{[EARLY\_FRAME\_SEQUENCE]}\\
Late frames (end of the turn segment):\\
\mbox{[LATE\_FRAME\_SEQUENCE]}\\
\\
Answer each question with Yes or No only.\\
\\
Q1 (Transition Visibility): Can a clear change in perspective be observed between the early and late frames?\\
Q2 (Target Consistency): Does the late-frame perspective match the target perspective ([TARGET\_PERSPECTIVE])?\\
Q3 (Quality Compliance): Does the new perspective satisfy expected structural properties? For third-person, is the subject centered with a following camera? For first-person, are there no self-view artifacts?
\end{tcolorbox}
\end{tcolorbox}

\begin{figure}[H]
\centering
\includegraphics[width=\linewidth]{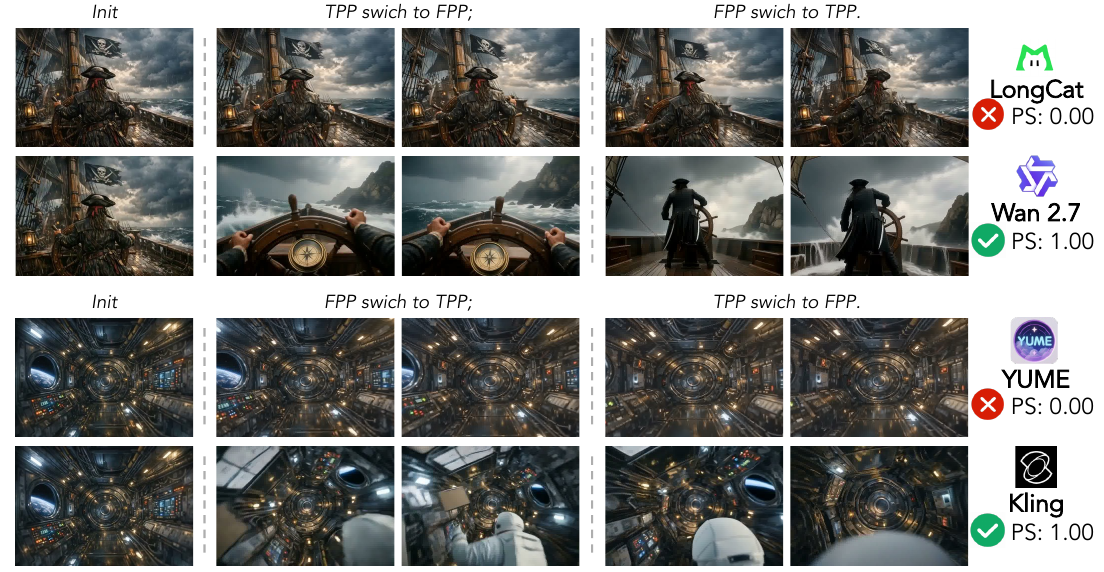}
\caption[Qualitative comparisons on perspective switching]{Qualitative comparisons on perspective switching.}
\label{fig:qual_ps}
\end{figure}

\myparagraph{Qualitative Results.}
\cref{fig:qual_ps} shows some text-driven model cases on the same switching instruction. Successful cases satisfy all three checks with a clear perspective change, a scene matching, and a structurally clean framing, while failures either preserve the source perspective.

\subsection{Consistency}
\label{app:consistency_details}

All seven consistency sub-metrics measure frame-level temporal coherence within each generated clip.

\subsubsection{Subject Consistency}
\label{app:subject_consistency_detail}

\myparagraph{Definition.}
For each frame with a valid SAM2 mask (area $\geq 10$\,px), we crop the subject region (background filled with gray) and extract DINOv2 ViT-B/14 and CLIP ViT-B/16 features.
We compute two similarity signals:
(i) DINOv2 adjacent-frame cosine similarity $s_i^{\mathrm{dino}} = \cos(\mathbf{f}_{i-1}, \mathbf{f}_i)$, and
(ii) CLIP first-frame anchored similarity $s_i^{\mathrm{clip}} = \cos(\mathbf{f}_0, \mathbf{f}_i)$.
The per-frame score is $(s_i^{\mathrm{dino}} + s_i^{\mathrm{clip}})/2$, averaged across all valid frames.

\subsubsection{Background Consistency}
\label{app:background_consistency_detail}

\myparagraph{Definition.}
Following VBench, we extract CLIP ViT-B/32 features from each frame and compute the mean cosine similarity between consecutive frame pairs: $\mathrm{score} = \frac{1}{N-1}\sum_{i=1}^{N-1} \cos(\mathbf{f}_i, \mathbf{f}_{i+1})$.

\subsubsection{Spatial Consistency}
\label{app:spatial_detail}

\begin{figure}[H]
\centering
\includegraphics[width=\linewidth]{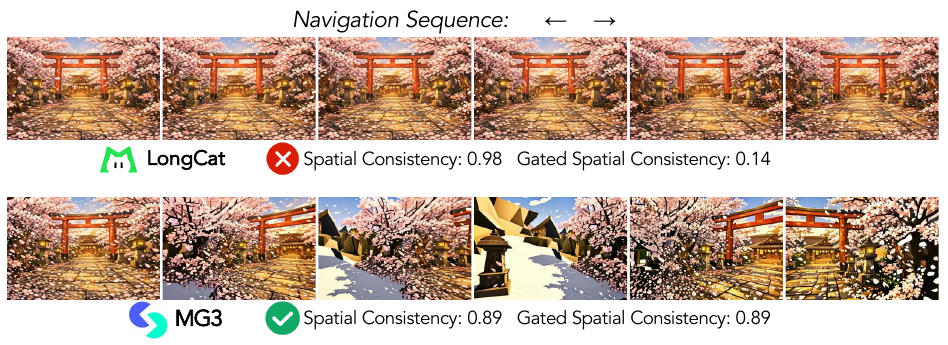}
\caption[Qualitative comparisons on spatial consistency]{Qualitative comparisons on spatial consistency. The gated score penalises the static scene.}
\label{fig:qual_spatial}
\end{figure}

\myparagraph{Definition.}
For roundtrip trajectories, we use MegaSaM-estimated camera poses to identify the \emph{return frame} in the final turn, i.e., the frame whose rotation matrix $R_k$ minimizes $\arccos\!\bigl((\mathrm{tr}(R_0^\top R_k) - 1)/2\bigr)$ relative to the first frame.
Let $s_{\mathrm{ret}}$ be the DreamSim similarity between the first frame and this return frame, computed as $1/(1+d)$ where $d$ is the DreamSim distance.
We uniformly sample 10 intermediate frames and compute the minimum similarity $s_{\mathrm{min}}$ to the first frame.
A motion gate prevents trivially high scores from near-static videos:
\begin{equation}
  \mathrm{score} = s_{\mathrm{ret}} \cdot \min\!\Bigl(1,\; \frac{1 - s_{\mathrm{min}}}{\tau}\Bigr), \quad \tau = 0.15.
\end{equation}
\myparagraph{Qualitative Results.}
\cref{fig:qual_spatial} illustrates the motion gate at work. The near-static return frame genereted by LongCat-Video earns a high $s_{\mathrm{ret}}$ but is penalised by the gated factor. Genuinely revisited model receives a higher gated spatial consistency score.

\subsubsection{Segment Continuity}
\label{app:segment_continuity_detail}

\myparagraph{Definition.}
TransNetV2 predicts per-frame scene-boundary probabilities.
Frames exceeding a confidence threshold of $0.5$ are flagged as cut candidates, with a minimum scene length of 10 frames to suppress spurious detections.
Each video receives a binary score: $1$ if no cut is detected, $0$ otherwise.
The model-level score is the fraction of cut-free videos: $\mathrm{score} = 1 - n_{\mathrm{cut}} / N$.

\subsubsection{Perspective Consistency}
\label{app:perspective_consistency}

\myparagraph{Definition.}
For each frame where the SAM2-tracked target is present (mask area $\geq 10$\,px), we record the normalized centroid $(c_x, c_y)$ within the mask region.
Let $p = n_{\text{valid}} / n_{\text{total}}$ denote the target presence rate.
The score measures how stably the subject remains positioned across frames:
\begin{equation}
  s_{\text{centroid}} = \max\!\Bigl(0,\; 1 - \frac{\sqrt{\sigma_{c_x}^2 + \sigma_{c_y}^2}}{0.3}\Bigr) \times p,
\end{equation}
where $\sigma_{c_x}$ and $\sigma_{c_y}$ are the standard deviations of the normalized centroid coordinates over valid frames.
The presence weighting ensures that models where the target frequently disappears receive lower scores, while the threshold constant ($0.3$) is set so that moderate drift already incurs significant penalty.

\subsubsection{Reconstruction Consistency}
\label{app:reconstruction_detail}

\myparagraph{Definition.}
Depth Anything 3~\cite{lin2025depthanything3} jointly estimates per-frame depth maps and camera poses.
We evaluate 3D coherence via two complementary signals that share the same depth-based reprojection pipeline.

For a pixel $\mathbf{u}_i = (u_i, v_i, 1)^\top$ in frame $i$ with depth $d_i(\mathbf{u}_i)$, we first back-project it to 3D in camera $i$,
\begin{equation}
  \mathbf{X}_i = d_i(\mathbf{u}_i) K^{-1} \mathbf{u}_i,
\end{equation}
where $K$ is the camera intrinsic matrix.
Using the relative pose from frame $i$ to frame $j$, we transform the point into camera $j$,
\begin{equation}
  \mathbf{X}_j = R_{ji} \mathbf{X}_i + \mathbf{t}_{ji},
\end{equation}
and project it onto frame $j$,
\begin{equation}
  \hat{\mathbf{u}}_j = \pi(K \mathbf{X}_j),
\end{equation}
where $\pi([x, y, z]^\top) = (x/z, y/z)$ denotes perspective projection.

\emph{Geometric consistency} measures the reprojection displacement between the projected location $\hat{\mathbf{u}}_j$ and its matched target location $\mathbf{u}_j$ in frame $j$.
For each valid pair, we normalize the displacement by the image diagonal $D = \sqrt{H^2 + W^2}$ and compute
\begin{equation}
  e_{\mathrm{rel}}(\mathbf{u}_i) = \frac{\lVert \hat{\mathbf{u}}_j - \mathbf{u}_j \rVert_2}{D},
  \qquad
  \bar{e}_{\mathrm{rel}} = \frac{1}{N} \sum_{n=1}^{N} e_{\mathrm{rel}}(\mathbf{u}_i^{(n)}),
\end{equation}
where $N$ is the number of valid projected points.
The geometric consistency score is then
\begin{equation}
  s_{\mathrm{geo}} = \frac{1}{1 + \bar{e}_{\mathrm{rel}}}.
\end{equation}

\emph{Photometric consistency} warps the RGB value of frame $i$ to the projected location in frame $j$ using the same reprojection mapping.
Let $\hat{I}_{i \rightarrow j}$ denote the warped image and $I_j$ denote the observed frame.
We evaluate their agreement using PSNR,
\begin{equation}
  s_{\mathrm{photo}} = \mathrm{PSNR}(\hat{I}_{i \rightarrow j}, I_j)
  = 10 \log_{10} \frac{255^2}{\mathrm{MSE}(\hat{I}_{i \rightarrow j}, I_j)}.
\end{equation}

The two signals are complementary.
Geometric consistency is sensitive to structural distortion caused by incorrect depth or pose, while photometric consistency captures appearance instability such as texture flickering or color shift after geometric alignment.
Our formulation shares the reprojection-based evaluation philosophy with recent video generation alignment methods. VGGRPO~\cite{an2026vggrpo} computes a depth-space reprojection reward by comparing rendered depth maps with predicted depths to enforce cross-view geometric coherence, while VideoGPA~\cite{du2026videogpa} constructs a pixel-space reprojection signal by rendering a colored point cloud back into each view and measuring MSE and LPIPS against the original frames. In contrast, we unify both geometric and photometric evaluation within a single depth-based reprojection pipeline, using displacement error for structural fidelity and PSNR for appearance stability.
Compared with WorldScore~\cite{duan2024worldscore}, we keep both terms within the same depth-based reprojection framework.

\subsection{Physical}
\label{app:physical_detail}

The physical dimension is evaluated via two sub-metrics: causal fidelity (VLM-based) and visual plausibility (fine-tuned Qwen3VL).

\subsubsection{Causal Fidelity}
\label{app:causal_fidelity_detail}

\myparagraph{Definition.}
Causal fidelity uses \texttt{Doubao-Seed-2.0-lite} with video frames sampled at 3\,fps.
It is scored along two complementary tracks, both on a 0--3 scale. Track~1 assigns a global score for rendering-level physics and causal consistency. Track~2 averages 0--3 scores over a case-specific subset of seven physics dimensions, selected in a text-only step so that the same dimensions are evaluated across all models. When Track~2 is applicable, the per-turn score is $(s_{\text{track1}} + \bar{s}_{\text{track2}})/2$; otherwise it falls back to $s_{\text{track1}}$. The case score is the mean over all evaluated turns and lies in $[0, 3]$, and is subsequently normalised to $[0, 1]$.

\myparagraph{Prompt.} The used prompots are shown as below.

\begin{tcolorbox}[d5box, title={Causal Fidelity --- Track 1: Rendering Physics \& Causal Consistency (0--3)}]
\textbf{System:}
\begin{tcolorbox}[systemprompt]
You are evaluating the RENDERING-LEVEL visual plausibility and causal consistency of an AI-generated video segment. You will be shown a sequence of frames in temporal order.\\
\\
IMPORTANT RULES:\\
- Do NOT judge whether depicted actions are possible in the real world. Fantasy / sci-fi elements are acceptable if the setting calls for them.\\
- Do NOT penalize camera behaviour (tilting, shaking, panning, zooming, rotation). Camera motion is a cinematography choice, NOT a physics issue of the world.\\
- ONLY evaluate the physics of OBJECTS and CHARACTERS within the scene.\\
\\
Before scoring, mentally scan the frames for notable events, unexplained appearances / disappearances, and whether every visible effect has a plausible cause and every depicted action produces its expected consequences.\\
\\
Evaluate the following aspects:\\
\emph{A.~Rendering physics.} (1) solid-body integrity, (2) consistent gravity, (3) motion continuity, (4) object permanence, (5) natural deformation, (6) character physics.\\
\emph{B.~Causal consistency.} (7) no effect without cause (for non-instructed changes), (8) no cause without effect (the instructed action must produce its consequences), (9) no unrelated objects or effects appearing from nothing. The instructed action and its direct consequences are EXPECTED changes and must NOT be penalised.
\end{tcolorbox}
\vspace{4pt}
\textbf{User:}
\begin{tcolorbox}[userprompt]
Scene context: [SCENE\_DESC]\\
The instructed action for this segment is: `[ACTION]'.\\
Special physics rule: [RULE\_DESC]\\
\\
\mbox{[FRAME\_SEQUENCE]}\\
\\
First, note any notable events, state changes, or objects appearing / disappearing across the frames. Then rate the overall rendering physics AND causal consistency of this segment from 0 to 3:\\
  3 -- Good: physics and causality are natural and consistent, no significant artifacts or causal errors.\\
  2 -- Fair: noticeable problems such as some clipping, jerky motion, or 1--2 causal errors, but main actions still make sense.\\
  1 -- Poor: multiple physics violations and / or causal breakdowns, objects appear from nowhere, frequent clipping, missing consequences.\\
  0 -- Failure: physics and causality are mostly or completely broken.
\end{tcolorbox}
\end{tcolorbox}

\vspace{6pt}

\begin{tcolorbox}[d5box, title={Causal Fidelity --- Track 2: Scene-Aware Physics Dimensions (0--3, averaged)}]
\textbf{Stage 2A (dimension selection, text-only; precomputed once per case):}
\begin{tcolorbox}[systemprompt]
You are a physics evaluation assistant for AI-generated videos. Given scene context, environment details, and an action description, identify which specific physics phenomena are relevant and should be evaluated. Consider BOTH the action AND the environment, for example rain implies fluid / reflection, snow implies surface tracks, wind implies environmental forces. Select only dimensions that are clearly relevant, but do not miss obvious ones.
\end{tcolorbox}
\vspace{4pt}
\begin{tcolorbox}[userprompt]
Scene description: [SCENE\_DESCRIPTION]\\
Interaction sequence: [FULL\_INTERACTION\_SEQUENCE]\\
Environment details: [ENV\_DETAILS]\\
\\
Select ALL applicable dimensions from:\\
(1) Fluid \& Smoke: liquids, smoke, fire, steam, and underwater behaviour (buoyancy, bubbles, drifting).\\
(2) Collision \& Clipping: objects or characters physically colliding, striking, pushing, or overlapping. Do NOT select for scenes without physical contact.\\
(3) Surface Tracks \& Imprints: movement over soft, deformable surfaces (snow, sand, mud, wet ground, dust). Do NOT select for hard surfaces.\\
(4) Deformation \& Destruction: breaking, shattering, crumbling, tearing, or bending, including explosion debris.\\
(5) Wind \& Environmental Forces: wind acting on hair, cloth, leaves, particles; currents and air resistance.\\
(6) Reflection \& Lighting: water / mirror reflections, explicit point lights, directional shadows, metallic / glass reflections, or fire / lava illumination.\\
(7) Human Motion \& Expression: clearly visible human or humanoid figures whose body motion and face coherence can be judged across frames.
\end{tcolorbox}
\vspace{6pt}
\textbf{Stage 2B (shared system prompt and user template for per-dimension scoring):}
\begin{tcolorbox}[systemprompt]
You are evaluating a specific physics phenomenon in an AI-generated video segment. Score based ONLY on the specified physics dimension, not on overall video quality. These are AI-generated videos; minor rendering artifacts are acceptable. Focus on whether the specific physical behaviour is plausible.
\end{tcolorbox}
\vspace{4pt}
\begin{tcolorbox}[userprompt]
World setting: [SCENE\_DESCRIPTION]\\
\\
\mbox{[FRAME\_SEQUENCE]}\\
\\
Evaluate dimension [DIM\_ID]: [DIM\_NAME].\\
Description: [DIM\_DESCRIPTION]\\
\mbox{[DIM\_SCORING\_PROMPT]}
\end{tcolorbox}
\vspace{4pt}
The seven dimension-specific \texttt{[DIM\_SCORING\_PROMPT]} templates are reproduced verbatim below.
\end{tcolorbox}

\vspace{6pt}

\begin{tcolorbox}[d5box, title={Track 2 Rubric --- Fluid \& Smoke}]
\ttfamily\scriptsize
Rate fluid, smoke, and related effects (0--3).\\
3 = Realistic flow, splash, smoke rise, underwater drift.\\
2 = Minor artifacts but behaviour is recognisable.\\
1 = Notable issues: liquid defies gravity, smoke static, no bubbles.\\
0 = Fluid / smoke behaviour completely unrealistic or absent, or completely unrealistic.
\end{tcolorbox}

\vspace{4pt}

\begin{tcolorbox}[d5box, title={Track 2 Rubric --- Collision \& Clipping}]
\ttfamily\scriptsize
Rate collision and solid-body integrity (0--3). ONLY evaluate collisions / contacts that actually occur.\\
3 = All contacts are natural, no clipping or pass-through.\\
2 = Minor clipping but mostly plausible contacts.\\
1 = Obvious clipping: limbs pass through objects or surfaces.\\
0 = Severe clipping, objects completely ignore each other.
\end{tcolorbox}

\vspace{4pt}

\begin{tcolorbox}[d5box, title={Track 2 Rubric --- Surface Tracks \& Imprints}]
\ttfamily\scriptsize
Rate surface interaction and imprint accuracy (0--3).\\
3 = Clear, appropriate marks visible (footprints in snow, tracks in sand).\\
2 = Some marks visible but incomplete or inconsistent.\\
1 = No marks visible on a surface that should clearly show them.\\
0 = Surface shows no response at all to contact.
\end{tcolorbox}

\vspace{4pt}

\begin{tcolorbox}[d5box, title={Track 2 Rubric --- Deformation \& Destruction}]
\ttfamily\scriptsize
Rate deformation and destruction effects (0--3).\\
3 = Realistic shattering, crumbling, bending, debris scatter.\\
2 = Minor issues but deformation is recognisable.\\
1 = Deformation clearly unnatural or inconsistent.\\
0 = No deformation when expected, or completely unrealistic.
\end{tcolorbox}

\vspace{4pt}

\begin{tcolorbox}[d5box, title={Track 2 Rubric --- Wind \& Environmental Forces}]
\ttfamily\scriptsize
Rate wind and environmental force effects (0--3).\\
3 = Natural swaying, drifting, fluttering of affected objects.\\
2 = Minor issues but environmental effects are recognisable.\\
1 = Effects are stiff, static, or inconsistent with the scene.\\
0 = No environmental effects visible when expected.
\end{tcolorbox}

\vspace{4pt}

\begin{tcolorbox}[d5box, title={Track 2 Rubric --- Reflection \& Lighting}]
\ttfamily\scriptsize
Rate the accuracy of SPECIAL reflection and lighting effects (0--3). Evaluate ONLY optical effects that are actually VISIBLE in the video; do NOT penalise for effects that are absent from the scene.\\
Check for:\\
- Water / mirror reflections: do they match the shape and motion of the source?\\
- Point light sources: is the illumination range and falloff plausible?\\
- Shadows: do they point in the correct direction relative to the light?\\
- Metallic / glass surfaces: do they show plausible environment reflections?\\
\\
3 = All visible optical effects are accurate and physically consistent.\\
2 = Minor errors in one effect (e.g. reflection slightly misaligned).\\
1 = Clear errors: reflections show wrong content, shadows point wrong way, or lighting does not respond to scene changes.\\
0 = Optical effects are completely wrong or absent when clearly expected.
\end{tcolorbox}

\vspace{4pt}

\begin{tcolorbox}[d5box, title={Track 2 Rubric --- Human Motion \& Expression}]
\ttfamily\scriptsize
Rate human motion naturalness and facial expression quality (0--3). Focus on MOVEMENT ACROSS FRAMES, not just single-frame anatomy:\\
- Does the character move fluidly between poses, or jerk / teleport?\\
- Are body movements biomechanically plausible (natural gait, realistic reach, believable weight shifts)?\\
- Do limbs move in coordinated, purposeful ways, or flail randomly?\\
- Is the face coherent (no melting, distortion, or uncanny shifts)?\\
Minor structural issues (slight hand distortion, brief extra finger) should NOT heavily impact the score if overall motion is natural.\\
\\
3 = Smooth, natural human motion throughout; face is coherent.\\
2 = Mostly natural motion with minor stiffness or brief glitches.\\
1 = Clearly unnatural: jerky limbs, puppet-like movement, frozen poses, or obvious face distortion.\\
0 = Severely broken: limbs flail impossibly, body contorts unnaturally, face melts or deforms grotesquely.\\[4pt]
\rmfamily\normalsize
\textit{\small Track 2 score $\bar{s}_{\text{track2}}$ is the arithmetic mean of the 0--3 scores over all selected dimensions. Dimension selection is precomputed once per case, manually verified, and shared across all models to ensure comparability. When at least one dimension is selected, the per-turn causal fidelity score is $s = (s_{\text{track1}} + \bar{s}_{\text{track2}}) / 2$; otherwise it falls back to $s_{\text{track1}}$. The per-case score is the mean over all evaluated turns and lies in $[0, 3]$, then normalised to $[0, 1]$.}
\end{tcolorbox}

\begin{figure}[H]
\centering
\includegraphics[width=0.88\linewidth]{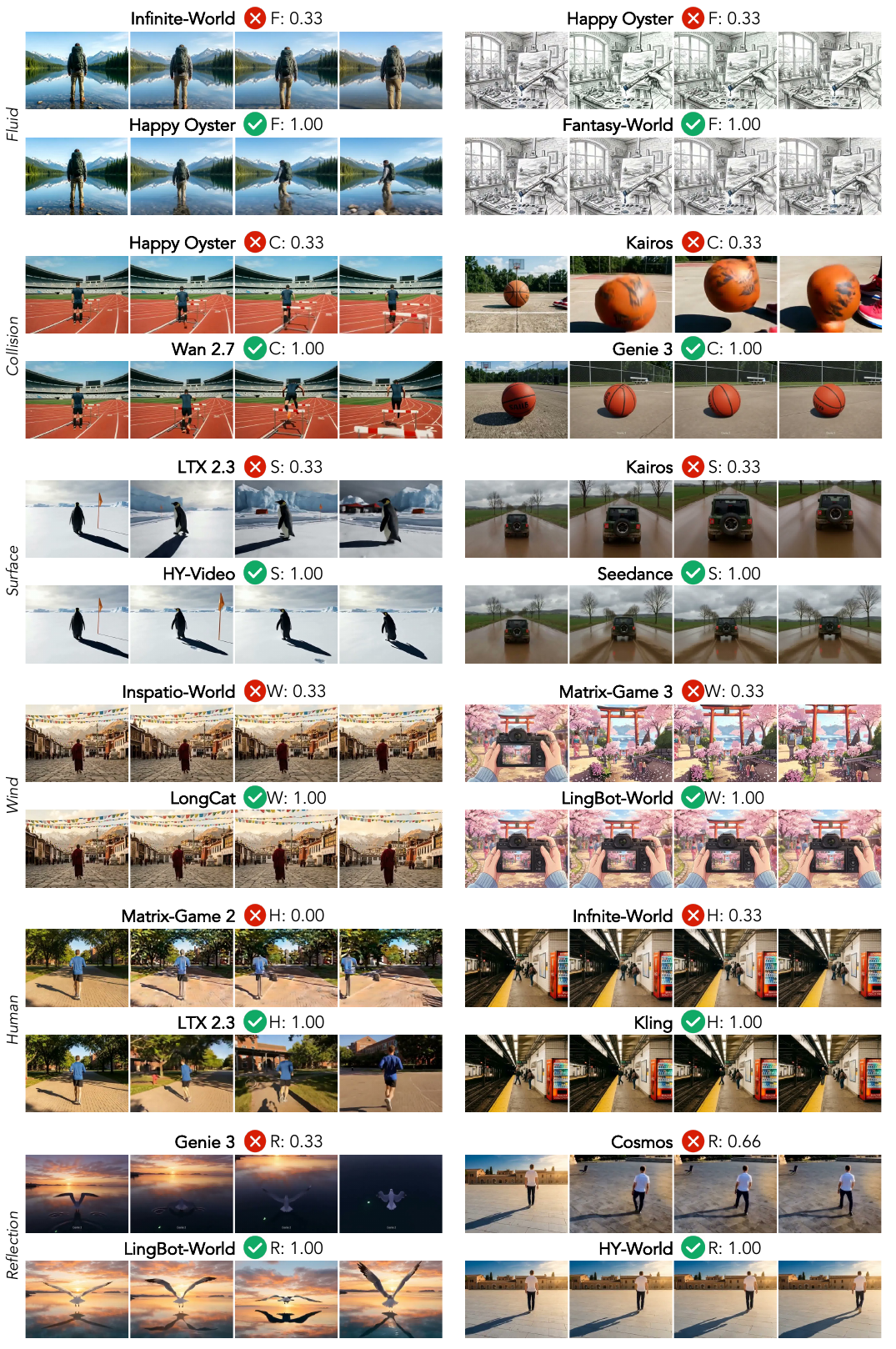}
\caption[Qualitative comparisons on physics compliance]{Qualitative comparisons on physics compliance.}
\label{fig:qual_pf}
\end{figure}

\myparagraph{Qualitative Results.}
\cref{fig:qual_pf} shows the comparison of good and bad cases, generated by different models on each sub-dimension evaluated in Track 2.

\myparagraph{Results.}
\cref{tab:track2_per_case_activation} lists every one of the \textbf{50} cases that receive Track 2 scoring, together with a short scene description and the seven dimension flags. A \cmark\ indicates that the dimension is included in that case's Stage 2B scoring pool, and a \xmark\ indicates that it is skipped.

\begin{figure*}[!htbp]
  \centering
  \includegraphics[width=\linewidth]{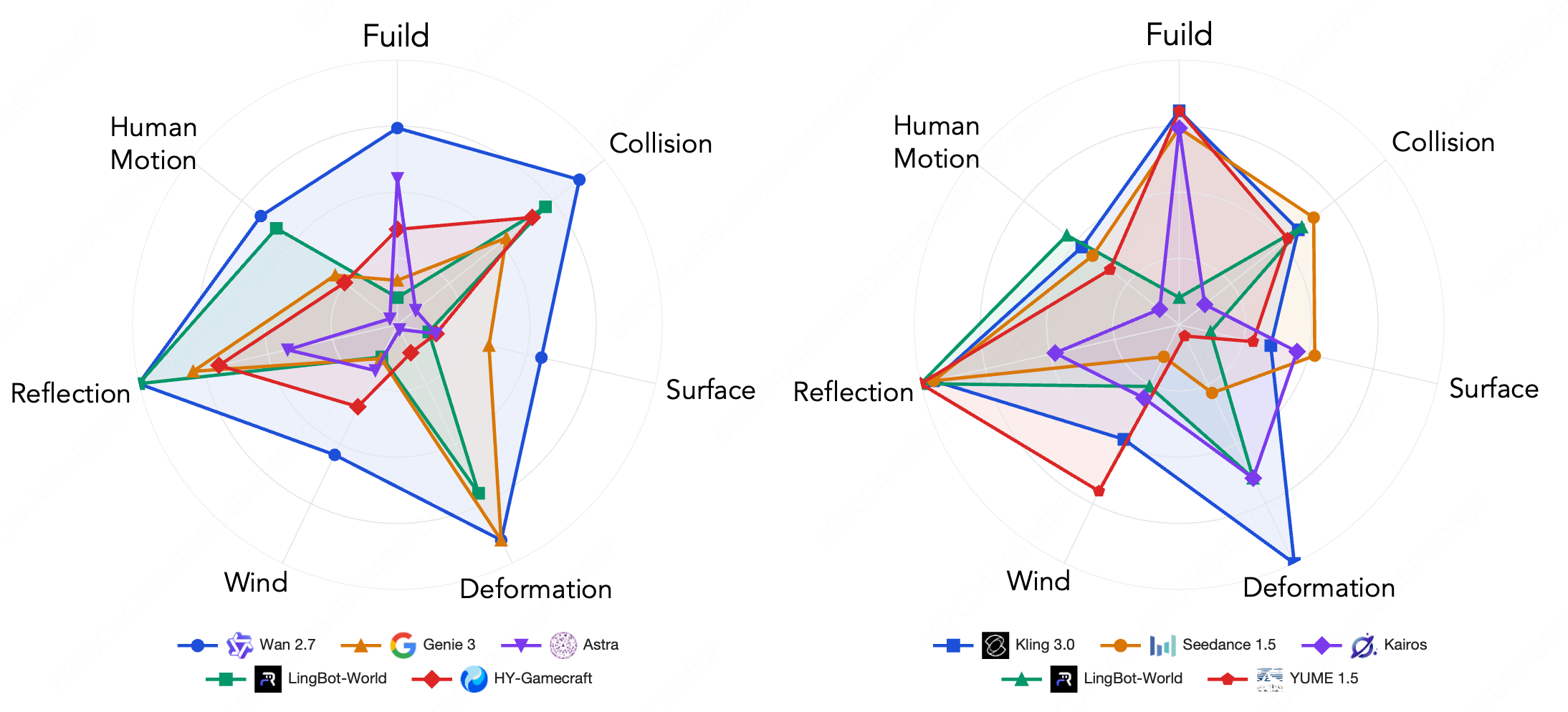}
  \caption[Causal Fidelity decomposed by Track 2 sub-dimensions]{Causal Fidelity decomposed along the seven Track~2 sub-dimensions. \textit{Left:} a top-to-bottom tier across paradigms. \textit{Right:} models at a similar overall tier that trade off strengths across sub-dimensions. Each axis is rescaled to its own visible range to make the per-sub-dimension differences easier to read.}
  \label{fig:radar_physics_per_dim}
\end{figure*}

\Cref{fig:radar_physics_per_dim} further decomposes Track~2 causal fidelity across the seven physics sub-dimensions. The results reveal heterogeneous failure modes across models. Reflection~\& Lighting is nearly saturated for most models, while Deformation~\& Destruction shows much larger variation despite being activated in only $4$ of the $50$ Track~2 cases, suggesting that destructive event dynamics remain challenging. Tier-wise gaps are also axis-dependent: strong models such as Wan~2.7 lead on most dimensions, but their advantage is much larger on Deformation and Human Motion than on Reflection, indicating that aggregate scores can obscure the underlying failure axes. Moreover, models with similar overall performance exhibit distinct trade-offs. Flagship text-driven systems are stronger on Fluid and Collision, LingBot-World performs best on Human Motion and Reflection but lags on Fluid and Surface, and YUME~1.5 stands out on Wind, consistent with its outdoor navigation-oriented fine-tuning. These patterns show that causal fidelity is a multi-dimensional capability, where similar overall scores may reflect qualitatively different physics profiles.

\begin{table*}[!htbp]
\centering
\scriptsize
\setlength{\tabcolsep}{3pt}
\renewcommand{\arraystretch}{1.05}
\begin{tabular}{@{}l p{0.55\linewidth} ccccccc c@{}}
\toprule
\textbf{ID} & \textbf{Scene description} & \textbf{F} & \textbf{C} & \textbf{S} & \textbf{D} & \textbf{W} & \textbf{R} & \textbf{H} & \textbf{\#Dims} \\
\midrule
1   & blocky game world, Mario jumping over pipes and blocks                & \xmark & \cmark & \xmark & \xmark & \xmark & \xmark & \cmark & 2 \\
2   & foggy British street at night, handheld mockumentary view             & \xmark & \cmark & \xmark & \xmark & \xmark & \cmark & \cmark & 3 \\
3   & rainy anime city street, young woman walking amid neon                & \cmark & \cmark & \cmark & \xmark & \xmark & \cmark & \cmark & 5 \\
4   & realistic soccer pitch, player tracked in third person                & \xmark & \cmark & \xmark & \xmark & \xmark & \xmark & \cmark & 2 \\
5   & crowded anime subway interior, camera swaying with train              & \xmark & \cmark & \xmark & \xmark & \xmark & \cmark & \cmark & 3 \\
6   & living room with clothing rack, heated iron on table                  & \xmark & \cmark & \xmark & \cmark & \xmark & \cmark & \xmark & 3 \\
7   & busy urban street, dark rain cloud above pedestrians                  & \xmark & \cmark & \cmark & \xmark & \xmark & \cmark & \cmark & 4 \\
8   & sterile chemistry lab, glassware and pressure vial under lights       & \xmark & \cmark & \xmark & \xmark & \xmark & \cmark & \cmark & 3 \\
9   & Olympic red track, athlete approaching hurdles in lane                & \xmark & \cmark & \xmark & \xmark & \xmark & \xmark & \cmark & 2 \\
10  & Antarctic ice sheet, emperor penguin walking on snow                  & \xmark & \xmark & \cmark & \xmark & \cmark & \xmark & \cmark & 3 \\
11  & Tibetan monastery courtyard at golden hour, maroon-robed monk         & \xmark & \cmark & \xmark & \xmark & \cmark & \cmark & \cmark & 4 \\
12  & autumn park with fallen leaves, toddler in red running                & \xmark & \cmark & \xmark & \xmark & \cmark & \xmark & \cmark & 3 \\
13  & African savanna at sunset, elephant on dusty trail                    & \cmark & \xmark & \cmark & \xmark & \cmark & \cmark & \xmark & 4 \\
14  & white office desk, small plush keychain with green accents            & \xmark & \cmark & \xmark & \cmark & \xmark & \xmark & \xmark & 2 \\
15  & Mediterranean stone plaza, young man walking in sunlight              & \xmark & \cmark & \xmark & \xmark & \xmark & \cmark & \cmark & 3 \\
16  & mountain lake shore at dawn, hiker with green backpack                & \cmark & \cmark & \cmark & \xmark & \xmark & \cmark & \cmark & 5 \\
17  & muddy rural road after rain, dark green off-road jeep                 & \cmark & \xmark & \cmark & \xmark & \xmark & \cmark & \xmark & 3 \\
18  & African savanna trail, chestnut horse with safari rider               & \xmark & \cmark & \cmark & \xmark & \cmark & \xmark & \cmark & 4 \\
19  & misty Finnish lake at dawn, man rowing wooden boat                    & \cmark & \cmark & \xmark & \xmark & \xmark & \cmark & \cmark & 4 \\
20  & wooden beach boardwalk at golden hour, young man walking              & \cmark & \xmark & \xmark & \xmark & \cmark & \cmark & \cmark & 4 \\
21  & city park walkway, woman in red jacket walking in sun                 & \xmark & \cmark & \xmark & \xmark & \xmark & \cmark & \cmark & 3 \\
22  & seaside stone promenade, young skateboarder with board in hand        & \cmark & \cmark & \xmark & \xmark & \xmark & \cmark & \cmark & 4 \\
23  & tree-lined campus pathway, college student jogging in sun             & \xmark & \cmark & \xmark & \xmark & \xmark & \xmark & \cmark & 2 \\
24  & desert highway shoulder at sunset, backpacker walking                 & \xmark & \xmark & \cmark & \xmark & \xmark & \xmark & \cmark & 2 \\
25  & wide snowy field in winter, person on fat-tire bicycle                & \xmark & \xmark & \cmark & \xmark & \xmark & \xmark & \cmark & 2 \\
26  & hot desert midday, monitor lizard crawling on golden sand             & \xmark & \xmark & \cmark & \xmark & \xmark & \xmark & \xmark & 1 \\
27  & outdoor basketball court, orange ball bouncing on concrete            & \xmark & \cmark & \xmark & \xmark & \xmark & \cmark & \xmark & 2 \\
28  & subway platform, FPP view of vending machine by tiled wall             & \xmark & \cmark & \xmark & \xmark & \xmark & \cmark & \cmark & 3 \\
29  & cozy antique bookstore, brass globe on wooden stand                   & \xmark & \cmark & \xmark & \xmark & \xmark & \cmark & \cmark & 3 \\
30  & bustling Asian night market, red lanterns and moving crowd            & \cmark & \cmark & \xmark & \xmark & \xmark & \cmark & \cmark & 4 \\
31  & science classroom interior, human skeleton model near blackboard      & \xmark & \cmark & \xmark & \xmark & \xmark & \cmark & \cmark & 3 \\
32  & coastal city aerial view, FPP seabird flight with beak                 & \cmark & \cmark & \xmark & \xmark & \xmark & \xmark & \xmark & 2 \\
33  & abandoned warehouse, racing quadcopter flying between pillars         & \xmark & \cmark & \xmark & \xmark & \xmark & \cmark & \xmark & 2 \\
34  & countryside farmland, colorful hot air balloon rising                 & \xmark & \xmark & \xmark & \xmark & \xmark & \xmark & \cmark & 1 \\
35  & calm ocean at sunset, white seagull diving above water                & \cmark & \xmark & \xmark & \xmark & \xmark & \cmark & \xmark & 2 \\
36  & supermarket aisle, tall shelves with stacked red cans                 & \xmark & \cmark & \xmark & \xmark & \xmark & \cmark & \cmark & 3 \\
37  & cherry blossom garden in spring, viewer with camera viewfinder        & \xmark & \cmark & \xmark & \xmark & \cmark & \xmark & \cmark & 3 \\
38  & war-torn city ruins, FPP view holding military grenade                 & \cmark & \cmark & \xmark & \cmark & \cmark & \xmark & \cmark & 5 \\
39  & misty lakeside at dawn, viewer holding bamboo fishing rod             & \cmark & \cmark & \xmark & \xmark & \xmark & \cmark & \cmark & 4 \\
40  & sunlit artist studio, viewer holding paintbrush near canvas           & \xmark & \cmark & \xmark & \xmark & \xmark & \cmark & \cmark & 3 \\
41  & snowy tundra under aurora borealis, green purple lights               & \xmark & \cmark & \cmark & \xmark & \xmark & \cmark & \xmark & 3 \\
42  & floating sky island anime, fairy girl with dragonfly wings            & \cmark & \xmark & \xmark & \xmark & \cmark & \cmark & \cmark & 4 \\
43  & lavender field at golden hour, woman in white, oil-painting style     & \xmark & \xmark & \cmark & \xmark & \cmark & \cmark & \cmark & 4 \\
44  & magical toy workshop, wooden puppet walking among stations            & \xmark & \cmark & \xmark & \xmark & \xmark & \cmark & \cmark & 3 \\
45  & underwater ancient ruins (CG), mermaid among marble columns           & \cmark & \cmark & \cmark & \xmark & \xmark & \cmark & \cmark & 5 \\
46  & moonlit rooftops, ink-wash ninja running across tiles                 & \xmark & \cmark & \xmark & \xmark & \xmark & \xmark & \cmark & 2 \\
47  & Middle Eastern bazaar (flat), fruit merchant pushing cart             & \xmark & \cmark & \xmark & \xmark & \xmark & \xmark & \cmark & 2 \\
48  & surreal melting landscape, brass robot among dripping clocks          & \xmark & \cmark & \cmark & \cmark & \xmark & \xmark & \cmark & 4 \\
49  & spooky haunted garden, glowing ghost dog among jack-o-lanterns        & \xmark & \cmark & \xmark & \xmark & \xmark & \cmark & \xmark & 2 \\
50  & volcanic crater edge (CG), dragon hatchling flying along rim          & \cmark & \xmark & \xmark & \xmark & \cmark & \cmark & \cmark & 4 \\
\midrule
\textbf{Total} & \textbf{(50 cases)} & 15 & 38 & 14 & 4 & 11 & 32 & 39 & 153 \\
\bottomrule
\end{tabular}
\caption[Per-case Track 2 dimension activation]{Per-case Track 2 dimension activation for all \textbf{50} cases that receive Track 2 scoring. Columns: F~= Fluid~\& Smoke, C~= Collision~\& Clipping, S~= Surface Tracks, D~= Deformation~\& Destruction, W~= Wind~\& Environmental Forces, R~= Reflection~\& Lighting, H~= Human Motion~\& Expression. The final row aggregates the activation count of each dimension over all 50 cases, summing to 153 dimension-case pairs (mean $\approx 3.06$ dimensions per case).}
\label{tab:track2_per_case_activation}
\end{table*}

\subsubsection{Visual Plausibility}
\label{app:visual_plausibility_detail}

\myparagraph{Definition.}
Visual plausibility measures whether a generated video is visually realistic and physically reasonable. It focuses on the overall appearance quality, temporal consistency, object structure, motion coherence, and whether the depicted dynamics conform to common-sense physical principles. We evaluate visual plausibility using a fine-tuned Qwen3-VL-30B-A3B model.

\myparagraph{Prompt.}
All videos are evaluated with a fixed prompt. The input video is sampled at 2 FPS, and the maximum number of pixels per frame is set to 602,112. The prompt used for both training and inference is shown below.

\begin{tcolorbox}[d5box, title={Visual Plausibility --- Fine-tuned Qwen3-VL-30B-A3B}]
\ttfamily\scriptsize
Suppose you are an expert in judging and evaluating the quality of AI-generated videos, please watch the above provided video and give scores for the video's truthfulness and rationality, i.e., whether the video's overall appearance and motion are consistent with our common-sense and physical principles.\\
\\
Your rating should be chosen from the following five categories: Perfect, Good, Fair, Poor, and Bad. Now please rate this video:
\end{tcolorbox}

\myparagraph{Training Data.}
To train the visual plausibility scorer, we construct an internal annotation set consisting of approximately 6K videos generated by multiple closed-source and open-source image-to-video models. The videos cover diverse subjects, scenes, visual styles, camera motions, and interaction patterns. Internal expert annotators are first trained with unified rating guidelines, and then each video is independently rated by three annotators on a 1--5 scale, where 5 denotes the highest visual plausibility and 1 denotes the lowest. The final ground-truth score of each video is computed as the average of the three annotator scores.

\myparagraph{Score Prediction and Optimization.}
We conduct full-parameter fine-tuning upon the pre-trained Qwen3-VL-30B-A3B checkpoint~\footnote{\url{https://huggingface.co/Qwen/Qwen3-VL-30B-A3B-Instruct.}} to regress the human-annotated visual plausibility score. Given a video and the above prompt, we extract the next-token prediction distribution at the final prompt token. Let the five rating tokens be
\(\{\texttt{Perfect}, \texttt{Good}, \texttt{Fair}, \texttt{Poor}, \texttt{Bad}\}\),
corresponding to scores \(\{5,4,3,2,1\}\), respectively. We take the model probabilities assigned to these five tokens and renormalize them over the rating set:
\[
\tilde{p}_c =
\frac{p_c}{\sum_{c' \in \mathcal{C}} p_{c'}},
\quad
\mathcal{C}=\{\texttt{Perfect}, \texttt{Good}, \texttt{Fair}, \texttt{Poor}, \texttt{Bad}\}.
\]
The predicted visual plausibility score is then computed as the expectation over the five ordered categories:
\[
\hat{s}
=
5\tilde{p}_{\texttt{Perfect}}
+4\tilde{p}_{\texttt{Good}}
+3\tilde{p}_{\texttt{Fair}}
+2\tilde{p}_{\texttt{Poor}}
+1\tilde{p}_{\texttt{Bad}}.
\]
Given the ground-truth human score \(s\), the model is optimized with mean squared error:
\[
\mathcal{L}_{\mathrm{VP}} = (\hat{s} - s)^2.
\]
This formulation preserves the ordinal structure of the five rating categories while allowing the model to produce a continuous visual plausibility score aligned with averaged human judgments. The trained model achieves a Pearson Linear Correlation Coefficient~(PLCC) of 0.92 against ground-truth label, demonstrating strong human preference alignment.

\section{Additional Experimental Results}
\label{app:results}

\subsection{Full Split Results on Text-Driven Models}
\label{app:full_split_text}

\cref{tab:full_split_text} reports the complete per-sub-metric scores for all nine text-driven models evaluated on the \textbf{full test set} (\numvideo cases). Unlike \cref{tab:full_results_transposed}, which is restricted to the navigation subset shared across paradigms, this table includes all interaction types (navigation, event editing, subject action, and perspective switching) and uses the full-split consistency metrics (gated spatial consistency, geometric consistency, and photometric consistency).

Several patterns emerge. Kling~3.0 and Wan~2.7 achieve the strongest interaction adherence, particularly in event editing and subject action, while perspective switching remains universally difficult (average 30.7). LongCat-Video and HunyuanVideo lead consistency metrics but exhibit limited dynamic degree, confirming a trade-off between motion magnitude and temporal stability. Wan~2.7 leads causal fidelity by a wide margin (83.3 \vs the 74.7 average), while visual plausibility scores cluster within a narrow 4.6-point range across all models. No single text-driven model dominates across all evaluation axes.

\begin{table}[H]
\caption[Full split results on text-driven models]{Full per-sub-metric results for text-driven models on the \textbf{full test set} (\numvideo cases). All scores $\in [0,100]$, higher is better. \textbf{Bold} = best, \underline{underline} = second best per row.}
\vspace{+1mm}
\label{tab:full_split_text}
\centering
\renewcommand{\arraystretch}{1.4}
\setlength{\tabcolsep}{7.0pt}
\renewcommand{\icon}[1]{\raisebox{-0.15em}{\makebox[1.2em][c]{\includegraphics[height=1.1em,width=1.1em,keepaspectratio]{figures/icon/#1}}}\hspace{0.1em}}
\newcommand{\iconbgB}[2]{\cellcolor{#1}\rule{0pt}{1.6em}\raisebox{-0.05em}{\makebox[1.2em][c]{\includegraphics[height=1.1em,width=1.1em,keepaspectratio]{figures/icon/#2}}}}
\newcommand{\gB}[1]{\textcolor{gray}{#1}}
\begin{tabular}{@{}cl| *{9}{c} c @{}}
\thickhline
\textbf{} & \textbf{Metrics}
  & \rotatebox{90}{\textbf{Seedance 1.5}}
  & \rotatebox{90}{\textbf{Wan 2.7}}
  & \rotatebox{90}{\textbf{Kling 3.0}}
  & \rotatebox{90}{\textbf{YUME 1.5}}
  & \rotatebox{90}{\textbf{HY-Video 1.5}}
  & \rotatebox{90}{\textbf{LTX 2.3}}
  & \rotatebox{90}{\textbf{LongCat-Video}}
  & \rotatebox{90}{\textbf{Kairos}}
  & \rotatebox{90}{\textbf{Cosmos 2.5}}
  & \rotatebox{90}{\textit{Average}} \\
\noalign{\vskip -1pt}
\textbf{} & \textbf{}
  & \iconbgB{blue!10}{bytedance.png}
  & \iconbgB{blue!10}{wan.png}
  & \iconbgB{blue!10}{kling.jpeg}
  & \iconbgB{blue!10}{shlab.png}
  & \iconbgB{blue!10}{hunyuan.png}
  & \iconbgB{blue!10}{lightrix.jpeg}
  & \iconbgB{blue!10}{longcat.png}
  & \iconbgB{blue!10}{kairos.png}
  & \iconbgB{blue!10}{cosmos.png}
  & \\
\hline
\multirow{7}{*}{\rotatebox{90}{\textbf{Video Quality}}}
  & Aesthetic   & 59.7 & 59.6 & 61.3 & 59.3 & \underline{61.9} & 56.9 & \textbf{64.7} & 58.4 & 60.1 & \gB{60.2} \\
  & Imaging     & \underline{69.8} & 68.1 & 67.7 & 65.7 & 67.4 & 62.3 & \textbf{69.8} & 63.6 & 67.2 & \gB{66.8} \\
  & Flickering  & 93.4 & 93.0 & 94.5 & 94.8 & 95.5 & 94.1 & 94.9 & \textbf{96.3} & \underline{96.0} & \gB{94.7} \\
  & Dynamic     & \underline{98.3} & \textbf{99.3} & 89.9 & 86.1 & 68.8 & 94.4 & 59.7 & 63.5 & 42.4 & \gB{78.0} \\
  & Smoothness  & 97.6 & 96.6 & 97.9 & 97.7 & \textbf{98.8} & 96.8 & 97.7 & 97.9 & \underline{98.3} & \gB{97.7} \\
  & HPSv3-Norm  & \underline{72.9} & 69.4 & 68.8 & 62.0 & 67.5 & 57.7 & \textbf{76.3} & 58.8 & 65.9 & \gB{66.6} \\
  & \textbf{Average} & \textbf{82.0} & \underline{81.0} & 80.0 & 77.6 & 76.6 & 77.1 & 77.2 & 73.1 & 71.7 & \gB{77.2} \\
\hline
\multirow{3}{*}{\rotatebox{90}{\textbf{Setting}}}
  & Scene       & 71.6 & \underline{88.3} & \textbf{89.0} & 53.1 & 77.6 & 81.3 & 53.1 & 52.2 & 72.4 & \gB{71.0} \\
  & Subject     & \underline{94.3} & \textbf{94.6} & 92.9 & 91.7 & 93.6 & 89.2 & 91.5 & 88.5 & \underline{94.3} & \gB{92.3} \\
  & \textbf{Average} & 82.9 & \textbf{91.5} & \underline{91.0} & 72.4 & 85.6 & 85.2 & 72.3 & 70.3 & 83.3 & \gB{81.6} \\
\hline
\multirow{5}{*}{\rotatebox{90}{\textbf{Interaction}}}
  & Navigation  & 68.0 & 66.0 & 70.3 & \textbf{72.0} & \underline{71.8} & 67.6 & 63.1 & 65.1 & 64.1 & \gB{67.6} \\
  & Event Editing  & 80.4 & \textbf{84.0} & \underline{81.4} & 57.8 & 63.8 & 53.0 & 50.4 & 46.8 & 48.2 & \gB{62.9} \\
  & Subject Action & 80.0 & \underline{83.4} & \textbf{85.6} & 47.0 & 55.6 & 51.8 & 48.4 & 41.4 & 41.6 & \gB{59.4} \\
  & Persp.\ Switching & \underline{45.0} & \textbf{55.0} & \textbf{55.0} & 16.7 & 27.6 & 25.0 & 18.3 & 13.3 & 20.0 & \gB{30.7} \\
  & \textbf{Average} & 68.3 & \underline{72.1} & \textbf{73.1} & 48.4 & 54.7 & 49.4 & 45.1 & 41.6 & 43.5 & \gB{55.1} \\
\hline
\multirow{9}{*}{\rotatebox{90}{\textbf{Consistency}}}
  & Background  & 89.6 & 89.5 & \underline{92.7} & 92.0 & 92.4 & 89.3 & \textbf{94.7} & 91.8 & 92.4 & \gB{91.6} \\
  & Spatial     & 72.7 & 71.0 & 75.3 & 71.5 & \underline{79.2} & 70.2 & \textbf{83.3} & 76.8 & 78.1 & \gB{75.3} \\
  & Gated Spatial & 72.4 & 71.0 & \textbf{75.1} & 71.4 & \textbf{75.1} & 70.2 & 66.2 & 62.0 & \underline{74.3} & \gB{70.9} \\
  & Perspective & 62.7 & 62.2 & 76.8 & 48.0 & \textbf{86.6} & 69.8 & \underline{81.5} & 76.3 & 84.3 & \gB{72.0} \\
  & Segment     & 92.4 & 65.6 & 92.7 & \textbf{99.3} & \textbf{99.3} & 77.8 & \underline{98.6} & 94.1 & 93.1 & \gB{90.3} \\
  & Geometric   & 83.6 & 82.6 & 89.4 & 91.1 & \underline{94.4} & 81.1 & \textbf{94.7} & 91.5 & 94.2 & \gB{89.2} \\
  & Photometric & 76.7 & 75.5 & 80.4 & \textbf{84.1} & 81.4 & 79.4 & 81.5 & \underline{82.1} & \underline{82.1} & \gB{80.4} \\
  & Subject Cross & 89.3 & 88.7 & 88.5 & 89.4 & \underline{91.5} & 86.7 & \textbf{92.4} & 90.7 & 91.8 & \gB{89.9} \\
  & \textbf{Average} & 79.9 & 75.8 & \underline{83.9} & 80.9 & \textbf{87.5} & 78.1 & 86.6 & 83.2 & 86.3 & \gB{82.4} \\
\hline
\multirow{3}{*}{\rotatebox{90}{\textbf{Physical}}}
  & Causal Fidelity     & 76.0 & \textbf{83.3} & \underline{78.0} & 72.7 & 75.0 & 74.0 & 76.0 & 62.7 & 74.7 & \gB{74.7} \\
  & Visual Plausibility & \underline{60.5} & 59.8 & 60.4 & 58.1 & 59.3 & 56.3 & \textbf{60.8} & 58.2 & 59.3 & \gB{59.2} \\
  & \textbf{Average} & 68.2 & \textbf{71.6} & \underline{69.2} & 65.4 & 67.1 & 65.1 & 68.4 & 60.5 & 67.0 & \gB{67.0} \\
\thickhline
\end{tabular}
\end{table}

\subsection{Human-Preference Annotation Platform and Protocol}
\label{app:human_annotation}

This section provides additional details on the human-preference annotation study described in Section~\ref{sec:human_validation}, including the annotation platform interface, task design, annotator training, and quality control procedures.

\begin{figure}[H]
\centering
\includegraphics[width=0.7\textwidth]{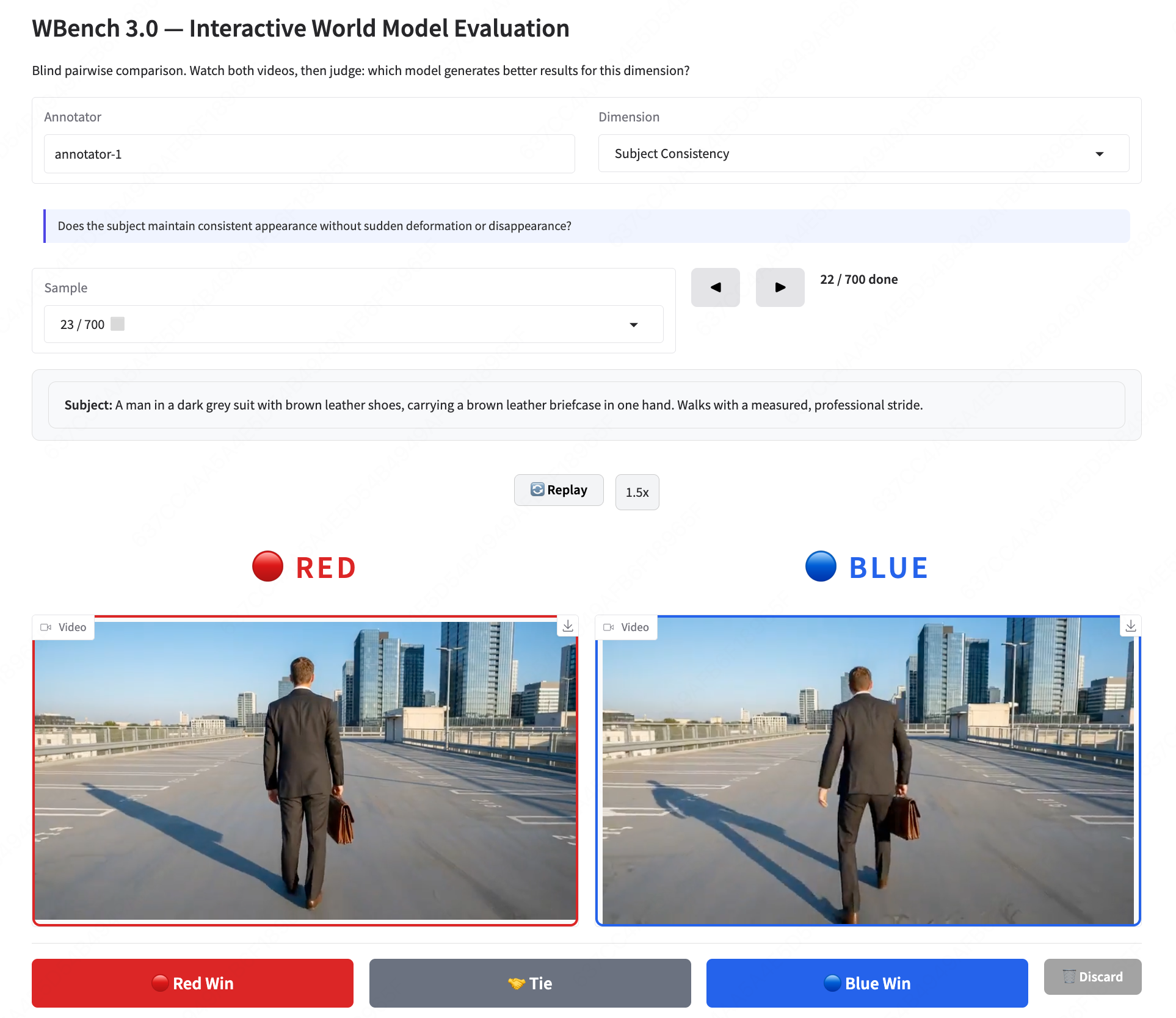}
\caption[Human-preference annotation platform]{Human-preference annotation platform. Each task is a blind pairwise comparison between two models, with the annotator choosing among \emph{A better}, \emph{B better}, \emph{Tie}, or \emph{Discard}.}
\label{fig:human_platform}
\end{figure}

\myparagraph{Platform overview.}
We deploy a custom web-based annotation platform (\cref{fig:human_platform}) built on top of an internal labeling system.
The platform supports side-by-side video playback with synchronized controls, allowing annotators to play, pause, seek, and replay both videos simultaneously before making a judgment.
Each annotation task presents a single pairwise comparison for one evaluation dimension.

\myparagraph{Task design.}
We organize the human study into nine dimension-level task sheets that together cover the four interaction-adherence metrics, two setting-level metrics, aesthetic quality, spatial consistency, and physics compliance. Each sheet contains a set of pairwise comparisons drawn from a fixed pool of evaluated models and navigation/semantic cases, and all pairs within a sheet follow the same per-dimension question. For most dimensions, we enumerate all $\binom{10}{2}=45$ pairs among the ten sampled models; for interaction adherence, only four text-driven I2V models expose the full set of non-navigation interactions and we enumerate all $\binom{4}{2}=6$ pairs. The left-right assignment of Video~A and Video~B is randomized independently for each task to eliminate position bias. Overall, the nine sheets amount to roughly $13.5$K pairwise annotation tasks.

\myparagraph{Dimension-specific instructions.}
Each task is presented with a short dimension-specific question, a contextual hint that discloses only the information needed for that dimension (so that annotators are not biased by unrelated axes), and a three-line checklist (\emph{focus / ignore / preference}) that explicitly decouples the target dimension from other quality signals. Each dimension is expanded below, with the contextual hint shown in italics and the \emph{focus / ignore / preference} lines listed verbatim from the platform.

\myparagraph{Annotator training.}
We recruit a large-scale crowdsource annotation pool of 400 annoors are required to complete a training phase consisting of:
\begin{itemize}[nosep,leftmargin=*]
  \item \textbf{Orientation session}: A 30-minute briefing explaining the five evaluation dimensions, the annotation interface, and common pitfalls (e.g., confusing visual quality with interaction accuracy).
  \item \textbf{Practice tasks}: Each annotator completes 20 practice comparisons with pre-determined gold labels and must achieve $\geq 80\%$ agreement to proceed. During this phase, the research team reviews annotator responses, identifies common mistakes and ambiguities, and provides clarifications to ensure consistent understanding of the annotation workflow and sub-dimension-specific criteria.

\end{itemize}

\myparagraph{Quality control.}
We employ several mechanisms to ensure annotation quality throughout the study:
\begin{itemize}[nosep,leftmargin=*]
  \item \textbf{Triple redundancy}: Every comparison is independently judged by three annotators; the final label is determined by majority vote.
  \item \textbf{Gold-standard checks}: 5\% of tasks are seeded with gold-standard pairs, where the quality difference is, where the an is clear and verified by the authors. Annotators whose accurwhose accuracy on gold-standard tasks falls below 85\% tasks agged for further further review.
  \item \textbf{Expert sampling review}: In addition to automatic quality checks, expert reviewers conduct sampled inspections over the completed annotations. For batches whose sampled accuracy is below 80\%, the corresponding annotations are rejected and returned for re-annotation. Experts also correct inconsistent or ambiguous labels when necessary, thereby mitigating residual noise from crowdsource judgments.
\end{itemize}

\myparagraph{Annotation statistics.}
In total, the nine dimension-level sheets contain $13{,}515$ pairwise comparison tasks spanning $8$ evaluated models and $10$ evaluation dimensions. These tasks are completed by a 400-person crowdsource annotation pool under the quality-control protocol described above, including triple redundancy, gold-standard checks, completion-time filtering, and expert sampled review. The per-dimension human-metric alignment that results from this study is reported in \cref{fig:human_alignment} in the main text.

\section{Broader Impact}
\label{app:broader_impact}

\benchmark is an evaluation benchmark that diagnoses the capabilities and limitations of interactive world models without generating or distributing synthetic media itself. All test data are constructed from synthetic or openly licensed imagery without depicting identifiable real individuals. By exposing fine-grained failure modes in physics, controllability, and consistency, the benchmark can guide the development of more reliable world models with downstream benefits for simulation, robotics, gaming, and education. While progress in interactive world modeling may indirectly benefit the creation of realistic synthetic media, we mitigate this risk by releasing only the evaluation toolkit and benchmark metadata rather than generative model weights, and by designing test cases that span diverse visual styles rather than optimizing for photorealistic identity synthesis.

\end{document}